\documentclass{article}

\PassOptionsToPackage{numbers,square,semicolon,sort&compress}{natbib}

\usepackage[main,final]{neurips_2026}

\usepackage[utf8]{inputenc} 
\usepackage[T1]{fontenc}    
\usepackage[colorlinks=true,
]{hyperref}
\usepackage{xr-hyper}
\usepackage{url}            
\usepackage[para]{footmisc}
\usepackage{booktabs}       
\usepackage{amsfonts}       
\usepackage{nicefrac}       
\usepackage{microtype}      
\usepackage{xcolor}         

\usepackage{algorithm}
\usepackage{algorithmic}

\usepackage{mathrsfs}
\usepackage{mathtools}
\usepackage{graphicx}
\usepackage{enumitem}

\usepackage{caption}
\usepackage{subcaption}

\usepackage{cancel}

\usepackage{academicons}
\usepackage{orcidlink}

\usepackage[disable,textsize=small]{todonotes}

\usepackage{macros}

\title{Energy-Tweedie: Score meets Score, Energy meets Energy}

\author{%
    Andrej Leban\\
    Department of Statistics,\\
    University of Michigan,\\
    Ann Arbor, MI, United States\\
    \texttt{leban@umich.edu} \\
}

\hypersetup{pdfauthor={Andrej Leban}}

\begin{document}

    \date{}

    \maketitle

    \begin{abstract}
        Denoising and score estimation are classically linked through Tweedie’s
        formula, which relates the posterior mean under Gaussian noise to the Stein score of the noisy marginal.
        In this work, we extend this perspective beyond Gaussian noise to a broad class of Gibbs (energy-based) noise distributions, with the generalized Gaussian family as an example.
        We derive the Energy-Tweedie identity: when the denoising posterior is
        viewed through the lens of scoring rules, the path derivative of a kernel scoring rule defined by the noise potential recovers the Stein score
        of the noisy marginal.
        Thus, the familiar correspondence between Gaussian noise, posterior means, squared loss, and Tweedie’s formula is lifted to a distributional correspondence between Gibbs noise distributions, full posterior laws, kernel scoring rules, and the
        Energy-Tweedie identity, which gives a Tweedie-style relation for each noise potential.
        Among its consequences, this identity gives a posterior-samples-to-score route to score estimation, yields a principled criterion for
        estimating unknown noise parameters, and enables diffusion-style sampling along user-chosen paths through the noise-parameter space, supplying the score-based perspective on recent generative methods trained with scoring rules.%
        \footnote{Code repository: \url{https://github.com/andleb/et}.}
    \end{abstract}

    \section{Introduction}

    Denoising, empirical Bayes, and score-based generative modeling are classically tied together by Tweedie’s formula. Under additive Gaussian noise $Y=X+\epsilon$, the posterior mean $\EE[X\mid Y=y]$  is the MSE-optimal denoiser and its residual determines the Stein score $s_m(y)=\nabla_y\log m(y)$ of the noisy marginal.
    This correspondence is one reason Gaussian denoising occupies such a central position in score-based diffusion.
    Once the noise is non-Gaussian, however, the posterior mean is typically no longer the natural object exposed by the score. Our main result, the Energy-Tweedie identity,  connects the score of the noisy marginal to the path derivative of a scoring rule determined by the noise potential (such as the \textit{energy score}), and evaluated at the denoising posterior.%
    \footnote{Somewhat coincidentally,  we thus connect two pairs of previously-unrelated homonyms: both \textit{energy} and \textit{score} have double meanings.}
    The familiar correspondence between Gaussian noise, posterior means, squared loss, and Tweedie's formula is thus lifted to
    a correspondence between Gibbs noise, full posterior laws, kernel scoring rules, and the Energy-Tweedie identity,
    as established in Theorem~\ref{thm:et}. Cor.~\ref{cor:gen_gaussian} provides a summary of the correspondence using the contrast between the Gaussian and generalized Gaussian cases.
    The main contributions of this work are thus:
    \begin{itemize}[nosep,leftmargin=*]
        \item \textbf{A general identity.} The Energy-Tweedie identity provides the noisy-marginal score for any normalizable Gibbs noise distribution (Thm.~\ref{thm:et}), establishing a connection between the former and scoring rules.
        \item \textbf{A posterior-samples-to-score route.} Any sufficiently accurate sampler of the denoising posterior becomes a score estimator.
        \item \textbf{Noise-parameter estimation.} We propose an M-estimation method recovering
       all the unknown noise parameters of any Gibbs noise distribution from noisy data, with theoretical guarantees for the generalized Gaussian case.
        \item \textbf{Diffusion along noise-parameter paths.} We show how posterior models can be used inside traditional score-based samplers beyond Gaussian noising, with the \textit{path} through the noise parameters a design variable chosen freely at sampling time.

    \end{itemize}

    \section{Background}
    \label{sec:background}

    \subsection{Denoising, Tweedie's Formula, and Score Matching}

    We denote the clean data as $X\sim p$ and the noise, independent of $X$, as $\epsilon \sim q$, where $q$ comes from a location-equivariant family.
    Let the noisy data be $Y = X + \epsilon$, with $X, \epsilon, Y \in \R^d$. The noisy marginal is then:
    \begin{equation}
        m(y) \;=\; (p*q)(y) \;=\; \int p(x)\,q(y-x)\,dx,
    \end{equation}
    where ``$*$'' denotes the convolution operation. The \textit{Stein score} of the noisy marginal is then defined as:
    \begin{equation}
        s_m(y) \; = \;\nabla_y \log m(y).
    \end{equation}
\todo{COMPACT 29: Reduce lead-in to one sentence; retain observation-versus-parameter differentiation.}
    A well-known result for $s_m$ is Fisher's formula \citep{louis_finding_1982, cappe_inference_2005}. In our case, we
    will be plugging in $y$ --- the ``noisy'' ambient variable --- instead of the parameter $\theta$; that is, switching
    from the Fisher to the Stein score. The result when using additive noise $q$ is:
\todo{COMPACT 13: Keep the expectation form only; retarget the path-derivative introduction and other references.}
    \begin{equation}
        s_m(y)  = \int_\gX \nabla_{y'} \log q(y' - x)\vert_{y'=y} ~ p(x\mid Y=y) \, dx,
        \label{eq:general_tweedie_int}
    \end{equation}
    or equivalently:
    \begin{equation}
        s_m(y)=\EE_{X \sim P(\cdot|Y=y)}\left[\nabla_y \log q(y-X)\right].
        \label{eq:general_tweedie}
    \end{equation}
    Eq.~\ref{eq:general_tweedie} can be seen as a ``generalized Tweedie'' formula. Setting $q = \gN(0, \Sigma)$, we
    arrive at the ``standard'' Tweedie's formula \citep{robbins_empirical_1956, efron_tweedies_2011}:
    \begin{equation}
        \EE[X\mid Y=y] \,=\, y + \Sigma \, s_m(y)
        \label{eq:tweedie}
    \end{equation}
    The derivation and more details can be found in App.~\ref{app:proof_gauss}.

    \subsection{Scoring Rules and the Energy Score}

    Scoring rules were introduced as a measure for the evaluation of \textit{probabilistic} forecasts: thus, they take
    in the predictive distribution and the value (taken to be from the ground truth distribution) that materializes,
    outputting a single value --- the \textit{score} \citep{gneiting_strictly_2007, parry_proper_2012, ehm_local_2012}.
    For a general \textit{kernel} $\rho(x, y)$, the \textit{kernel score} is defined as:
    \begin{equation}
        S_\rho(P,y) = \mathbb E_{Y\sim P}[\rho(Y,y)] - \frac12 \mathbb E_{Y,Y'\stackrel{\mathrm{iid}}{\sim}P} [\rho(Y,Y')],
        \label{eq:KS_def}
    \end{equation}
    where $P$ is the predictive distribution, and $y$ is a sample from the true distribution.
\todo{COMPACT 32: State the minimization sign convention parenthetically.}
    We have \textbf{adopted a negative sign} before the scoring rule in order to match the usual objective \textit{minimization} convention in
    machine learning.
    The first term measures the distance between the forecast and the observations and would be minimized by a collapsed point forecast. The second term rewards admitting uncertainty; together, the two terms make honest reporting optimal in expectation.
    In this convention, a scoring rule is said to be \textit{proper} if
    $$\EE_{y \sim Q} [S_\rho(P, y)] \geq \EE_{y \sim Q} [S_\rho(Q, y)].$$
    It is said to be \textit{strictly proper} if we have the equality above \textit{iff} $P \equiv Q$ --- the expected
    scoring rule is minimized \textit{uniquely} by the ground truth distribution.
    A common example is the \textit{energy score}, introduced in \citet{gneiting_strictly_2007},
    \begin{equation}
        \mathrm{ES}_\beta(P, y) = \EE_{Y\sim P}\left[\|Y - y\|_2^\beta \right] -\tfrac12\,\EE_{Y, Y^\prime \iid
            P}\left[ \|Y - Y'\|_2^\beta\right],
        \label{eq:ES_def}
    \end{equation}
    where $\beta > 0$, and $\rho(x, y) = \|x - y\|_2^\beta$ is the kernel in this case.
    While existing work has covered $\| \cdot \|_\alpha$  with $\alpha$ a scalar,
\todo{COMPACT 25: Inline the Mahalanobis-distance definition; retain positive definiteness.}
    we will be focusing on the \textit{Mahalanobis distance} (for some $\Sigma \succ 0$):
    \begin{equation}
        \rho_{\Sigma^{-1}}(x,y)=\|x-y\|_{\Sigma^{-1}}=\sqrt{(x-y)^\top \Sigma^{-1} (x-y)}
    \end{equation}
    and the $\beta$-th power of the corresponding norm $\| \cdot \|_{\Sigma^{-1}}$ in our examples.
\todo{COMPACT 14: Define by replacing the norm in the ordinary energy score; retarget references.}
    The corresponding (negative) scoring rule is:
    \begin{equation}
        \mathrm{ES}_{\Sigma^{-1}, \beta} (P, y) ~=~ \EE_{Y \sim P} \left[ \| Y - y \|_{\Sigma^{-1}}^\beta \right] -
            \tfrac{1}{2} \EE_{Y, Y^\prime \iid P} \left[ \left\| Y - Y^{\prime}\right\|_{\Sigma^{-1}}^\beta \right]
        \label{eq:non-eucl-ES}
    \end{equation}


\section{Tweedie-Like Formula for Gibbs Noise}
\label{sec:gibbs}

    We start this section by further specifying the family of $q$ we will be considering from now on:
    \begin{definition}[Noise potential and the Gibbs distribution]
        \label{def:ellip_dist}
        Let $\gE:\mathbb R^d\to[0,\infty)$ be a measurable and symmetric \textit{noise potential} with $\gE(0)=0$.
\todo{COMPACT 08: Keep density displayed; inline normalizing integral and admissible scales, preserving references.}
        The matched Gibbs distribution is defined as:
        \begin{equation}
            q_{\gE,c}(u)=\frac{1}{Z_\gE(c)} e^{-c\,\gE(u)},
            \label{eq:gibbs_def}
        \end{equation}
        where $c$ is the \textit{scale parameter} and
        \begin{equation}
            Z_\gE(c)=\int_{\mathbb R^d} e^{-c\,\gE(u)}\,du
            \label{eq:norm_const}
        \end{equation}
        is the normalizing constant, which we require to be finite. Thus, the admissible scales form the set:
        \begin{equation}
            \Lambda_\gE=\bigl\{c>0:\ Z_\gE(c)<\infty\bigr\}.
            \label{eq:Lambda_E}
        \end{equation}
\todo{COMPACT 22: Compress definition trailer; retain parameterization and energy-model connection.}
        That is, we (only) require \textit{normalizability}.
    \end{definition}\postdisplaystatementskip
    The density is thus part of the broader location-equivariant family.
    Note that the potential $\gE$  itself can include further parameters, hence the distribution is parameterized by $(\gE, c)$,
    where the former is to be understood as a particular choice of the functional form.
    In machine learning, this distributional form is a version of what is commonly known as an \textit{energy model}
    \citep{murphy_probabilistic_2023}.
    We get for the score of the noising kernel:
    \begin{equation}
        \nabla_y\log q_{\gE,c}(y - X) = -c\,\nabla_y \gE(y-X)
        \label{eq:gradE}
    \end{equation}
    We will be plugging this into Eq.~\ref{eq:general_tweedie}, and adopting the ``path derivative'' notation from
    Variational Inference literature \citep{roeder_sticking_2017}: let $\nabla^{\mathrm{PD}}$ denote a ``fixed-measure''
    gradient --- taking the gradient of the function, but not the measure, exactly as the operation is performed in
    Eq.~\ref{eq:general_tweedie_int}:
    \begin{equation}
        \nabla^{\mathrm{PD}}_y \,\EE_{X\sim P(\cdot\mid Y=y)}[f(y,X)] \Def ~ \int_\gX \nabla_{y^\prime} f(y^\prime,
            x)\vert_{y'=y} ~ p(x\mid Y=y) \, dx.
        \label{eq:path_derivative}
    \end{equation}
\todo{COMPACT 09: Replace the stop-gradient display with a parenthetical; keep the fixed-measure definition.}
    Another terminology (stemming from the implementation) is the \textit{stop-gradient} (\texttt{sg}) operation applied
    to the distribution; namely:
    $$
    \nabla_y^{\mathrm{PD}} \EE_{X \sim P(\cdot \mid Y=y)}[f(y, X)]=\EE_{X \sim P(\cdot \mid Y=y)}\left[\nabla_y f(y,
        \operatorname{sg}(X))\right].
    $$
    We then arrive at the following Tweedie-like formula for Gibbs noise:
    \begin{proposition}[Tweedie-Like Formula for Gibbs Noise]
    \label{prop:gibbs_tweedie}
    For Gibbs noise distributions under the following assumptions:
        \vspace{-6pt}
        \begin{itemize}
        \setlength{\itemsep}{-4pt}
        \setlength{\parsep}{-4pt}
        \item (A1) $c \in \Lambda_\gE$;
        \item (A2) the generalized Tweedie formula (Eq.~\ref{eq:general_tweedie}) holds at $y$ for $q_{\gE, c}$;
        \item (A3) $\gE$ is differentiable outside a Lebesgue-null set $N$ with $P((y - X) \in N \mid Y=y) = 0$;
        \item (A4) $\EE\bigl[\|\nabla\gE(y-X)\|\mid Y=y\bigr]<\infty$.
        \end{itemize}
        \vspace{-6pt}
        Eq.~\ref{eq:general_tweedie} becomes:
        \begin{equation}
            \boxed{
                    s_m(y)
                    = \EE_X \left[- c\,\nabla_y \gE(y-X) \mid Y=y\right]
                    = - c \, \nabla_y^{\mathrm{PD}}~ \EE_X \left[\gE(y -X)\mid Y=y\right]
            }
           \label{eq:gibbs_tweedie}
        \end{equation}
    \end{proposition}\postdisplaystatementskip%
    A short proof can be found in App.~\ref{app:proof_gibbs_tweedie}.
\todo{COMPACT 26: Remove restatement; retain the following Efron positioning.}
    In words, the score at $y$ coincides with the negative
    path-gradient (w.r.t. $y$) of the expected (w.r.t. the posterior) potential centered around the same $y$. In the case of the exponential family of distributions,
    \citet{efron_tweedies_2011} has derived related Tweedie's formulas for the first two moments.
    We would like to note that a Gibbs distribution is not necessarily an exponential family
    distribution when viewed as a family in the location parameter $x$; this result thus provides Tweedie identities beyond \citep{efron_tweedies_2011}. Some examples follow below.

    \subsection{Examples}
    \label{sec:elliptic_examples}

\todo{COMPACT 37: Consider inlining Gaussian and Laplace formulas only if first-reading clarity is preserved.}
    Three familiar special cases illustrate which summary of the denoising posterior the score encodes;
    derivations are deferred to App.~\ref{app:examples_derivations}. For Gaussian noise,
    $\gE(u)=\|u\|_{\Sigma^{-1}}^2$, with $c = \frac12$ constant, and Eq.~\ref{eq:gibbs_tweedie} recovers the ordinary Tweedie
    residual:
    $$
    s_m(y)=\Sigma^{-1}\left(\EE[X\mid Y=y]-y\right).
    $$
    The score therefore points toward the posterior mean in $\Sigma^{-1}$ geometry. For multivariate Laplace noise,
    $q(u)\propto \exp(-\lambda\|u\|_{\Sigma^{-1}})$, hence $\gE(u)=\|u\|_{\Sigma^{-1}}$ and  $c = \lambda$, and
    the score averages only directions from $y$ to posterior samples:
    $$
    s_m(y)=-\lambda\,\EE\!\left[\frac{\Sigma^{-1}(y-X)}{\|y-X\|_{\Sigma^{-1}}}\,\middle|\,Y=y\right].
    $$
    This replaces the Gaussian's \textit{mean-seeking} behavior with a more robust, approximately
    \textit{median-seeking} behavior, pointing toward the posterior geometric median in the same metric.
    \vspace{-10pt}
    \paragraph{Generalized Gaussian}
    We use the parameterization:
    \begin{equation}
        q(u)\ \propto\ \exp \Big( - \frac{\lambda}{\beta} ~ \|u\|_{\Sigma^{-1}}^{\beta} \Big),
        \label{eq:gen_gaussian}
    \end{equation}
\todo{COMPACT 33: Remove unused conventional-scale conversion; retain the distribution citation.}
    for $\beta, \lambda > 0$, which relates to the common scale parameter $\alpha$ via
    $\alpha = ({\beta} / {\lambda})^{1/\beta}$ \citep{zhang_multivariate_2013}.
    Thus $\gE(u)=\|u\|_{\Sigma^{-1}}^\beta$ and $c = \frac{\lambda}{\beta}$.
    The normalizing constant is finite for all $c>0$, so $\Lambda_\gE=(0,\infty)$.
\todo{COMPACT 30: Remove repeated qualification here; retain the Efron discussion and Gaussian specialization.}
    This is a location family, but not generally an exponential family distribution in the denoising setting with unknown location parameter $x$;
    the Gaussian case is recovered at $(\beta=2, \lambda=1)$.
    \begin{lemma}
        Inheriting the assumptions of Prop.~\ref{prop:gibbs_tweedie}, generalized Gaussian noise has score
        \begin{equation}
            s_m(y) = -\lambda ~ \Sigma^{-1} \,
            \EE\left[  \|y - X\|_{\Sigma^{-1}}^{\beta - 2} ~ (y - X) \mid Y=y \right].
            \label{eq:gen_gaussian_score}
        \end{equation}
        \label{lem:gen_gaussian_score}
    \end{lemma}\postdisplaystatementskip%
\vspace{-10pt}
    A short proof and the verification of the assumptions can be found in App.~\ref{app:proof_gen_gaussian_score}.
    The factor $\|y-X\|_{\Sigma^{-1}}^{\beta-2}$ controls which posterior samples dominate: $1<\beta<2$ values trade off local (Laplace-like) robustness against distance-sensitive (Gaussian-like) residual weighting. This generalized Gaussian family will be used in the practical examples.

    \section{The Energy-Tweedie Identity}%
    \label{sec:ET}

\todo{COMPACT 11: Inline the discrepancy; shorten the derivative display to first and last expressions.}
    Define the discrepancy induced by the noise potential $\gE$ as:
    \begin{equation}
        \rho(x,y) = \gE(y-x).
    \end{equation}
    As the noise potential is symmetric with $\gE(0)=0$, we have $\rho(x,y) = \rho(y,x)$ and $\rho(x,x)=0$.
    Next, the path-derivative of the associated kernel score (Eq.~\ref{eq:KS_def}) is:
    \begin{equation}
        \nabla^{\mathrm{PD}}_y S_{\rho} (P, y) ~=~
        \nabla^{\mathrm{PD}}_y \left( \EE_{X \sim P} [\gE(y - X)] - \tfrac12\EE_{X, X' \iid P} [\gE(X' - X)]  \right)~=~
        \EE_{X \sim P} [ \nabla_y \gE(y - X)].
    \end{equation}
\todo{COMPACT 28: Reduce to one sentence introducing evaluation at the posterior.}
    We are interested in what this scoring rule tells us in the denoising context: i.e., adopt $P$ to be the denoising
    posterior $P = P(X \mid Y=y)$ and examine the path-gradient above when the latter is supplied with a noisy sample $y$.
    We obtain after comparison with Prop.~\ref{prop:gibbs_tweedie}:

    \begin{theorem}[The Energy-Tweedie Identity]
        For data $X \sim p$ noised with Gibbs noise (Eq.~\ref{eq:gibbs_def}) with the noise potential $\gE$,
        the associated discrepancy $\rho$ and the scale parameter $c$, assume at a  noisy data sample $y$ a finite $\rho$-moment:
        $\mathbb E_{X,X'\stackrel{\mathrm{iid}}{\sim}P(X \mid Y=y)}\left[\rho(X,X')\right] <\infty.$
        Under the assumptions of Prop.~\ref{prop:gibbs_tweedie}, the Energy-Tweedie identity then links the \emph{Stein score} of
        the noisy marginal to the (path-derivative) gradient of the induced kernel score at the denoising posterior:
        \begin{equation}
            \boxed{
                s_m(y)\;=\;-c\,\nabla_y^{\mathrm{PD}}\,S_\rho(P(X \mid Y=y),\,y).
            }
            \label{eq:thm_et}
        \end{equation}
        \label{thm:et}
\todo{COMPACT 21: Remove theorem restatement; retain assumptions and identity.}
        In other words, the Stein score of the noisy marginal corresponds to the path-gradient of a matched kernel
        score induced by the noise potential with respect to the \emph{true} posterior. We thus obtain a \emph{family}
        of Tweedie-style relations, determined by the noise potential $\gE$.
    \end{theorem}
\todo{COMPACT 41: Shorten; retain loss-choice freedom and distinguish it from the Gaussian mean-only exception.}
    On propriety: the Energy-Tweedie identity (Eq.~\ref{eq:thm_et}) is algebraic and does not depend on the
    propriety of the matched scoring rule $S_\rho$. If one were, however, to plug in an estimator $P_{\theta}(X \mid Y)$,
    it would be preferable for it to be trained via a \textit{strictly proper} scoring rule; it being $S_\rho$ is \textit{unnecessary}, though.
    The following proposition shows that propriety is determined by the induced discrepancy $\rho$ and hence the noise potential $\gE$ alone.
    \begin{proposition}[Propriety of the Matched Kernel Score]
        Independently of (A1)--(A4) in Prop.~\ref{prop:gibbs_tweedie}, for the induced $S_{\rho}$ in Thm.~\ref{thm:et}
        and distributions with finite $\rho$-moment $\EE_{Z,Z'\iid P}[\rho(Z,Z')]<\infty$:
        \vspace{-6pt}
        \begin{enumerate}
        \setlength{\itemsep}{-4pt}
        \setlength{\parsep}{-4pt}
            \item $S_\rho$ is \emph{proper} iff $\rho$ is a \emph{conditionally negative definite} (CND) kernel:\\
                $\rho$ is symmetric, with $\sum_{i=1}^n \sum_{j=1}^n a_i a_j \, \rho(z_i, z_j) \leq 0$ for all $n$, all
                $z_1, \ldots, z_n \in \R^d$, and all $a_1, \ldots, a_n \in \R$ such that $\sum_i a_i = 0$;
            \item $S_\rho$ is \emph{strictly proper} iff $\rho$ has \textit{strong negative type}: the
                \emph{generalized energy distance}
                $$D_\rho(P,Q)=2\,\EE_{Z\sim P,\,W\sim Q}[\rho(Z,W)] - \EE_{Z, Z' \iid P}[\rho(Z,Z')]-\EE_{W, W' \iid Q} [\rho(W,W')]$$
                vanishes only at $P=Q$.
        \end{enumerate}
        \vspace{-6pt}
        If either holds, the finite $\rho$-moment condition of Thm.~\ref{thm:et} is automatically satisfied at the denoising posterior.
        \label{prop:matched_propriety}
    \end{proposition}
    The proof is given in App.~\ref{app:proof_matched_propriety}.

    \begin{corollary}[The (generalized) Gaussian case]
        For the \emph{generalized Gaussian}: $\gE(u)=\|u\|_{\Sigma^{-1}}^{\beta}$ and $c=\lambda/\beta$, the matched kernel score is  the \emph{Mahalanobis energy score} (Eq.~\ref{eq:non-eucl-ES}):
        $S_\rho=\mathrm{ES}_{\Sigma^{-1},\beta}.$
         $\mathrm{ES}_{\Sigma^{-1},\beta}$ is \emph{proper} for $\beta \in (0, 2]$ and \emph{strictly proper} for $\beta \in (0, 2)$.
        In this case, the Energy-Tweedie identity (Eq.~\ref{eq:thm_et}) becomes:
        \begin{equation}
            s_m(y)=-\frac{\lambda}{\beta}\,\nabla_y^{\mathrm{PD}}\,\mathrm{ES}_{\Sigma^{-1},\beta}(P(X \mid Y=y),\,y)
            \;=\; -\lambda ~ \Sigma^{-1} \, \EE\left[  \|y - X\|_{\Sigma^{-1}}^{\beta - 2} ~ (y - X) \mid Y=y \right].
            \label{eq:es_tweedie}
        \end{equation}%
\todo{COMPACT 42: Shorten to one sentence retaining mere propriety and identification of the mean only.}
        As is expected, we recover the ``classical'' relations in the ordinary Gaussian case $\beta=2, \lambda=1$:
        $D_\rho(P,Q)=2\,\|\mathbb E_{Z\sim P}[Z]-\mathbb E_{W\sim Q}[W]\|_{\Sigma^{-1}}^2$ vanishes as soon as the \emph{means} agree.
        As the rule is no longer \emph{strictly} proper, it pins down only the posterior mean.
\todo{CUT CANDIDATE: Replace the correspondence table in this corollary with an inline list.}
        We summarize the correspondence below:
        \begin{center}
        \renewcommand{\arraystretch}{1.15}
        \begin{tabular}{@{}p{.15\linewidth}p{.42\linewidth}p{.33\linewidth}@{}}
        \toprule
        Object & Generalized & Gaussian \\
        \midrule
        Noise distribution & $\propto
            \exp(-\frac{\lambda}{\beta}\|u\|_{\Sigma^{-1}}^\beta)$
            & $\propto\exp(-\frac12u^\top\Sigma^{-1}u)$ \\
        Objective & $\mathrm{ES}_{\Sigma^{-1},\beta}(P(X\mid Y=y),x)$
            & $\|\EE[X\mid Y=y]-x\|_{\Sigma^{-1}}^2$ \\
        Posterior object & $P(X\mid Y=y)$ & $\EE[X\mid Y=y]$ \\
        Score identity &
            $-\frac{\lambda}{\beta}\nabla_y^{\mathrm{PD}}
            \mathrm{ES}_{\Sigma^{-1},\beta}(P(X\mid Y=y),y)$
            & $\Sigma^{-1}(\EE[X\mid Y=y]-y)$ \\
        \bottomrule
        \end{tabular}
        \end{center}
        \label{cor:gen_gaussian}
    \end{corollary}

    Proofs of the propriety of $\mathrm{ES}_{\Sigma^{-1},\beta}$ and the $\beta=2$ collapse can be found in App.~\ref{app:proof_gen_gaussian}.
\todo{COMPACT 44: Shorten but retain the Laplace/ordinary-energy-score connection; it is distinct from the earlier example.}
    For $\beta=1$ and $\Sigma=I_d$ we obtain a correspondence between isotropic Laplace noise and  the \textit{ordinary energy score}:
    $s_m(y)=-\lambda\nabla^{\mathrm{PD}}_y\mathrm{ES}\big(P(X\mid Y=y),y\big)$.
    In the Gaussian case $\beta=2, \lambda=1$, plugging in an \textit{estimator} trained with a \textit{merely} proper ($\beta=2$) energy score
    suffices; one example is the denoising diffusion model trained via score matching.
    Thm.~\ref{thm:et} also has a geometric reading, which we develop in App.~\ref{app:connections} for the generalized Gaussian case: in the Riemannian geometry induced by $\Sigma^{-1}$, the noisy score is related to the path-derivative of the posterior mean $\beta$-energy. \todo{CUT CANDIDATE: this is too verbose}

    \begin{corollary}[The Student-t case]
        Take $\gE(u)=\log\bigl(1+\|u\|_{\Sigma^{-1}}^2\bigr)$ with $c \in \Lambda_\gE=(d/2,\infty)$.
        Then $q_{\gE, c}$ can be identified with the \emph{multivariate Student-t} distribution by choosing the degrees of freedom $\nu=2 c - d$
        and the scale matrix $\Sigma / \nu$, and the identity (Eq.~\ref{eq:thm_et}) holds for every data distribution $p$:
        \begin{equation}
            s_m(y)\;=\; -\,c\,\nabla_y^{\mathrm{PD}}\,S_\rho\bigl(P(X\mid Y=y),\,y\bigr)
            \;=\;-\,2c\,\Sigma^{-1}\,\EE_{X\sim P(\cdot\mid Y=y)} \!\left[\frac{y-X}{1+\|y-X\|_{\Sigma^{-1}}^2}\right].
            \label{eq:student_t_score}
        \end{equation}
        The induced $S_\rho$ is \emph{strictly proper} relative to the class of probability measures $P$ on $\R^d$ with $\EE_{Z\sim P}[\log(1+\|Z\|_{\Sigma^{-1}}^2)]<\infty$.
        \label{cor:student_t}
    \end{corollary}
    Proofs can be found in App.~\ref{app:proof_student_t}.
\todo{modified: removed dangling propriety transition.}
    The Student-t case is an example where the theorem gives the score for a \emph{heavy-tailed noise distribution even if its mean is undefined},
    with a \textit{strictly proper} scoring rule.

    \section{Applications and Empirical Consequences}
    \label{sec:consequences}

    In this section, we will focus on the generalized Gaussian noise case (Cor.~\ref{cor:gen_gaussian}) for the concrete examples, as it is a flexible family that includes many of the most common
    noise distributions, and the matched kernel score includes the energy score, perhaps the most commonly used multivariate scoring rule.

    \subsection{Score Estimation}
    \label{sec:score_estimation}

    Using Theorem~\ref{thm:et} one can get a noisy score estimate at any noising distribution
    parameter by plugging in a well-specified trained posterior estimator $P_\theta(\cdot
    \mid Y)$. Using samples from a posterior model $X^{(i)} \sim P_\theta(X \mid Y=y)$, the noisy
    score Monte Carlo approximation in the generalized Gaussian case (Cor.~\ref{cor:gen_gaussian}, with the conditional score from Lemma~\ref{lem:gen_gaussian_score}) is:
\todo{modified: lemma cite}
    \begin{equation}
        \widehat{s}_m (y) = - \lambda ~\Sigma^{-1} ~ \frac{1}{N} \sum_{i=1}^N \left( \| y - X^{(i)}
            \|_{\Sigma^{-1}}^{\beta - 2} ~(y - X^{(i)}) \right)
        \label{eq:sm_finite_sample}
    \end{equation}
    The bias-variance decomposition for this estimator is given in App.~\ref{app:bias_variance_score}.
    We wish to examine the \textit{clean} score estimation aspect.
    Write $\Sigma=\varepsilon^2\Sigma_0$, and let $q_\varepsilon$ be the corresponding generalized Gaussian noising
    density. Under suitable regularity conditions, made explicit in App.~\ref{app:proof_richardson}, the noisy score
    converges to the clean data score as $\varepsilon \downarrow 0$:
    \begin{equation}
        s_{m_\varepsilon}(y)=\nabla_y\log(p*q_\varepsilon)(y)\longrightarrow \nabla_y\log p(y).
        \label{eq:small_noise_score_limit}
    \end{equation}
\todo{COMPACT 39: Shorten positioning while retaining freedom to use a suitable posterior model.}
    This approach can be viewed as a  distinct route to score estimation: score estimation from the posterior, joining direct score estimation in the line of \citet{hyvarinen_estimation_2005} and denoising score estimation in the line of \citet{vincent_connection_2011}. While adjacent to the latter, it applies to any posterior model $P_\theta(X \mid Y )$ and does not prescribe the particular training objective.
    The Energy-Tweedie identity also allows denoising score estimation, which we denote \emph{Energy-Tweedie Denoising Score Matching} (ET-DSM). Using Theorem~\ref{thm:et}, we learn a neural score field $s_\theta(y,\eta)$ from clean--noisy pairs $(X, Y)$ by minimizing
    \begin{equation}
        \mathcal{L}_{\mathrm{ET\text{-}DSM}}(\theta;\eta)
        = \EE_{X\sim p,\,Y\sim q_\eta(\cdot\mid X)}
        \left[\left\|s_\theta(Y,\eta)+c\,\nabla_Y^{\mathrm{PD}}S_\rho(\delta_X,Y)\right\|_2^2\right].
        \label{eq:et_dsm_loss}
    \end{equation}
    Here $\delta_X$ is the point mass at $X$, and $\eta$ denotes the noise parameters for the chosen noise distribution.
    For generalized Gaussian noise, the regression target is
    given by Eq.~\ref{eq:sm_finite_sample} with $N=1$ and $X^{(1)}=X$.
\todo{COMPACT 03: Keep the Richardson estimator and bias-order claim; move explanation and extra display to appendix.}
    In practice, the clean score limit (Eq.~\ref{eq:small_noise_score_limit}) can be approximated by either (or \textit{both}) (i)  letting the estimator $P_{\theta,
        \varepsilon}$ take $\varepsilon$ as a parameter, and (ii) using numeric techniques such as \textit{Richardson
        extrapolation} \citep{press_numerical_2007}.%
    \footnote{Note: this works because $q$ is in the symmetric location family.}%
    Namely, under sufficient regularity conditions,  we can define:
    \begin{equation}
        \hat{s}_{\mathrm{RE}}(y) \Def \frac{\varepsilon_2^2 \hat{s}_{\varepsilon_1}(y)-\varepsilon_1^2
            \hat{s}_{\varepsilon_2}(y)}{\varepsilon_2^2-\varepsilon_1^2} 
        \label{eq:richardson_extrapolation}
    \end{equation}
    Then:
    $$
    \hat{s}_{\mathrm{RE}}(y) = \nabla \log p(y) + \OO(\varepsilon^4) + (\text{estimation error}).
    $$
    Further details and the derivation for the error order can be found in App.~\ref{app:proof_richardson}.
    We examine this on the synthetic two-dimensional Eight-Gaussians ring distribution commonly used as an example benchmark in generative modeling \citep{izmailov_semi-supervised_2020}.
\todo{COMPACT 02: Move reference-bank and sweep details to appendix; inline the shape matrix.}
    Throughout the experiments presented in this section,
    the generalized Gaussian noise distribution is parameterized by the five-parameter tuple
    $\eta=(\beta,\lambda,\sigma,u,v)$ where $\Sigma=\sigma^2S(u,v)$, and $\sigma$ sets the noise magnitude. The noise shape ranges over symmetric positive-definite (SPD) matrices of the trace-normalized form
    $$
        S(u,v)=
        \begin{pmatrix}
            1+u & v\\
            v & 1-u
        \end{pmatrix}.
    $$
    We compare the score estimates from the Energy-Tweedie posterior route with ET-DSM and Hyv\"arinen score matching in Table~\ref{tab:score_full5_accuracy}.
    We report magnitude and direction accuracy over five parameter sweeps,
    using an importance-sampled ground truth from $160{,}000$ clean particles (App.~\ref{app:score_comparison}), reporting the NMSE (MSE over the
    mean squared ground-truth score norm) and the cosine similarity.
\todo{COMPACT 02: Shorten the following table caption's parameter ranges; retain metric and uncertainty definitions.}
    While the direct estimators attain lower errors, the posterior route provides accurate scores; for the former, we verified near-optimal validation losses
    (App. Fig.~\ref{fig:score_comparison_training}), while the posterior might be slightly undertrained, so the comparison is conservative.

    \begin{table*}[ht!]
        \centering
        \footnotesize
        \setlength{\tabcolsep}{2pt}
        \renewcommand{\arraystretch}{1.15}
        \newcommand{\scorecell}[2]{\begin{tabular}[t]{@{}c@{}}$#1$\\$\pm #2$\end{tabular}}
        \begin{minipage}[t]{0.49\textwidth}
            \centering
            \textbf{(a) Score NMSE $\downarrow$}\par\smallskip
            \begin{tabular}{@{}lccccc@{}}
                \toprule
                Estimator & $\beta$ & $\lambda$ & $\sigma$ & $u$ & $v$ \\
                \midrule
                ET posterior & \scorecell{0.050}{0.010} & \scorecell{0.039}{0.004} & \scorecell{0.077}{0.002} & \scorecell{0.037}{0.004} & \scorecell{0.029}{0.004} \\
                ET-DSM & \scorecell{0.019}{0.003} & \scorecell{0.014}{0.000} & \scorecell{0.017}{0.001} & \scorecell{0.018}{0.001} & \scorecell{0.015}{0.003} \\
                Hyv\"arinen & \scorecell{0.027}{0.002} & \scorecell{0.022}{0.003} & \scorecell{0.023}{0.001} & \scorecell{0.025}{0.002} & \scorecell{0.023}{0.003} \\
                \midrule
                GT disagr. & $0.0003$ & $0.0003$ & $0.0092$ & $0.0003$ & $0.0003$ \\
                \bottomrule
            \end{tabular}
        \end{minipage}\hfill
        \begin{minipage}[t]{0.49\textwidth}
            \centering
            \textbf{(b) Score cosine $\uparrow$}\par\smallskip
            \begin{tabular}{@{}lccccc@{}}
                \toprule
                Estimator & $\beta$ & $\lambda$ & $\sigma$ & $u$ & $v$ \\
                \midrule
                ET posterior & \scorecell{0.971}{0.008} & \scorecell{0.975}{0.000} & \scorecell{0.951}{0.000} & \scorecell{0.977}{0.004} & \scorecell{0.982}{0.006} \\
                ET-DSM & \scorecell{0.990}{0.000} & \scorecell{0.992}{0.001} & \scorecell{0.990}{0.003} & \scorecell{0.990}{0.000} & \scorecell{0.990}{0.002} \\
                Hyv\"arinen & \scorecell{0.987}{0.002} & \scorecell{0.989}{0.001} & \scorecell{0.987}{0.001} & \scorecell{0.987}{0.000} & \scorecell{0.987}{0.001} \\
                \bottomrule
            \end{tabular}
        \end{minipage}
        \caption{\textbf{Noisy-score accuracy across five parameter sweeps.} Each cell is the mean
        over two models, sample std below. Range of parameters:
        $\beta\in[1.3,2]$, $\lambda\in[1,2.5]$, $\sigma\in[0.1,0.75]$, and $u,v\in[-0.6,0.6]$.
        GT disagreement is the NMSE between ground-truth scores computed on separate halves of the clean particles.
        }
        \label{tab:score_full5_accuracy}
    \end{table*}
    Details, full comparisons and further results are presented in App.~\ref{app:score_comparison}.

    \subsection{Noise Parameter Estimation}
    \label{sec:param_estimation}

\todo{COMPACT 19: Keep general objective; state specialization in prose and move expanded argmin to appendix.}
    Equation~\ref{eq:thm_et} also allows us to use the mismatch between the two sides as a loss to learn unknown parameters of the noising distribution:
    \begin{equation}
        \EE_{y \sim
            m} \Big[ \Big\|\,
        \widehat{s}_m (y)\; +
            c\,\nabla_y^{\mathrm{PD}}\,S_\rho(P(X \mid Y=y),\,y)
        \Big\|_2^2\Big],
        \label{eq:param_optim_general}
    \end{equation}
\todo{COMPACT 34: Keep Hyvarinen example; omit sliced-score-matching example here.}
    where $\widehat{s}_m$ is another score estimator that \textit{does not require} knowledge of the noise parameters,
    for example the original \textit{score-matching estimator} in \citet{hyvarinen_estimation_2005} or \textit{sliced
    score matching} in \citet{song_sliced_2019}.
    Specializing again to the generalized Gaussian case, we obtain the objective:%
    \begin{equation}
        (\widehat{\beta}, \widehat{\lambda}, \widehat{\Sigma}) ~ \in ~ \argmin_{\beta, \lambda, \Sigma}\ \EE_{y \sim
            m} \Big[ \Big\|\, \widehat{s}_m (y)\; +
            \;\frac{\lambda}{\beta}\nabla^{\mathrm{PD}}_y\mathrm{ES}_{\Sigma^{-1},\beta}\big(P_\theta(\cdot\mid
            Y=y),y\big)\,\Big\|_2^2\Big].
        \label{eq:param_optim}
    \end{equation}
\todo{COMPACT 23: Inline the scale orbit and fold trace-normalization motivation; retain identifiability qualification.}
    In equation~\ref{eq:param_optim}, we encounter the well-studied scale invariance issue for elliptical
    distributions; specifically for the generalized Gaussian (Eq.~\ref{eq:gen_gaussian}), the following set of parameter choices leads to the same density:
    $$\{(\beta, c^{\beta/2} \, \lambda, c \, \Sigma) \mid c \in \R^+\}$$   
    After adopting a restriction --- in our case trace normalization $\Tr(\Sigma)=d$ --- this orbit has a unique representative, and $(\beta^*,\lambda^*,\Sigma^*)$ is the unique trace-normalized optimizer. The formal
    identifiability statement is Thm.~\ref{thm:identifiability} in Appendix~\ref{app:param}.
    Restricting the trace to the dimensionality of the space is a very common choice in similar settings~\citep{chen_regularized_2011}.
\todo{COMPACT 18: Move profiling displays to appendix; retain closed-form claim and pointer. Alternative: inline G only.}
    The scalar scale $\lambda$ admits a closed-form solution. Define:
    \begin{equation}
        G(y; \beta, \Sigma) = - \tfrac{1}{\beta}\nabla^{\mathrm{PD}}_y
        \mathrm{ES}_{\Sigma^{-1},\beta}\big(P(\cdot\mid Y=y),y\big) =
        \Sigma^{-1}\EE_{X \sim P(\cdot\mid Y=y)}[\|X - y\|^{\beta - 2}_{\Sigma^{-1}} (X - y)].
        \label{eq:G(y)}
    \end{equation}
    The unconstrained optimizer for $\lambda$ in Eq.~\ref{eq:param_optim} at the population level is then given by:
    \begin{equation}
        \lambda^*(\beta, \Sigma) =
        \frac{\EE_{Y \sim m}\left[\langle s_m(Y), G(Y; \beta, \Sigma) \rangle\right]}
        {\EE_{Y \sim m}\left[\|G(Y; \beta, \Sigma)\|^2_2\right]}.
        \label{eq:lambda_profile}
    \end{equation}
    The formal statement and Gaussian simplifications are given in Cor.~\ref{cor:lambda_profile}.
    We also show in Prop.~\ref{prop:plugin_consistency} that this noise-parameter estimator is consistent: under empirical $L_2$ consistency of the auxiliary
    score estimator $\widehat s_m$ and the Monte Carlo posterior field, approximate minimizers of the plug-in criterion converge to the normalized true parameter tuple $(\beta^*,\lambda^*,\Sigma^*)$.
    The implementation is summarized in Alg.~\ref{alg:noise_param} in Appendix~\ref{app:param}: parameterize a trace-normalized $\Sigma$, estimate $G$ from posterior samples, profile $\lambda$, and take gradient steps over the remaining  parameters.
    We test this profiled estimator in the same full five-parameter setting
    $(\beta,\lambda,\sigma,u,v)$ as in Sec.~\ref{sec:score_estimation}.
    One joint fit over three known $\sigma\in\{0.25,0.45,0.75\}$ levels estimates $(\beta,\lambda,u,v)$.
    Here, we write $\Sigma_j=\sigma_j^2S$ and normalize $\operatorname{tr}(S)=d$ as $\sigma_j$ are known factors.
    The results are presented in Table~\ref{tab:param_estimation_SPD}.
    For the score estimator, we use both an oracle importance-sampled and a learned Hyvärinen estimator, and likewise for the posterior, which is either importance-sampled or an Engression model. The first four rows isolate plug-in nuisance estimation within the correctly specified generalized family, while the final three isolate family misspecification while retaining \textit{oracle} nuisance components.
    \begin{table*}[ht!]
        \centering
        \setlength{\tabcolsep}{3pt}
        \resizebox{\textwidth}{!}{%
        \begin{tabular}{@{}llccccc@{}}
            \toprule
            Fitted family & \shortstack{Score/\\Posterior} & $\widehat\beta$ & $\widehat\lambda$
                & $\widehat u$ & $\widehat v$ & NMSE \\
            \midrule
            generalized & \shortstack{oracle/\\oracle} & $1.4094 \pm 0.0062$ & $1.7814 \pm 0.0041$
                & $0.2537 \pm 0.0043$ & $0.3505 \pm 0.0037$ & $0.0061 \pm 0.0004$ \\
            generalized & \shortstack{learned/\\oracle} & $1.4118 \pm 0.0075$ & $1.7733 \pm 0.0145$
                & $0.2493 \pm 0.0075$ & $0.3512 \pm 0.0059$ & $0.0061 \pm 0.0005$ \\
            generalized & \shortstack{oracle/\\learned} & $1.4379 \pm 0.0111$ & $1.7952 \pm 0.0098$
                & $0.2557 \pm 0.0036$ & $0.3463 \pm 0.0056$ & $0.0145 \pm 0.0018$ \\
            generalized & \shortstack{learned/\\learned} & $1.4398 \pm 0.0120$ & $1.7875 \pm 0.0150$
                & $0.2512 \pm 0.0061$ & $0.3472 \pm 0.0066$ & $0.0146 \pm 0.0017$ \\
            \shortstack[l]{forced Gaussian\\anisotropic} & \shortstack{oracle/\\oracle} & $\equiv 2$ & $1.4697 \pm 0.0210$
                & $0.2588 \pm 0.0087$ & $0.3466 \pm 0.0043$ & $0.0342 \pm 0.0016$ \\
            \shortstack[l]{forced isotropic\\generalized} & \shortstack{oracle/\\oracle} & $1.1206 \pm 0.0240$ & $1.8078 \pm 0.0154$
                & $\equiv 0$ & $\equiv 0$ & $0.1324 \pm 0.0025$ \\
            \shortstack[l]{forced isotropic\\Gaussian} & \shortstack{oracle/\\oracle} & $\equiv 2$ & $1.3650 \pm 0.0332$
                & $\equiv 0$ & $\equiv 0$ & $0.2081 \pm 0.0019$ \\
            \bottomrule
        \end{tabular}%
        }
        \caption{\textbf{Full-SPD noise-parameter estimation.} Ground truth: $(\beta^*,
    \lambda^*,u^*,v^*)=(1.4,1.8,0.25,0.35)$.  Estimated parameters and held-out score NMSE for the generalized family and three misspecified alternatives. NMSE is computed against the oracle noisy score, pooled over the three $\sigma$ levels.
        Standard deviations are over the five data replicates for the oracle posterior,
        and over the fifteen fits (3 trained posterior models $\times$ 5 replicates) when it is learned.
        }
        \label{tab:param_estimation_SPD}
    \end{table*}

    The fully learned generalized fit performs better than every misspecified fit even though the latter use oracle scores and posteriors: their error ranges from $2.3$ to $14.2$ times the fully learned generalized one.
    Forcing Gaussianity compensates for the imposed light tails by shifting $\widehat\lambda$ down, while imposing isotropy drags $\widehat\beta$ down to $1.12$, imitating the missing anisotropy with heavier tails.
    What provides accuracy is thus being able to fit the correct noise family, enabled by our approach, which ``survives'' plugging in an auxiliary score estimator almost untouched.
     Tail weight and anisotropic geometry are common in real noise, yet methods for estimating
     unknown noise parameters typically estimate only the scalar noise scale, fixing an isotropic Gaussian or a similar exponential-family model \citep{kim_noise2score_2021}.
    Such methods operate within the isotropic Gaussian case of
     Table~\ref{tab:param_estimation_SPD}; here that is fitted with oracle values against an M-estimation parameter, so no estimator confined to that class is likely to do better.
    The complete $16$-row comparison in App.~\ref{app:param} shows that the same family ordering persists across all four source combinations, confirming that the misspecification is due to the noise family restriction.

    \subsection{Energy-Tweedie Diffusion}
    \label{sec:diffusion}

    One can use Thm.~\ref{thm:et} to sample from $p(x)$ in a diffusion-like approach by using a time-parameterized model $P_\theta(X\mid Y,t)$.
    Specializing again to the generalized Gaussian case, the construction is:

\todo{COMPACT 06: Fold the following Train/Sample table into the training and sampling introductions.}
    \noindent\textbf{Noise-parameter path.} Choose $t\mapsto(\beta_t,\lambda_t,\Sigma_t)$, $t\in[0,1]$, and pass $t$ (or the corresponding noise parameters) as conditioning information.
    
    \smallskip
    \noindent
    \begin{tabular}{@{}p{.46\linewidth}@{\hspace{.05\linewidth}}p{.46\linewidth}@{}}
        \textbf{Train.} Fit $P_\theta(\cdot\mid Y,t)$ with a proper scoring rule, for example the matched
        non-Euclidean energy score in Eq.~\ref{eq:ES_diffusion_train}.
        &
        \textbf{Sample.} Draw $X^{(i)} \sim P_\theta(\cdot\mid Y=y_k,t_k)$ and use the Monte Carlo
        Energy-Tweedie score in Eq.~\ref{eq:ES_diffusion_sample} inside annealed Langevin or another score-based
        sampler.
    \end{tabular}
    The matched training objective is
    \begin{equation}
        \min_\theta\ \EE_{t \sim \mathrm{Unif}([0,1])}\,\EE_{x\sim p}\,\EE_{y\mid
            x,t}\Big[\,\mathrm{ES}_{\Sigma_t^{-1},\beta_t}\big(P_\theta(\cdot\mid Y=y,t),\,x\big)\Big],
        \label{eq:ES_diffusion_train}
    \end{equation}
    where $y=x+\epsilon_t,\ \epsilon_t\sim q_{\beta_t, \lambda_t, \Sigma_t}$.
    At reverse step $k$, the Energy-Tweedie score estimate is
    \begin{equation}
        \widehat{s}_{t_k}(y_k)
        =-\frac{\lambda_{t_k}}{\beta_{t_k}}\nabla_y^{\mathrm{PD}}
        \mathrm{ES}_{\Sigma_{t_k}^{-1},\beta_{t_k}}\big(P_\theta(\cdot\mid y_k,t_k),y_k\big)
        \approx-\frac{\lambda_{t_k}}{N}\Sigma_{t_k}^{-1}
        \sum_{i=1}^N
        \|y_k-X^{(i)}\|_{\Sigma_{t_k}^{-1}}^{\beta_{t_k}-2}(y_k-X^{(i)}).
        \label{eq:ES_diffusion_sample}
    \end{equation}
    Note that it is not strictly necessary that the estimator $P_{\theta}(\cdot \mid Y, t)$ was trained under the induced scoring rule  --- any form of distributional matching would suffice. However, it must have been trained over the range of
    $\left(\beta_t, \lambda_t, \Sigma_t\right)$ it is expected to see at sampling time.
    Despite this, matching the energy score to the noise-parameter path at training \textit{might} prove useful: for example, when $\beta\approx 1$, the energy score will encourage robustness to extreme values, which \textit{are} likely to come up at that step due to the noising distribution. Subsequent to the preprint versions of this work, \citet{martorella_improved_2026} modify the energy-score parameters of a scoring-rule diffusion model along the diffusion schedule while keeping isotropic Gaussian noise, and report improved few-step generation at ImageNet scale, in particular when only $\beta_t$ varies.%
\todo{NOTE: the De Bortoli positioning.}
    Plugging the Energy-Tweedie score into an annealed Langevin sampler thus yields a diffusion-style procedure over a \textit{noise-parameter path}.
    In the isotropic Gaussian case, this can be viewed through the usual reverse-SDE interpretation from ``ordinary diffusion'' \citep{song_score-based_2021} as the Energy-Tweedie identity (Eq.~\ref{eq:es_tweedie}) simplifies to the Tweedie identity; in the general case, it should be viewed as score-based MCMC over a sequence of intermediate targets as was the original perspective in NCSN/SMLD \citep{song_generative_2019}.
    The use of scoring rules for diffusion-like generative modeling has been explored in \textit{Reverse Markov Learning} \citep{shen_reverse_2025} and \textit{Distributional Diffusion Models with Scoring Rules} \citep{de_bortoli_distributional_2025}.
    In contrast to the procedure above, neither runs a score-based
    generation procedure; RML uses \textit{ancestral sampling} and is thus roughly analogous to DDPM, while \citet{de_bortoli_distributional_2025} employ an approach analogous to DDIM using the learned full posterior $p(x_0 \mid x_t)$. Our proposed approach thus provides the missing score-based perspective analogous to NCSN/SMLD.

\begin{figure}[ht!]
        \centering
        \includegraphics[width=\linewidth]{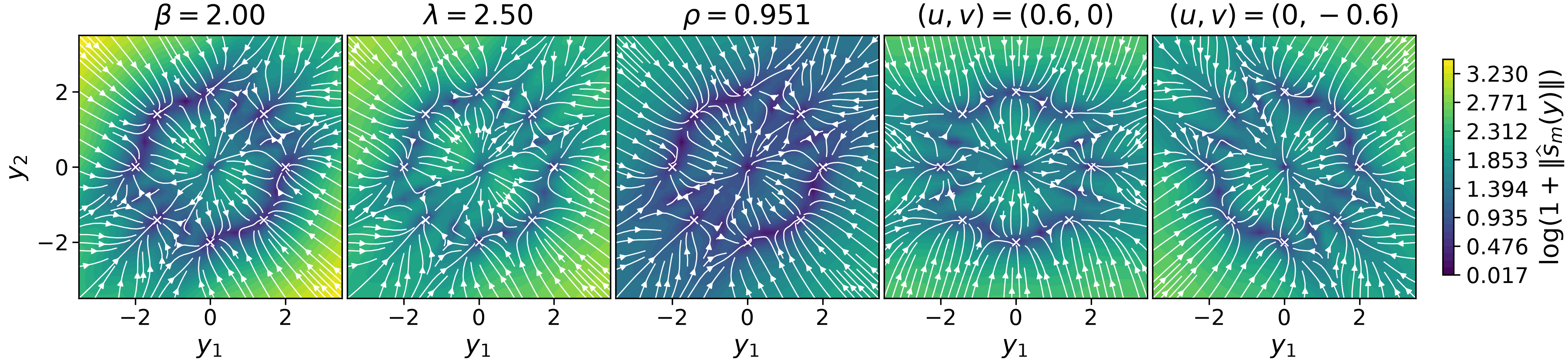}
        \caption{\textbf{Energy-Tweedie score fields} from the five-parameter Eight-Gaussians posterior model as each noise parameter is varied. The colorbar represents the score magnitude (log scale).}
        \label{fig:diffusion_score}
    \end{figure}

    We reuse the Eight-Gaussians posterior model of Sec.~\ref{sec:score_estimation} and train a corresponding one on MNIST \citep{lecun_gradient-based_1998}, both conditioned on $\eta=(\beta,\lambda,\sigma,u,v)$.
    Through Eq.~\ref{eq:ES_diffusion_sample}, each
    model supplies the Energy-Tweedie score anywhere in the noise-parameter space, and hence along an \emph{arbitrary} noise-parameter path $t\mapsto(\beta_t,\lambda_t,\Sigma_t)$  instead of a single scalar noise schedule.
    Figure~\ref{fig:diffusion_score} displays the learned score field for the Eight-Gaussians data as each
    parameter is varied in turn; in particular, the mean-seeking Gaussian field ($\beta=2$) concentrates mass toward  the ring, while for $\beta<2$ the field tends toward the geometric-median regime and targets the mixture centers directly, as discussed in Sec.~\ref{sec:elliptic_examples}. The displayed base parameter combination is $(\beta,\lambda,\rho,u,v)=(1.4,1.8,0.57,0,0.6)$ ($\rho$ is the reference RMS of Eq.~\ref{eq:noise_rms_scales}).
\todo{COMPACT 45: Consider reduced figure height and shorter caption only if labels and score-field detail remain legible.}
    For generation, we run annealed Langevin over four reverse paths through the noise-parameter space, matched in noise RMS: an isotropic fixed-tail reference, a fixed non-diagonal covariance, a covariance completing a full rotation, and a joint path along which all five parameters vary simultaneously. The path definitions, setup and more results are given in App.~\ref{app:diffusion_full5}.

\begin{figure}[ht!]
        \centering
        \includegraphics[width=\linewidth,trim=0 0 0 0.18in,clip]{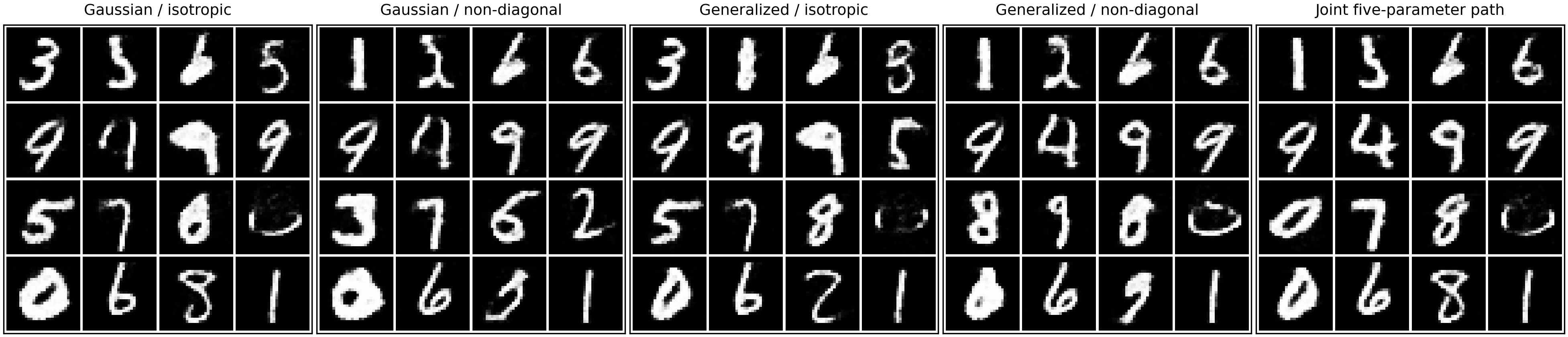}
        \caption{\textbf{Unselected MNIST samples} from the Gaussian/isotropic, Gaussian/non-diagonal,
            generalized/isotropic, generalized/non-diagonal, and joint paths (left to right).}
        \label{fig:mnist_samples}
    \end{figure}

\todo{COMPACT 20: Shorten path descriptions; retain shared schedule, model, budget, and reported findings.}
    The same construction carries over to MNIST. We examine five paths with a common noise magnitude: four fixed combinations of Gaussian or generalized radial law with
  isotropic or non-diagonal covariance, together with a jointly varying path that transforms Gaussian diffusion \textit{continuously} into a heavy-tailed anisotropic one (definitions  in App.~\ref{app:diffusion_mnist}).
    Figure~\ref{fig:mnist_samples} shows unselected samples from all five
    paths, each producing recognizable, class-diverse digits.
    On quantitative metrics (Table~\ref{tab:mnist_quantitative} in App.~\ref{app:diffusion_mnist}), the generalized/non-diagonal path attains the best point estimates on all metrics. As the model, noise magnitude schedule and sampling budget are shared, this comparison shows that the noise-parameter path, \textit{enabled by the Energy-Tweedie identity}, is a consequential modeling choice. Additional results are presented in App.~\ref{app:diffusion}.

    \section{Related Work}%
    \label{sec:related_work}

\todo{COMPACT 07: Compress repeated Tweedie/Fisher history and score-matching background; retain citations.}
    Classical work on Tweedie-type identities and empirical Bayes goes back to \citet{robbins_empirical_1956} and refinements such as \citet{efron_tweedies_2011}.
     Score matching \citep{hyvarinen_estimation_2005, song_sliced_2019} provides a way to estimate score fields and forms the basis of modern score-based generative modeling \citep{song_score-based_2021}.
    Self-supervised denoising handles unknown noise mainly either through risk surrogates or through Tweedie-based score extraction. SURE-type methods estimate or optimize denoising risk under Gaussian noise, with UNSURE relaxing the need to know the noise level \citep{stein_estimation_1981, tachella_unsure_2024}. Noise2Score and Noise-Distribution Adaptive Denoising instead estimate the noisy marginal score using denoising autoencoder machinery and plug it into classic Tweedie formulas  \citep{kim_noise2score_2021, kim_noise_2022}. The use of denoising autoencoders provides a contrast to recent work on energy-score-based autoencoders \citep{shen_distributional_2024}, for which  non-asymptotic score alignment has been established \citep{leban_distributional_2025}, albeit in a different setting. Estimation of unknown noise parameters has typically been indirect, relying on surrogates such as reconstructed-image regularity (cf. Sec.~\ref{sec:param_estimation}).
    In addition to the scoring-rule-based diffusion approaches discussed in Sec.~\ref{sec:diffusion}, heavy-tailed diffusion methods replace Gaussian perturbations with richer noising processes: \citet{yoon_score-based_2023} use isotropic $\alpha$-stable Lévy processes and fractional denoising score matching, \citet{shariatian_heavy-tailed_2025} give a DDPM-style construction with $\alpha$-stable noise, and \citet{pandey_heavy-tailed_2024} use Student-$t$ perturbations to model heavy-tailed data. These works modify the forward noising process or reverse sampler, whereas our contribution is a score identity. For Student-$t$ noise in particular, Cor.~\ref{cor:student_t} supplies the corresponding identity and matched kernel score.
    Concurrent work also links generative modeling to score-induced geometry: Energy Matching \citep{balcerak_energy_2025} uses an interaction term to ensure diversity (conceptually similar to the energy score), and an optimal-transport term to encourage direct paths to the data manifold, which we observe for $\beta \approx 1$; dual score matching \citep{guth_learning_2025} learns normalized image energies whose spatial gradient is the score; and the Information--Estimation Metric \citep{ohayon_learning_2025} builds data-dependent distances by comparing denoising-error or score fields over Gaussian noise scales. In our case the metric is tied to the additive noise law, and the identity singles out a ``mean $\beta$-energy'' (App.~\ref{app:connections_geom}).

    \section{Discussion}%
    \label{sec:discussion}

    This work generalizes the Gaussian Tweedie correspondence from a posterior-mean identity to a distributional identity that holds for any additive Gibbs noise. The main object is the denoising posterior viewed through a kernel score induced by the noise potential.
    The resulting Energy-Tweedie identity then gives the score of the noisy marginal as the path-derivative of this matched scoring rule, with the noise potential alone determining propriety. The classical results are recovered in the Gaussian case.
    The identity gives a posterior-sample route to score estimation, turns the estimation of unknown noise parameters (for any Gibbs noise) into an M-estimation problem that moves beyond estimating a scalar noise level, and provides the missing score-based view of ``distributional'' diffusion approaches, where the process can be non-Gaussian, and the
    noise-parameter schedule becomes a path chosen freely at sampling time.
     These consequences also suggest promising additional uses. In inverse problems \citep{daras_survey_2024}, many approaches use the score of noisy distributions, which Energy-Tweedie provides without the Gaussian assumption.
    Similarly, in test-time denoising under noise distribution shift, the discrepancy in Eq.~\ref{eq:param_optim_general} could be used to estimate candidate parameters.
    While the identity is general, we have focused on the generalized Gaussian family for the examples. The theory for parameter estimation currently assumes that the posterior is correctly specified and that the auxiliary score estimator is sufficiently accurate; relaxing these could be a future direction. On the diffusion side, scaling to competitive large-scale image-generation benchmarks remains open.

    \newpage

    \begin{ack}
    The author would like to dedicate this work to the memory of Pavla Vidmar (1938–2025).
    The author thanks the anonymous reviewers for their productive feedback, for example suggesting the inclusion of the Energy-Tweedie denoising score matching approach.
    This work received no funding. The author declares no competing interests.
    \end{ack}

    \bibliography{bibliography}

    \clearpage
    \appendix
    \section{Appendix}

    \begin{table}[ht!]
    \centering
    \renewcommand{\arraystretch}{1.1}
    \caption{\textbf{Core notation used throughout the paper.}}
    \begin{tabular}{@{}p{3.6cm} p{9.5cm}@{}}
    \toprule
    Symbol & Meaning \\
    \midrule
    $d$ & Ambient data dimension, with $X,\epsilon,Y \in \R^d$. \\
    $X,\ Y,\ \epsilon$ & Clean data, noisy observation, and additive noise, with $Y=X+\epsilon$. \\
    $Z,\ W,\ldots$ & Generic random variables used in distributional statements such as $Z\sim P$ and $W\sim Q$. \\
    $p(x),\ q(u),\ m(y)$ & Clean-data density, noise density, and noisy marginal density. \\
    $P(X\mid Y=y)$ & Denoising posterior distribution of $X$ given noisy observation $y$. \\
    $s_m(y)$ & Stein score of the noisy marginal: $s_m(y)=\nabla_y \log m(y)$. \\
    $\gE(u),\ c$ & Noise potential and scale parameter of the Gibbs noise density
        $q_{\gE,c}(u)\propto e^{-c\,\gE(u)}$. \\
    $\rho(x,y)$ & Discrepancy induced by the noise potential: $\rho(x,y)=\gE(y-x)$. \\
    $S_\rho(P,y)$ & Kernel score with kernel $\rho$, evaluated at predictive law $P$ and observation $y$. \\
    $D_\rho(P,Q)$ & Generalized energy distance between distributions $P$ and $Q$, induced by $\rho$. \\
    $\Sigma,\ \Sigma_0$ & Positive-definite noise shape matrices; $\Sigma=\varepsilon^2\Sigma_0$ in the small-noise
        analysis. \\
    $\beta,\ \lambda$ & Generalized-Gaussian shape and scale parameters. \\
    $\eta$ & Noise-parameter tuple; $\eta=(\beta,\lambda,\sigma,u,v)$ in the generalized-Gaussian experiments. \\
    $\mathrm{ES}_{\Sigma^{-1},\beta}(P,y)$ & Mahalanobis energy score with exponent $\beta$, evaluated at predictive law
        $P$ and observation $y$. \\
    $\nabla_y^{\mathrm{PD}}$ & Path derivative: differentiation with respect to $y$ while keeping the measure fixed. \\
    $P_\theta(X\mid Y=y)$ & Learned approximation to the denoising posterior. \\
    \bottomrule
    \end{tabular}
    \end{table}


    \subsection{Proofs of Section~\ref{sec:background}}
    \label{app:proof_background}

    \subsubsection{Proof of the Generalized Tweedie Formula}
    \label{app:proof_gauss}

    The Fisher identity is given in \citet{cappe_inference_2005} as:
    \begin{equation}
        \left.\nabla_\theta \ell(\theta)\right|_{\theta=\theta^{\prime}}=\left.\int \nabla_\theta \log f(x ;
            \theta)\right|_{\theta=\theta^{\prime}} p\left(x ; \theta^{\prime}\right) \lambda(d x),
        \label{eq:fisher_theta}
    \end{equation}
    with $f(x; \theta)$ being an unnormalized density, $L(\theta)$ the normalization constant:
    $$L(\theta) = \int f(x ; \theta) \lambda(d x),$$
    $\ell(\theta)$ the log-likelihood:
    $$\ell(\theta) =\log L(\theta),$$
    and $p(x ; \theta) = f(x ; \theta) / L(\theta).$

    As Eq.~\ref{eq:fisher_theta} is an analytic identity, we can perform the following symbolic substitution: $\theta
    \rightarrow y, \theta^{\prime} \rightarrow y^{\prime}$.

    Without restricting ourselves to location family noise, we can use a general noising kernel $k(y|x)$. In our
    notation, we have the following:
    $$f(x; y) = k(y\mid x)\, p(x) = p_{X, Y}(x, y), $$
    $$L(y)=\int k(y\mid x)\,p(x)\,dx = m(y), $$ and
    $$ p(x; y) = k(y\mid x)\, p(x) \,/\, m(y) = p_{X \mid Y=y}(x).$$

    Plugging this into the Fisher identity~\ref{eq:fisher_theta}, we have:
    \begin{equation*}
        \left.\nabla_y \ell(y)\right|_{y=y^{\prime}}=\left.\int \nabla_y \left[\log k(y|x) + \underset{\neq f(y)}{\cancel{\log
            p(x)}} \right]\right|_{y=y^{\prime}} \underbrace{\frac{ k(y^\prime \mid x)\, p(x) }{ m(y^\prime)}}_{p(x
            \mid Y=y^{\prime})} ~dx
    \end{equation*}
    Switching $y$ and $y^{\prime}$ around and noting $\nabla_y \ell(y) = s_m(y)$, we thus have:
    \begin{equation}
        s_m(y) \;=\; \EE_{X \sim P(\cdot \mid Y=y)} \left[\;\nabla_y \log k(y\mid X)\right].
    \end{equation}
    For the location family, we have:
    $$k(y \mid X) = q(y - X), $$
    thus we recover Eq.~\ref{eq:general_tweedie}:
    \begin{equation}
        s_m(y) \;=\; \EE_{X \sim P(\cdot \mid Y=y)} \left[\;\nabla_y \log q(y - X)\right].
    \end{equation}

    To recover the ``classic'' Tweedie's formula, we plug in a multivariate Gaussian:
    $$k(y|x) = q(y - x) \propto \exp\left[-\frac{1}{2} (y - x)^\top \Sigma^{-1} (y - x)\right], $$
    we obtain:
    $$
    s_m(y) \;=\; \EE_{X \sim P(\cdot \mid Y=y)} \left[\;\nabla_y \left(-\frac{1}{2} (y - X)^\top
        \Sigma^{-1} (y - X) \right) \right]
    =\; \EE_{X \sim P(\cdot \mid Y=y)} \left[ - \Sigma^{-1} (y - X)  \right]$$
    $$
    = \Sigma^{-1} \EE \left[X \mid Y=y \right] - \Sigma^{-1}  y,
    $$
    hence:
    $$
    \EE \left[X \mid Y=y \right]  = y + \Sigma \, s_m(y).
    $$
    \qed

    \subsection{Proofs of Section~\ref{sec:gibbs}}
    \label{app:proof_gibbs}

    \subsubsection{Proof of Proposition~\ref{prop:gibbs_tweedie}}
    \label{app:proof_gibbs_tweedie}
    Substituting Eq.~\ref{eq:gradE} into Eq.~\ref{eq:general_tweedie} (Assumption (A2)) gives
    the first equality; by (A3) the substitution is valid $P(\cdot \mid Y=y)$-almost surely,
    and (A4) makes the expectation well defined. The second equality is the definition of the
    path derivative (Eq.~\ref{eq:path_derivative}) applied to $f(y,X)=\gE(y-X)$.
    \qed

    \subsubsection{Derivations for the Examples}
    \label{app:examples_derivations}

    The examples are special cases of $q(u)\propto \exp\{-V(\|u\|_{\Sigma^{-1}})\}$ with an \textit{elliptical} potential $V$.
    For such, Eq.~\ref{eq:gradE} becomes:
    \begin{equation}
    \nabla_y \log q(y-X) \;=\; -\,\nabla_y V\!\left(\|y-X\|_{\Sigma^{-1}}\right) =
        -\,\frac{V'(\|y-X\|_{\Sigma^{-1}})}{\|y-X\|_{\Sigma^{-1}}}\,\Sigma^{-1}(y-X),
    \end{equation}

    By the definition of the path derivative (Eq.~\ref{eq:path_derivative}), this is:
    \begin{equation}
        s_m(y) = -\,\EE\!\left[\nabla_y V\!\left(\|y-X\|_{\Sigma^{-1}}\right)\,\middle|\,Y=y\right]
        = -\,\nabla^{\mathrm{PD}}_y\,\EE\!\left[V\!\left(\|y-X\|_{\Sigma^{-1}}\right)\,\middle|\,Y=y\right],
        \label{eq:elliptical_tweedie}
    \end{equation}

    For Gaussian noise with $V(r)=r^2/2$ and $\Sigma=\sigma^2$ in one dimension, Eq.~\ref{eq:elliptical_tweedie} gives
    \begin{align*}
    s_m(y)&=-\nabla_y^{\mathrm{PD}}~ \EE\left[ \frac{(y-X)^2}{2 \sigma^2} \mid Y=y\right] \\
    &= -\nabla_y^{\mathrm{PD}} \left( \frac{y^2}{2\sigma^2} - \frac{2y}{2\sigma^2}\EE[X \mid Y=y] +
        \frac{1}{2\sigma^2} \EE[X^2 \mid Y=y]\right),
    \end{align*}
    hence
    $$
    s_m(y) = \frac{\EE[X\mid Y=y] - y } { \sigma^2 }.
    $$

    In the multivariate Gaussian case, $V(r)=\frac{1}{2}r^2$, and
    \begin{align*}
        s_m(y)
        &= -\nabla_y^{\mathrm{PD}}~ \EE\left[ \frac{1}{2}\| y - X \|_{\Sigma^{-1}}^2 \mid Y=y\right] \\
        &= -\nabla_y^{\mathrm{PD}} \left( \frac{1}{2} y^\top \Sigma^{-1} y - y^\top \Sigma^{-1} \EE[X \mid Y=y] +
            \frac{1}{2} \EE[X^\top \Sigma^{-1} X \mid Y=y] \right),
    \end{align*}
    so
    $$
    s_m(y) = \Sigma^{-1} \left( \EE[X\mid Y=y] - y \right).
    $$

    For multivariate Laplace noise, $q(u)\propto \exp(-\lambda \|u\|_{\Sigma^{-1}})$, so
    $V(r)=\lambda r$. Eq.~\ref{eq:elliptical_tweedie} yields
    $$
    s_m(y) \;=\; -\,\lambda\,\nabla^{\mathrm{PD}}_y \EE\!\left[ \, \|y - X\|_{\Sigma^{-1}}\,\middle|\,Y=y\right]
    ~=\;-\lambda\,\EE\!\left[\frac{\Sigma^{-1}(y-X)}{\|y-X\|_{\Sigma^{-1}}}\;\middle|\;Y=y\right].
    $$
    The vector $\Sigma^{-1}(y-X)/\|y-X\|_{\Sigma^{-1}}$ records direction in the $\Sigma^{-1}$ geometry while discarding
    magnitude, giving the median-seeking interpretation of Sec.~\ref{sec:elliptic_examples}.

    \subsubsection{Proof of Lemma~\ref{lem:gen_gaussian_score}}
    \label{app:proof_gen_gaussian_score}

    First, (A1) holds: whitening ($u=\Sigma^{1/2}v$) and polar coordinates reduce the normalizing
    constant to a Gamma integral:
    \begin{align*}
        Z_\gE(c) \;&=\; |\Sigma|^{1/2}\,\frac{2\pi^{d/2}}{\Gamma(d/2)}
        \int_0^\infty r^{d-1}\, e^{-c\,r^{\beta}}\,dr \\
        &=\; |\Sigma|^{1/2}\,\frac{2\pi^{d/2}}{\Gamma(d/2)}\,
        \frac{\Gamma(d/\beta)}{\beta}\; c^{-d/\beta}
        \;<\;\infty \qquad \text{for all } c>0,
    \end{align*}
    so $\Lambda_\gE=(0,\infty)$ as stated.
    By the Tweedie identity for the elliptical distributions (Eq.~\ref{eq:elliptical_tweedie}), we have:
    $$
    s_m(y) = - \nabla_y^{\mathrm{PD}} ~ \EE\left[ \frac{\lambda}{\beta}\| y - X \|_{\Sigma^{-1}}^\beta \mid Y=y\right] =
    \EE\left[ - \lambda \, \|y - X\|_{\Sigma^{-1}}^{\beta - 2} ~\Sigma^{-1} \, (y - X) \mid Y=y \right],
    $$
    hence:
    $$
    s_m(y) = -\lambda ~ \Sigma^{-1} \, \EE\left[  \|y - X\|_{\Sigma^{-1}}^{\beta - 2} ~ (y - X) \mid Y=y \right].
    $$
    \qed

    \textbf{Integrability:} We verify integrability condition (A4) of Prop.~\ref{prop:gibbs_tweedie}; we note that as the field
    behaves like $\|y-X\|^{\beta-1}$, we observe three regimes:
    \begin{itemize}
        \item $\beta>1$: no singularity at the diagonal $X = y$; the field goes to $0$.
        \item $\beta=1$: bounded off the diagonal; the only issue is pointwise non-differentiability at $X=y$, which is handled by (A3).
        \item $0<\beta<1$: singular at the diagonal, the question is whether that singularity is locally integrable
            under the posterior.
    \end{itemize}
    For the last case: at the single point $X=y$, the vector field may be undefined; throughout, it may be assigned an arbitrary value there,
    since under the posterior it has probability zero.
    Next, \textit{if} the posterior density $p(x\mid Y=y)$ is bounded,
    then for $\beta \in (0, 1]$, in the $\varepsilon$-neighborhood of the diagonal for any $y$ the expectation is locally
    bounded by the radial component of the integral:
    $$
    \int_{\|x-y\|\le \varepsilon}\|x-y\|^{\beta-1}\,p(x\mid Y=y)\,dx \;\le\; C\int_0^\varepsilon r^{d+\beta-2}\,dr,
    $$
    where $d \geq 1$ is the dimension of the ambient space. Furthermore, as for $y \approx x$, $p(x\mid Y=y) \propto q(y - x)
    p(x)$, this means that the identity is well-defined for $\beta \in (0, 1]$ \textbf{if the data distribution
    has a locally bounded density at $y$} as $q$ is continuous and finite at 0.

    \subsection{Proofs of Section~\ref{sec:ET}}
    \label{app:proof_ET}

    \subsubsection{Proof of Proposition~\ref{prop:matched_propriety}}
    \label{app:proof_matched_propriety}

    The claims follow from existing results:
    \begin{itemize}
        \item CND $\Rightarrow$ proper is given by Theorem~4 in \citet{gneiting_strictly_2007}.
        \item Proper $\Rightarrow$ CND: this is stated in Sect.~5.2 of \citet{gneiting_strictly_2007}, citing Theorem~1 in \citet{szekely_new_2005}.
        \item Strictly proper $\Leftrightarrow$ strong negative type: combining Eqs.~(29) and~(31) in
            \citet{gneiting_strictly_2007}, the expected score gap under the true law $Q$ is, in our negative
            orientation:
            $$\EE_{W \sim Q}[S_\rho(P, W)] - \EE_{W \sim Q}[S_\rho(Q, W)] = \tfrac12\, D_\rho(P, Q).$$
            The gap is strictly positive for all $P \neq Q$ exactly when $D_\rho(P, Q) > 0$ for all $P \neq Q$,
            which is the definition of strong negative type (Def.~28 in \citet{sejdinovic_equivalence_2013}).
    \end{itemize}

    It remains to show that for CND $\rho$, the finite $\rho$-moment condition of Thm.~\ref{thm:et}
    holds automatically at the denoising posterior. First, the first term of the kernel score is finite for
    \textit{any} measurable noise potential: since $\gE \geq 0$ and $t\,e^{-c t} \leq \tfrac{1}{c e}$ for all
    $t \geq 0$,
    $$
    \begin{aligned}
    \EE\left[\rho(X, y) \mid Y=y\right]
    &= \frac{1}{m(y)}\int \gE(y-x)\, q_{\gE,c}(y-x)\, p(x)\, dx \\
    &= \frac{1}{Z_\gE(c)\,m(y)}
        \int \gE(y-x)\,e^{-c\gE(y-x)}\,p(x)\,dx \\
    &\leq \frac{1}{c\,e\,Z_\gE(c)\,m(y)}\int p(x)\,dx \\
    &= \frac{1}{c\,e\,Z_\gE(c)\,m(y)} < \infty.
    \end{aligned}
    $$

    Next, a CND $\rho$ with $\rho(z,z)=0$ admits a Hilbert-space representation
    $\rho(z, z') = \|\varphi(z) - \varphi(z')\|_{\mathcal{H}}^2$ (Prop.~3 in \citet{sejdinovic_equivalence_2013}),
    so $\rho^{1/2}$ obeys the triangle inequality and
    $\rho(X, X') \leq 2\,\rho(X, y) + 2\,\rho(y, X')$ for posterior draws $X,X'$.
    Taking expectations over $X, X' \iid P(\cdot \mid Y=y)$ and using the above:
    $$
    \EE\left[\rho(X, X') \mid Y=y\right] \leq 4\, \EE\left[\rho(X, y) \mid Y=y\right] < \infty.
    $$

    \qed

    \subsubsection{Proofs of Cors.~\ref{cor:gen_gaussian} and~\ref{cor:student_t}}
    \label{app:proof_gen_gaussian}

    \paragraph{Proof of the Mahalanobis energy score propriety}

    As the Mahalanobis distance $\|\cdot\|_{\Sigma^{-1}}$ appears in both corollaries,
    we wish to show that the $\beta$-th power of the \textit{Mahalanobis distance}
    $$
    d_{\Sigma^{-1}}(z_i, z_j)=\|z_i - z_j\|_{\Sigma^{-1}} = \sqrt{(z_i- z_j)^\top \Sigma^{-1} (z_i- z_j)}
    $$
    is a conditionally negative definite (CND) kernel, as well as develop some general machinery.
     We assume $\Sigma$ to be a symmetric positive definite (PD) matrix $\Sigma \succ 0$, so $\Sigma^{-1}$ exists and is symmetric PD.

    A real-valued function $g$ on $\R^d \times \R^d$ is said to be a \textit{conditionally negative definite (CND) kernel} if it is (i)
        symmetric in its arguments and (ii) $$\sum_{i=1}^n \sum_{j=1}^n a_i a_j g\left(z_i, z_j\right) \leq 0$$ for all
        positive integers $n$, all $z_1, \ldots, z_n$, and all $a_1, \ldots, a_n \in \mathbb{R}$ that sum to $0$ (this
        is the ``conditionally'' part).

    To show symmetry, we can simply write:
    \begin{align*}
        g(z_j, z_i) &= \| z_j - z_i \|^\beta_{\Sigma^{-1}} = \left( \langle \Sigma^{-1/2} (z_j - z_i), \Sigma^{-1/2} (z_j
            - z_i) \rangle  \right)^{\tfrac \beta 2} \\
        &= \left( \langle \Sigma^{-1/2} z_j, \Sigma^{-1/2}z_j \rangle - 2 \langle \Sigma^{-1/2} z_j, \Sigma^{-1/2} z_i
            \rangle +  \langle \Sigma^{-1/2}z_i, \Sigma^{-1/2}z_i \rangle   \right)^{\tfrac \beta 2} \\
        &= \| z_i - z_j \|^\beta_{\Sigma^{-1}} = g(z_i, z_j)
    \end{align*}
        \paragraph{$\beta=2$ case:}
    For the second requirement, let's begin with the $\beta=2$ case:

    \begin{align*}
        \sum_{i=1}^n \sum_{j=1}^n a_i a_j ~ g(z_j, z_i) =
        \sum_{i=1}^n a_i  \sum_{j=1}^n a_j \big( \|\Sigma^{-1/2}z_i\|^2 - 2 \langle \Sigma^{-1/2}z_j, \Sigma^{-1/2}z_i
            \rangle + \|\Sigma^{-1/2} z_j\|^2 \big)=\\
        \underbrace{\sum_{i=1}^n a_i  \sum_{j=1}^n a_j \big( \|\Sigma^{-1/2}  z_i\|^2 + \|\Sigma^{-1/2}z_j\|^2
            \big)}_{A} -
        \underbrace{2 \sum_{i=1}^n a_i  \sum_{j=1}^n a_j z_j^\top \Sigma^{-1} z_i}_{B}
        \stackrel{?}{\leq} 0
    \end{align*}

    For the terms in A, we notice that:
    $$
    \sum_{i,j} a_i a_j\, z_i^\top \Sigma^{-1}  z_i = \sum_i a_i\, z_i^\top \Sigma^{-1}   z_i \;\underbrace{\sum_j
        a_j}_{=\,0}=0,
    $$
    and vice-versa for the $z_j$ norm term.

    Thus, we are left with the term in B:
    \begin{align*}
        - 2 \sum_{i=1}^n a_i  \sum_{j=1}^n a_j \langle  \Sigma^{-1/2} z_j, \Sigma^{-1/2} z_i \rangle =
        - 2 (\sum_{i=1}^n a_i z_i)^\top \Sigma^{-1} (\sum_{j=1}^n a_j z_j) \leq 0,
    \end{align*}
    Since $\Sigma^{-1}$ is PD, the above quadratic form is nonnegative and the expression is $\leq 0$ after multiplication by -2.
    Together with Theorem~4 in \citet{gneiting_strictly_2007}, this gives propriety of the scoring rule
    (Eq.~\ref{eq:non-eucl-ES}) at  $\beta=2$.

    \qed

    \paragraph{General case --- $\beta \in (0, 2)$:}

    First of all, we note that the Mahalanobis norm is \textit{not} a CND kernel for $\beta > 2$:

    Let $(a_1,a_2,a_3)=(1,1,-2)$ and
    $z_1=z$, $z_2=w$, $z_3=\bar z\Def\tfrac{z+w}{2}$, so $\sum_i a_i=0$. Since $\bar z$ is the midpoint of $z$ and
    $w$, the three nonzero pairwise distances are
    $$
    \|z_1-z_2\|_{\Sigma^{-1}}=\|z-w\|_{\Sigma^{-1}},
    \qquad
    \|z_1-z_3\|_{\Sigma^{-1}}=\|z_2-z_3\|_{\Sigma^{-1}}
    =\tfrac12\|z-w\|_{\Sigma^{-1}}.
    $$
    The diagonal terms vanish, and by symmetry each $(i,j)$ term  can be paired with the corresponding $(j,i)$ term. Therefore,
    \begin{align*}
        \sum_{i=1}^3 \sum_{j=1}^3 a_i a_j\,\|z_i-z_j\|_{\Sigma^{-1}}^\beta
        &=2\left[
            a_1a_2\|z_1-z_2\|_{\Sigma^{-1}}^\beta
            +a_1a_3\|z_1-z_3\|_{\Sigma^{-1}}^\beta
            +a_2a_3\|z_2-z_3\|_{\Sigma^{-1}}^\beta
        \right] \\
        &=2\left[
            \|z-w\|_{\Sigma^{-1}}^\beta
            -2\left(\tfrac12\|z-w\|_{\Sigma^{-1}}\right)^\beta
            -2\left(\tfrac12\|z-w\|_{\Sigma^{-1}}\right)^\beta
        \right] \\
        &=2\|z-w\|_{\Sigma^{-1}}^\beta\left(1-2^{2-\beta}\right).
    \end{align*}
    The CND inequality would require $1-2^{2-\beta}\leq0$, equivalently $\beta\leq2$. Thus, for $\beta>2$, the sum is strictly positive whenever $z\neq w$, which violates the CND inequality for every $\Sigma\succ0$.

    \qed

    Now, we tackle the general case for $\beta \in (0, 2)$:
    For this, we reproduce a classical result due to \citet{schoenberg_metric_1938}, using a more modern reference found
    in \citet{berg_stieltjes-pick-bernstein-schoenberg_2008}.

    We will be using the concept of a Bernstein function, which is a $\gC^\infty$ function whose derivative is
    completely monotone. Such functions $\phi$ admit the \textit{Levy-Khinchin} representation (Def.~5.1 in
    \citep{berg_stieltjes-pick-bernstein-schoenberg_2008}):
    $$
    \phi(t)\;=\;a + b\,t + \int_0^\infty \big(1-e^{-t s}\big)\,\mu(ds),
    $$
    for some $a, b\ge 0$ and a positive measure $\mu$ on $(0,\infty)$.

    Further, note that $\phi(t)=t^\alpha$ is Bernstein $\forall \alpha \in (0, 1]$ (Example on p. 16 in
    \citep{berg_stieltjes-pick-bernstein-schoenberg_2008}).

    Next, we claim that the composition of a Bernstein function and a CND kernel is a CND kernel.

    To see this, use the Levy-Khinchin representation. Denote for the CND kernel $g(z_i, z_j) \Def g_{ij}$. By the
    Levy-Khinchin representation:
    \begin{equation}
        \sum_{i,j} a_i a_j ~ \phi \circ g_{ij} = \sum_{i,j} a_i a_j\, \phi\big(g_{ij}\big) = a \sum_{i,j} a_i a_j + b
            \sum_{i,j} a_i a_j\, g_{ij}  + \int_0^\infty \sum_{i,j} a_i a_j\big(1-e^{-s g_{ij}}\big)\,\mu(ds).
        \label{eq:levy-khinchin}
    \end{equation}

    Next, we use Schoenberg's theorem (Thm.~2.2 in \citep{berg_stieltjes-pick-bernstein-schoenberg_2008}):
    \begin{quote}
        \textit{A function $g$ is CND $\Leftrightarrow$ $\exp (-s g)$ is positive definite for all $s > 0$. }
    \end{quote}

    So, denoting $e^{-s g_{ij}}$ as the PD kernel $K_s(z_i, z_j)$, we have for the terms in
    Eq.~\ref{eq:levy-khinchin}:
    \begin{itemize}
        \item $a \sum_{i,j} a_i a_j = a \underbrace{\sum_i a_i}_{ = 0 } \underbrace{\sum_j a_j}_{ = 0 } \leq 0$
        \item $b\sum_{i,j} a_i a_j g_{ij} \leq 0$ since $b \geq 0$ and $g$ is CND.
        \item $ \sum_{i,j} a_i a_j \big(1-K_s(z_i,z_j)\big)
        = \underbrace{\Big(\sum_i a_i\Big)^2}_{=\,0}\;-\;\sum_{i,j} a_i a_j K_s(z_i,z_j) \;\le\; 0.    $
    \end{itemize}
    Integration of the nonpositive term above ($\forall s$) with respect to the positive measure $\mu$ is thus nonpositive.
    We thus have that a composition of a Bernstein function and a CND kernel is a CND kernel.

    \qed

    We will first use this to prove that the Mahalanobis norm is a CND kernel for $\beta \in (0, 2)$, and thus the corresponding energy score is a proper scoring rule.

    Let $\alpha=\beta/2\in(0,1)$, and denote $k_\beta = (k_2)^\alpha$, and
    $$
    k_2(z,w) \;=\; \|Lz-Lw\|_2^2  = \| z - w\|_{\Sigma^{-1}}^2,
    $$
    where we are using the Cholesky decomposition $\Sigma^{-1} = L^\top L$.

    Using Schoenberg's theorem (see above) again, $k_2$ is CND as the corresponding factor $e^{-s \| u -
    u^\prime\|_2^2}$ is the well-known PD Gaussian kernel for $u = L z$ and $u^\prime=Lw$.

    Thus we have $\| z - w\|_{\Sigma^{-1}}^\beta = k_\beta = t^\alpha \circ k_2$ is CND by the composition rule for
    $\beta \in (0, 2)$.
    Then, following Theorem~4 in \citet{gneiting_strictly_2007}, the corresponding scoring rule
    (Eq.~\ref{eq:non-eucl-ES}) is proper.

    \qed

    For \textit{strict} propriety, we reference Theorem~5 in \citet{gneiting_strictly_2007}. It requires the scoring rule to be expressible as:
    $$
    S(P, w)=\EE_{Z \sim P} [\psi\left(\| Z - w\|_2^2\right)] - \tfrac1 2 \EE_{Z, Z^\prime \iid P} [\psi\left(\| Z -
        Z^\prime\|_2^2\right)]
    $$
    for some $\psi$ continuous on $[0, \infty)$ and $\psi'$ completely monotone and not constant.

    We have, for $T(z)=L z$:
    $$
    \|z-w\|_{\Sigma^{-1}}^\beta=\|T(z)-T(w)\|_2^\beta=\psi\left(\|T(z)-T(w)\|_2^2\right), \quad \psi(t)=t^{\beta / 2}
    $$
    $\psi(t)=t^{\beta / 2}$ is the example given for Theorem~5 in \citet{gneiting_strictly_2007}.
    Since $T(z)=Lz$ is invertible, equality of the whitened distributions is equivalent to equality of the original ones,  so strictness transfers to the originals.
    Hence, the Mahalanobis energy score is, much like the ordinary energy score, strictly proper for $\beta \in (0, 2)$.

    \qed

    \paragraph{$\beta=2$ case for the generalized Gaussian}

    We show the collapse of the generalized energy distance $D_\rho$ as stated in Cor.~\ref{cor:gen_gaussian}.
    At $\beta=2$ the induced discrepancy is $\rho(z,w)=\|z-w\|_{\Sigma^{-1}}^2$; $P$, $Q$ have finite second moments throughout (the finite $\rho$-moment class at $\beta=2$).

    We use two standard facts. A positive definite reproducing kernel Hilbert space (RKHS) kernel $k$ is said to \textit{generate} the discrepancy $\rho$ if
    $$
        \rho(z,w)=k(z,z)+k(w,w)-2\,k(z,w).
    $$
    First, Theorem~22 in \citet{sejdinovic_equivalence_2013} states that for any generating RKHS kernel, the generalized energy distance equals twice the squared \textit{maximum mean
    discrepancy} (MMD):
    $$D_\rho(P,Q)=2\,\mathrm{MMD}_k^2(P,Q).$$
    Second, the MMD is the distance between the \textit{kernel mean embeddings} $\mu_P, \mu_Q$ of the two
    distributions in the RKHS $\mathcal H_k$ (Lemma~4 in \citet{gretton_kernel_2012}):
    $$\mathrm{MMD}_k(P,Q)=\|\mu_P-\mu_Q\|_{\mathcal H_k}, $$
    where a kernel mean embedding is defined as $\mu_P=\EE_{Z\sim P}\left[k(Z,\cdot)\right].$

    The squared Mahalanobis distance is generated by the \textit{linear} kernel $k(z,w)=z^\top\Sigma^{-1}w$:
    $$
    k(z,z)+k(w,w)-2\,k(z,w) = z^\top\Sigma^{-1}z + w^\top\Sigma^{-1}w - 2\,z^\top\Sigma^{-1}w
    = \|z-w\|_{\Sigma^{-1}}^2.
    $$
    Its mean embedding is the ordinary mean, $\mu_P=\EE_{Z\sim P}[Z]$, and the RKHS norm reduces to
    $\|\cdot\|_{\Sigma^{-1}}$. Combining the two facts, the discrepancy collapses to a difference of means:
    $$
    D_\rho(P,Q) = 2\,\bigl\|\EE_{Z\sim P}[Z]-\EE_{W\sim Q}[W]\bigr\|_{\Sigma^{-1}}^2.
    $$

    \qed

    \paragraph{Proof of Student-t propriety}
    \label{app:proof_student_t}

    For the Student-t potential $\gE(u)=\log\bigl(1+\|u\|_{\Sigma^{-1}}^2\bigr)$, the noise density is
    $$q_{\gE, c}(u)\,=\frac{1}{Z_\gE  (c)}\,e^{-c\,\gE(u)}\,=\, \frac{1}{Z_\gE  (c)} \,(1+\|u\|_{\Sigma^{-1}}^2)^{-c}$$
    for $c \in \Lambda_\gE=(d/2,\infty)$. Setting $u = y - X$, the matched kernel score is:
    \begin{equation}
        S_\rho(P,y)\;=\;\mathbb E_{X\sim P}\!\left[\log\bigl(1+\|y - X\|_{\Sigma^{-1}}^2\bigr)\right]\;
        -\;\tfrac12\,\mathbb E_{X,X' \iid P}\!\left[\log\bigl(1+\|X-X'\|_{\Sigma^{-1}}^2\bigr)\right].
        \label{eq:student_t_kernel}
    \end{equation}
    For propriety, we follow the same approach as above with the composition rule: with $k_2(z,w)=\|z-w\|_{\Sigma^{-1}}^2$ CND as above,
    we have:
    $$
    \rho(z, w) \;=\; \log\bigl(1+\|z-w\|_{\Sigma^{-1}}^2\bigr) \;=\; \phi \circ k_2, \quad \phi(t)=\log(1+t).
    $$
    $\phi$ is Bernstein: it is smooth and nonnegative on $[0, \infty)$, and $\phi'(t)=(1+t)^{-1}$ is completely
    monotone, since $(-1)^n \phi^{(n+1)}(t)=n!\,(1+t)^{-(n+1)} \geq 0$. Thus $\rho$ is CND by the composition
    rule, and following Theorem~4 in \citet{gneiting_strictly_2007}, the matched kernel score
    (Eq.~\ref{eq:student_t_kernel}) is proper.

    \qed

    For \textit{strict} propriety, we again apply Theorem~5 in \citet{gneiting_strictly_2007} exactly as for the Mahalanobis energy score. For $T(z)=Lz$:
    $$
    \log\bigl(1+\|z-w\|_{\Sigma^{-1}}^2\bigr)=\psi\left(\|T(z)-T(w)\|_2^2\right), \quad \psi(t)=\log(1+t),
    $$
    with $\psi$ continuous on $[0, \infty)$ and $\psi'$ completely monotone and not constant, as verified above;
    unlike for $\psi(t)=t^{\beta/2}$, no parameter restriction arises.

    \paragraph{Strict propriety moment restriction for Student-t}

\todo{modified: introduced the explicit Student-t moment condition.}
    In the Student-t case, the general pairwise moment requirement in Prop.~\ref{prop:matched_propriety} reduces to the explicit logarithmic moment condition stated in Cor.~\ref{cor:student_t}.

    Using $\rho(z,w)=\psi\left(\|T(z)-T(w)\|_2^2\right)$ as above, Theorem~5 in \citet{gneiting_strictly_2007} gives strict propriety for
    distributions $P$ with finite pairwise moment $\EE_{Z,Z'\iid P}\left[\rho(Z,Z')\right]<\infty$.
    Since   $\rho(z,z')=\gE(z-z')$, this pairwise condition is equivalent to
    $\EE_{Z, Z' \iid P}\left[\gE(Z - Z')\right]<\infty$. We show below that this is equivalent to the stated condition $\EE_{Z \sim P}\left[\gE(Z)\right]<\infty$.

    For any $u,v\in\R^d$, the triangle inequality and $(a+b)^2\leq 2a^2+2b^2$ give
    $1+\|u+v\|_{\Sigma^{-1}}^2\leq 4\left(1+\|u\|_{\Sigma^{-1}}^2\right)\left(1+\|v\|_{\Sigma^{-1}}^2\right)$, hence
    $$
        \gE(u+v)\leq \gE(u)+\gE(v)+2\log 2.
    $$
    Taking $u=z$, $v=-z'$ (with $\gE$ even) gives $\rho(z,z')\leq\gE(z)+\gE(z')+2\log 2$, so
    $\EE_{Z\sim P}\left[\gE(Z)\right]<\infty$ implies $\EE_{Z,Z'\iid P}\left[\rho(Z,Z')\right]<\infty$.
    On the other hand, taking $u=z-z'$, $v=z'$ gives
    $$\gE(z)\leq\rho(z,z')+\gE(z')+2\log 2.$$

    For the converse direction, suppose $\EE_{Z,Z'\iid P}[\rho(Z,Z')]<\infty$.
    By Fubini’s theorem, there exists a
    fixed $z_0\in\R^d$ such that
    $$
    \EE_{Z\sim P}[\rho(Z,z_0)]
    =\EE_{Z\sim P}[\gE(Z-z_0)]<\infty.
    $$
    Applying the preceding inequality to $Z=(Z-z_0)+z_0$ gives
    $$
    \EE[\gE(Z)]
    \leq \EE[\gE(Z-z_0)]+\gE(z_0)+2\log 2<\infty.
    $$

    Thus $\EE_{Z,Z'\iid P}\left[\rho(Z,Z')\right]<\infty$ also implies $\EE_{Z\sim P}\left[\gE(Z)\right]<\infty$.

    \qed

    \todo{LATER: state propriety in terms of $\gE$ - not urgent.}

\todo{modified: moved assumptions verification after the moment restriction.}
    \paragraph{Verification of the assumptions (A1)-(A4) for Student-t}
    \label{app:proof_student_t_assumptions}

    Since $\nabla\gE$ is bounded, (A2)--(A4) hold for every data distribution $p$; we verify this below.
\todo{modified}

    For the Student-t potential $\gE(u)=\log(1+\|u\|_{\Sigma^{-1}}^2)$, (A1) is exactly the stated restriction
    $c>d/2$.
    Furthermore, the gradient satisfies
    $$
    \|\nabla\gE(u)\|
    =\frac{2\|\Sigma^{-1}u\|}{1+\|u\|_{\Sigma^{-1}}^2}
    \leq 2\|\Sigma^{-1/2}\|_{\mathrm{op}}
    \frac{\|u\|_{\Sigma^{-1}}}{1+\|u\|_{\Sigma^{-1}}^2}
    \leq \|\Sigma^{-1/2}\|_{\mathrm{op}}.
    $$
    Thus $\gE$ is smooth and has bounded gradient, which gives (A3) and (A4) for every data distribution.

    It remains to check (A2). Since $q_{\gE,c}\leq Z_\gE(c)^{-1}$ and
    $\nabla q_{\gE,c}=-c\,q_{\gE,c}\nabla\gE$, the gradient of the noise density is also bounded by a constant $C$.
    By the mean-value theorem, the difference quotients of the
    noise density in any coordinate direction $e_i$ then satisfy, for all $x$ and all $h\neq 0$:
    $$
    \left|\frac{q_{\gE,c}(y+h e_i-x)-q_{\gE,c}(y-x)}{h}\right| \;\leq\; C,
    $$
    and the constant $C$ is integrable under any data distribution $p$.
    Now, (A2) asserts that the generalized Tweedie formula (Eq.~\ref{eq:general_tweedie}) holds at $y$,
    and the Stein score is given by $s_m(y)=\nabla_y\log m(y)=\nabla m(y)/m(y)$.
    We can thus exchange the gradient with the integral in the convolution by dominated convergence:
    $$
    \nabla m(y)=\nabla_y \int q_{\gE,c}(y-x)\,p(x)\,dx=\int \nabla q_{\gE,c}(y-x)\,p(x)\,dx.
    $$
    Dividing the result by $m(y)$ and using Bayes' rule gives:
    $$
    s_m(y)
    =\frac{1}{m(y)}\int \nabla q_{\gE,c}(y-x)p(x)\,dx
    =\EE\left[\nabla\log q_{\gE,c}(y-X)\mid Y=y\right],
    $$
    which is (A2).

    \clearpage
    \subsection{Appendix for Score Estimation}
    \label{app:score}

    \subsubsection{Bias-Variance decomposition of the Monte Carlo Score estimator}
    \label{app:bias_variance_score}

    Let
    \begin{equation}
        h_\eta(y,x)=-\lambda\Sigma^{-1}\|y-x\|_{\Sigma^{-1}}^{\beta-2}(y-x)
        \label{eq:cc_score}
    \end{equation}
     be the clean-conditioned score and
    $$
        \widehat s_N(y)=\frac1N\sum_{i=1}^N h_\eta(y,X^{(i)}),
    $$
    where $X^{(i)}\iid P_\theta(\cdot\mid y)$ for $i=1,\ldots,N$. The conditional error separates exactly as
    \begin{equation}
    \begin{aligned}
        \EE_{X^{(i)}\iid P_\theta(\cdot\mid y)}
        \!\left[\|\widehat s_N(y)-s_m(y)\|_2^2\right]
        &=
        \left\|\EE_{X\sim P_\theta(\cdot\mid y)}[h_\eta(y,X)]-s_m(y)\right\|_2^2 \\
        &\quad+\frac1N\operatorname{tr}\operatorname{Cov}_{X\sim P_\theta(\cdot\mid y)}(h_\eta(y,X)).
    \end{aligned}
        \label{eq:score_mc_error_decomposition}
    \end{equation}

    To see this, let $\mu_\theta(y)=\EE_{X\sim P_\theta(\cdot\mid y)}[h_\eta(y,X)]$.
    Conditional i.i.d. sampling gives
    $$\EE_{X^{(i)} \iid P_\theta(\cdot\mid y)}[\widehat s_N(y)]=\mu_\theta(y)$$ and
    $$\operatorname{Cov}_{X^{(i)}\iid P_\theta(\cdot\mid y)}(\widehat s_N(y))
    = \tfrac1N \operatorname{Cov}_{X\sim P_\theta(\cdot\mid y)}(h_\eta(y,X)).$$
    Expanding the squared error around $\mu_\theta(y)$ gives
    \begin{align*}
        &\EE_{X^{(i)}\iid P_\theta(\cdot\mid y)}
        \!\left[\|\widehat s_N(y)-s_m(y)\|_2^2\right] \\
        &\quad =\EE_{X^{(i)}\iid P_\theta(\cdot\mid y)}
        \!\Big[\|\widehat s_N(y)-\mu_\theta(y)\|_2^2\Big]
        +\|\mu_\theta(y)-s_m(y)\|_2^2 \\
        &\qquad\quad
        +2\EE_{X^{(i)}\iid P_\theta(\cdot\mid y)}
        \!\Big[(\widehat s_N(y)-\mu_\theta(y))^\top(\mu_\theta(y)-s_m(y))\Big] \\
        &\quad =\operatorname{tr}\operatorname{Cov}_{X^{(i)}\iid P_\theta(\cdot\mid y)}(\widehat s_N(y))
        +\|\mu_\theta(y)-s_m(y)\|_2^2 \\
        &\qquad\quad
        +2\underbrace{\left(
        \EE_{X^{(i)}\iid P_\theta(\cdot\mid y)}[\widehat s_N(y)]-\mu_\theta(y)
        \right)^\top}_{= \mu_\theta(y) - \mu_\theta(y) = 0}(\mu_\theta(y)-s_m(y)) \\
        &\quad =\|\mu_\theta(y)-s_m(y)\|_2^2
        +\frac1N\operatorname{tr}\operatorname{Cov}_{X\sim P_\theta(\cdot\mid y)}(h_\eta(y,X)).
    \end{align*}

    \qed

    The first term is the squared bias from the learned posterior and does not depend on $N$.
    The second term is posterior Monte Carlo variance and decays as $1/N$.
     It requires $\beta>1-d/2$: $\|h_\eta(y,x)\|_2^2$ scales as $\|y-x\|_2^{2(\beta-1)}$ near $x=y$, hence the covariance can be infinite in 1 dimension.

    \subsubsection{Empirical Score Comparisons}
    \label{app:score_comparison}

    \paragraph{Estimators and evaluation.}
    As a reminder, we compare three approaches:
    \begin{itemize}
        \item the \textit{posterior route} uses the identity directly in conjunction with samples from the denoising posterior,  averaging the known clean-conditioned score $h_\eta(y,X)$ (Eq.~\ref{eq:cc_score}) over $192$ draws.
        \item \textit{ET-DSM} learns a neural network  $s_\theta(y,\eta)$ by regressing against $h_\eta(Y,X)$ using clean--noisy pairs.
        \item \textit{Hyv\"arinen score matching} uses a similar neural architecture
        but trains on noisy data alone. While in principle it does not require the noise parameters as an input (cf. Sec.~\ref{sec:param_estimation}), we condition the network on the noise parameters to reuse it across experiments (i.e., not have to retrain it in each setting).
    \end{itemize}

    \paragraph{Noise magnitude.}
    All comparisons in this appendix use the five-parameter generalized Gaussian noise of Sec.~\ref{sec:score_estimation},
    $\eta=(\beta,\lambda,\sigma,u,v)$ with $\Sigma=\sigma^2S(u,v)$, on the two-dimensional Eight-Gaussians data.
    The scale $\sigma$ alone does not determine how far the noise displaces an observation, since the displacement also
    depends on $\beta$ and $\lambda$. To compare parameter combinations at equal noise strength, we use the per-coordinate root mean
    square (RMS) displacement of a noise draw $r_\eta$:
    \begin{equation}
        c_{\beta,d}^2=\frac{\beta^{2/\beta}\,\Gamma((d+2)/\beta)}{d\,\Gamma(d/\beta)},
        \qquad \rho=c_{\beta,d}\,\sigma,
        \qquad r_\eta=\rho\,\lambda^{-1/\beta},
        \label{eq:noise_rms_scales}
    \end{equation}
    where $d$ is the ambient dimension, here $d=2$.
    We call $\rho$, the value of $r_\eta$ at $\lambda=1$, the reference RMS; it is the magnitude coordinate on which the five-parameter models are conditioned (App.~\ref{app:training}). In the experimental appendices and figures, $\rho$ denotes this scalar, unrelated to the kernel discrepancy of Sec.~\ref{sec:ET}, as should be apparent from context.

    \paragraph{Noise-parameter combinations and ground truth.}
    The sweeps of Table~\ref{tab:score_full5_accuracy} vary one parameter at a time around the default combination
    $(\beta,\lambda,\sigma,u,v)=(1.4,\,1.8,\,0.45,\,0,\,0.6)$: $\beta\in\{1.3,1.5,1.75,2\}$,
    $\lambda\in\{1,1.4,1.8,2.5\}$, $\sigma\in\{0.1,0.25,0.45,0.75\}$, and $u,v\in\{-0.6,-0.3,0,0.3,0.6\}$, where the other  coordinate is held at $0$ while $u$ or $v$ is swept.
    All combinations are specified through the reference RMS $\rho$ and converted to $\sigma$ at the combination's
      $\beta$; the $\beta$ sweep therefore holds $\rho=0.570$ fixed while the tail shape changes, and the magnitude sweep
      $\rho\in\{0.13,0.32,0.57,0.95\}$ is reported as $\sigma\in\{0.1,0.25,0.45,0.75\}$ at $\beta=1.4$.
    In total there are $22$ parameter combinations with $192$ noisy observations each.

    When the noise is Gaussian, the noised score is analytic. This is not the case for generalized Gaussian noise, so to obtain a high-quality reference for evaluation we construct a numerical ground truth by importance sampling from the clean data distribution:
    a bank of $160{,}000$ clean particles is reweighted by the generalized Gaussian likelihood at the observation, and the ground-truth score is the weighted average of the clean-conditioned score $h_\eta$ (Eq.~\ref{eq:cc_score}) over the bank.
\todo{modified}
    The bank is split into two independent halves. The ground truth is the average of the two half-bank estimates, and the NMSE between the halves, reported as the ground-truth disagreement in Table~\ref{tab:score_full5_accuracy}, measures the precision of the ground truth itself.
    The posterior route averages $192$ posterior draws per observation. The direct estimators are evaluated at the same parameter combinations with the same number of observations and the same bank
    size, but with separately generated observations and ground truths, so Table~\ref{tab:score_reference_diagnostics} lists the ground-truth precision of each evaluation.

\begin{figure}[ht!]
        \centering
        \includegraphics[width=\textwidth]{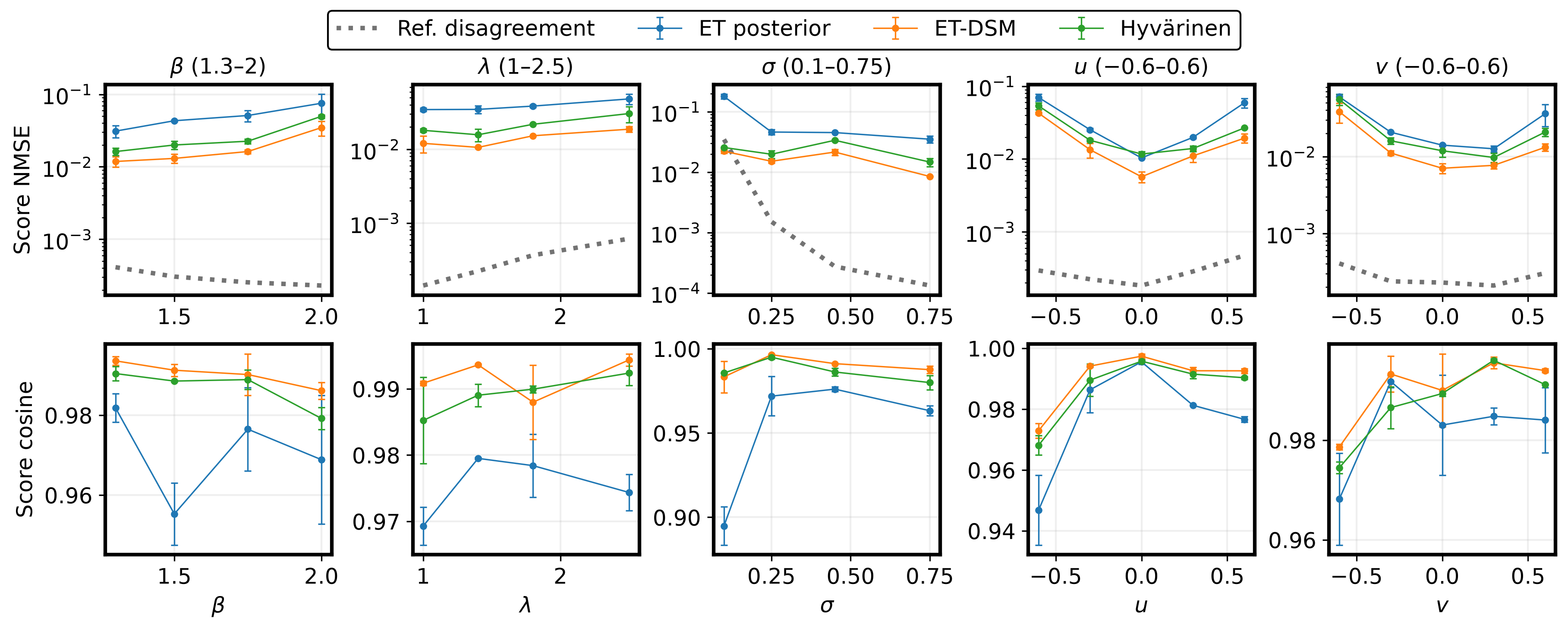}
        \caption{\textbf{Score NMSE (top, logarithmic scale) and cosine similarity (bottom) of the three estimators at every
        parameter combination.} Error bars are sample SDs across two training seeds.
        The dotted gray curves show the half-bank ground-truth NMSE for the posterior evaluation.}
        \label{fig:score_comparison_sweeps}
    \end{figure}

    \paragraph{Accuracy across the sweeps.}
    Figure~\ref{fig:score_comparison_sweeps} shows NMSE and cosine similarity at every parameter combination, which is the data behind the means presented in Table~\ref{tab:score_full5_accuracy} in the main text.   
    ET-DSM has the lowest NMSE in every sweep, followed by Hyv\"arinen score matching and then the posterior route.  The three estimators agree in direction and magnitude: the sweep-mean cosine similarity exceeds $0.95$ for every  estimator, and the norm ratios lie within a few percent of one (Table~\ref{tab:score_norm_ratios}).
    The largest error of the posterior route occurs at the smallest scale $\sigma=0.1$, where the importance-sampled ground truth is  also least precise (Table~\ref{tab:score_reference_diagnostics}): as the noise shrinks, the posterior concentrates on  few bank particles, the effective sample size drops, and the two half-bank estimates disagree by an NMSE far larger than at the other combinations. As noted in the main text, while the two direct estimators were ensured to be well-trained in order for a maximally rigorous comparison (see \textit{Training of the direct estimators} below), no such procedure was done for the posterior, so its slightly lower accuracy might be due to undertraining.

    We can also compare the posterior model draws and the importance posterior directly. Table~\ref{tab:score_posterior_diagnostics} reports the error of the
    posterior mean, the ratio of the posterior covariance traces, the total variation distance between the assignments of posterior mass to the nearest mixture centers, and the energy distance between the two posteriors, together with its excess over the energy distance between two independent resamples of the importance posterior. The results show that the learned posterior is centered and assigns modes correctly; its remaining error is a mild overdispersion (trace ratio $1.04$--$1.11$), consistent with the nonzero score NMSE that the posterior route retains as $N\to\infty$ (Table~\ref{tab:score_mc_budget}) and pointing to mild undertraining.

    \begin{table}[ht!]
        \centering
        \small
        \begin{tabular}{@{}lccc@{}}
            \toprule
            Sweep & ET posterior & ET-DSM & Hyv\"arinen \\
            \midrule
            $\beta$ & $0.970 \pm 0.011$ & $0.989 \pm 0.002$ & $0.985 \pm 0.013$ \\
            $\lambda$ & $0.989 \pm 0.003$ & $1.003 \pm 0.001$ & $1.006 \pm 0.005$ \\
            $\sigma$ & $1.036 \pm 0.013$ & $0.990 \pm 0.003$ & $0.982 \pm 0.001$ \\
            $u$ & $0.987 \pm 0.003$ & $0.991 \pm 0.002$ & $0.991 \pm 0.006$ \\
            $v$ & $0.998 \pm 0.003$ & $0.995 \pm 0.003$ & $0.995 \pm 0.003$ \\
            \bottomrule
        \end{tabular}
        \caption{\textbf{Ratio of mean estimated-score norm to mean ground-truth score norm.} Values are means
        $\pm$ sample SDs across training seeds after equal averaging within each sweep.}
        \label{tab:score_norm_ratios}
    \end{table}
    \begin{table}[ht!]
        \centering
        \small
        \setlength{\tabcolsep}{3pt}
        \resizebox{\textwidth}{!}{%
        \begin{tabular}{@{}lcccccc@{}}
            \toprule
            Sweep & Mean MSE & Trace ratio & Mode TV & ED & ED SD & ED excess \\
            \midrule
            $\beta$ & $0.0057 \pm 0.0012$ & $1.112 \pm 0.033$ & $0.044 \pm 0.004$ & $0.0127 \pm 0.0008$ & $0.0256 \pm 0.0018$ & $0.0129 \pm 0.0008$ \\
            $\lambda$ & $0.0042 \pm 0.0003$ & $1.042 \pm 0.001$ & $0.043 \pm 0.002$ & $0.0087 \pm 0.0006$ & $0.0170 \pm 0.0011$ & $0.0088 \pm 0.0006$ \\
            $\sigma$ & $0.0037 \pm 0.0004$ & $1.046 \pm 0.007$ & $0.029 \pm 0.002$ & $0.0066 \pm 0.0009$ & $0.0164 \pm 0.0053$ & $0.0069 \pm 0.0009$ \\
            $u$ & $0.0042 \pm 0.0004$ & $1.051 \pm 0.008$ & $0.034 \pm 0.001$ & $0.0092 \pm 0.0005$ & $0.0274 \pm 0.0000$ & $0.0094 \pm 0.0005$ \\
            $v$ & $0.0035 \pm 0.0004$ & $1.045 \pm 0.009$ & $0.032 \pm 0.002$ & $0.0067 \pm 0.0009$ & $0.0168 \pm 0.0025$ & $0.0068 \pm 0.0009$ \\
            \bottomrule
        \end{tabular}}
        \caption{\textbf{Posterior diagnostics of the learned posterior against the importance posterior, means $\pm$ sample
        SDs across training seeds.} Mean MSE is the squared error of the posterior mean; trace ratio is the ratio of
        mean posterior covariance traces; mode TV is the total variation distance between nearest-mixture-center
        assignments. ED is the unbiased energy distance from the learned to the importance posterior, evaluated at
        $48$ observations per parameter combination; ED SD is its standard deviation across those observations.
        ED excess subtracts the ED between two ground-truth resamples of Table~\ref{tab:score_reference_diagnostics}.}
        \label{tab:score_posterior_diagnostics}
    \end{table}
    \begin{table}[ht!]
        \centering
        \small
        \setlength{\tabcolsep}{4pt}
        \begin{tabular}{@{}lccrrc@{}}
            \toprule
            & \multicolumn{2}{c}{Split ground-truth NMSE} & \multicolumn{2}{c}{Importance ESS} & \\
            \cmidrule(lr){2-3}\cmidrule(lr){4-5}
            Sweep & Posterior & Direct & Median & Minimum & Ground-truth ED \\
            \midrule
            $\beta$ & $0.0003$ & $0.0003$ & $14,713$ & $4,581$ & $-0.0002 \pm 0.0029$ \\
            $\lambda$ & $0.0003$ & $0.0003$ & $14,491$ & $5,906$ & $-0.0002 \pm 0.0029$ \\
            $\sigma$ & $0.0092$ & $0.0111$ & $10,965$ & $4,291$ & $-0.0003 \pm 0.0020$ \\
            $u$ & $0.0003$ & $0.0003$ & $14,523$ & $5,277$ & $-0.0002 \pm 0.0027$ \\
            $v$ & $0.0003$ & $0.0002$ & $14,307$ & $5,475$ & $-0.0001 \pm 0.0027$ \\
            \bottomrule
        \end{tabular}
        \caption{\textbf{Ground-truth diagnostics averaged over parameter combinations within each sweep.} Split ground-truth NMSE is the NMSE  between the two independent half-bank score estimates, for the posterior and the direct evaluations.
        ESS columns average the per-combination median and minimum importance effective sample sizes of the posterior evaluation. Ground-truth ED is the energy distance between independent resamples of the importance posterior; the value after $\pm$ is the average within-combination SD over observations.}
        \label{tab:score_reference_diagnostics}
    \end{table}

\begin{figure}[ht!]
        \centering
        \includegraphics[width=\textwidth]{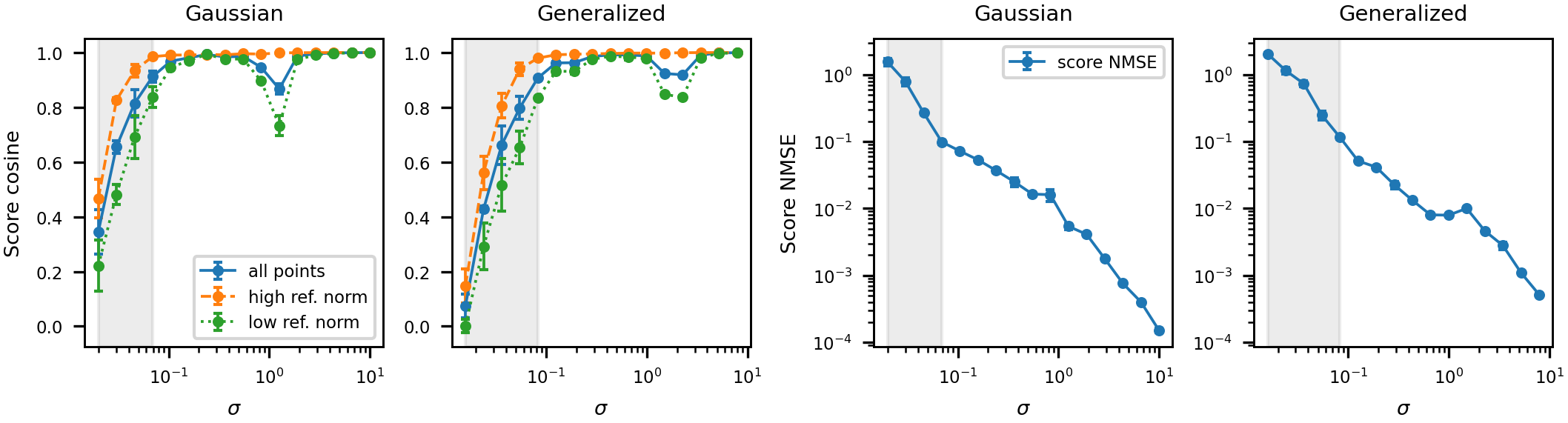}
        \caption{\textbf{Posterior-route score cosine (left two panels) and score NMSE (right two panels) against noise magnitude for isotropic Gaussian and generalized Gaussian noise.} Cosine is shown over all observations and separately split into cases where the corresponding ground truth is  above- or below-median in norm. Error bars are standard deviations across two independently trained posterior models. Gray shading marks magnitudes at which the half-bank ground-truth NMSE exceeds $0.01$.}
        \label{fig:score_alignment}
    \end{figure}

    \paragraph{Noise-magnitude range.}
    The sweeps above span less than one decade of noise magnitude. 
    Figure~\ref{fig:score_alignment} supplements this by evaluating the posterior
    route over three decades, using the isotropic Gaussian and generalized Gaussian noise with the remaining    parameters fixed. Cosine similarity dips at intermediate magnitudes, with the dip confined to the observations whose ground-truth score has below-median norm: where the ground-truth score is small, a small absolute error produces a large angular error, and the score NMSE in this region stays at or below $0.010$. Replacing the learned posterior by the exact one (analytic for Gaussian noise, importance-sampled for generalized Gaussian noise) reproduces the dip (Table~\ref{tab:score_dip_control}), which places the responsibility in the Monte Carlo estimate at those observations rather than in the learned model. At the smallest magnitudes, shaded gray, the half-bank ground-truth NMSE exceeds $0.01$ and the comparison is  thus limited by the ground truth.
    \begin{table}[ht!]
        \centering
        \small
        \setlength{\tabcolsep}{3pt}
        \begin{tabular}{@{}llcccc@{}}
            \toprule
            Noise & Posterior & NMSE & Cosine (all) & Cosine (high norm) & Cosine (low norm) \\
            \midrule
            Gaussian, $\sigma=1.260$ & learned & $0.0054$ & $0.867$ & $0.999$ & $0.732$ \\
            Gaussian, $\sigma=1.260$ & exact & $0.0055$ & $0.875$ & $0.999$ & $0.752$ \\
            Generalized, $\sigma=2.277$ & learned & $0.0046$ & $0.919$ & $0.999$ & $0.836$ \\
            Generalized, $\sigma=2.277$ & importance-sampled & $0.0036$ & $0.904$ & $1.000$ & $0.809$ \\
            \bottomrule
        \end{tabular}
        \caption{\textbf{Cosine-dip control at the dip location of Fig.~\ref{fig:score_alignment}, with $N=192$ posterior draws
        per observation.} ``Learned'' is the five-parameter posterior model; ``exact'' and ``importance-sampled'' draw
        from the true posterior  and apply the same Monte Carlo score estimator. ``High/low norm'' correspond to the same two cases as in Fig.~\ref{fig:score_alignment}.}
        \label{tab:score_dip_control}
    \end{table}

    \paragraph{Training of the direct estimators.}
    ET-DSM and Hyv\"arinen score matching share the architecture, initialization, and data stream described in
    App.~\ref{app:training}. With the factor $\tfrac12$ included by convention, the objectives at a fixed parameter combination $\eta$ are
    \begin{align}
        D_\eta(\theta)&=\tfrac12\mathcal{L}_{\mathrm{ET\text{-}DSM}}(\theta;\eta),\nonumber\\
        H_\eta(\theta)&=\EE_{Y\sim m_\eta}
        [\tfrac12\|s_\theta(Y,\eta)\|_2^2+\operatorname{div}_Y s_\theta(Y,\eta)],
        \label{eq:score_comparison_objectives}
    \end{align}
    with $\mathcal{L}_{\mathrm{ET\text{-}DSM}}$ from Eq.~\ref{eq:et_dsm_loss} and $m_\eta=p*q_\eta$ the noisy marginal.

    Each minibatch draws one parameter combination from the law used to train the five-parameter Eight-Gaussians posterior (App.~\ref{app:training}). The objectives are averaged over combinations with the weight $r_\eta^2$ in order for the small-noise cases not to dominate the loss.

\begin{figure}[ht!]
        \centering
        \includegraphics[width=\textwidth]{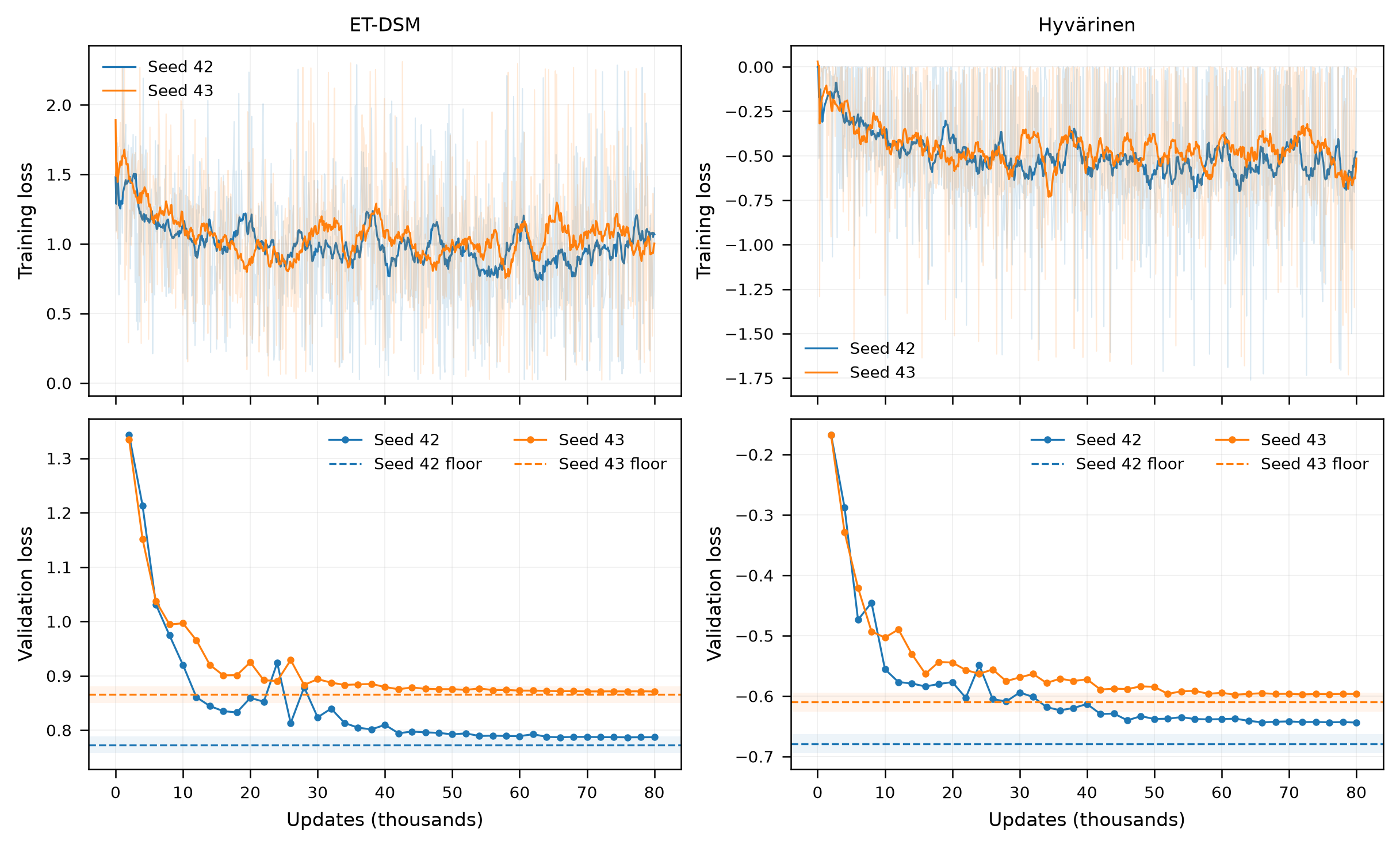}
        \caption{\textbf{$r_\eta^2$-weighted training (top) and fixed-validation (bottom) objectives for the two direct
        estimators.} Faint curves are logged minibatch losses; solid training curves are trailing means over
        $20$ logged batches. Validation uses eight fixed parameter combinations per seed, with $1{,}024$ observations
        per combination. Dashed lines are the estimated population floors.}
        \label{fig:score_comparison_training}
    \end{figure}

    Both objectives have known population minima, so we can ensure that the training has converged.
    Let $s_\eta=\nabla\log m_\eta$ be the noisy score,
    $I_\eta=\EE_{Y\sim m_\eta}[\|s_\eta(Y)\|_2^2]$ its expected squared norm, and
    $C_\eta=\EE_{X\sim p,\,Y\sim q_\eta(\cdot\mid X)}[\|h_\eta(Y,X)\|_2^2]$ that of the clean-conditioned score (Eq.~\ref{eq:cc_score}).
    Under the integration-by-parts conditions of score matching, we get \citep{hyvarinen_estimation_2005, vincent_connection_2011}:
    \begin{equation}
    \begin{aligned}
        H_\eta^*&=-I_\eta/2,\qquad D_\eta^*=(C_\eta-I_\eta)/2,\\
        H_\eta(\theta)-H_\eta^*&=D_\eta(\theta)-D_\eta^*
        =\tfrac12\EE_{Y\sim m_\eta}[\|s_\theta(Y,\eta)-s_\eta(Y)\|_2^2],
    \end{aligned}
        \label{eq:score_comparison_floors}
    \end{equation}
    so both objectives exceed their minima by half the mean squared score error, and this excess divided by $I_\eta/2$ is the score NMSE at that combination.
    For the two-dimensional generalized Gaussian,
    $$
        C_\eta=\frac{\lambda^2\operatorname{tr}(S^{-1})}{2\sigma^2}
        \left(\frac{\beta}{\lambda}\right)^{2-2/\beta}
        \frac{1}{\Gamma(2/\beta)},
    $$
    while $I_\eta$ is estimated on the validation set from the inner product of the two half-bank ground-truth
    scores.
    Validation is evaluated every $2{,}000$ updates.
    We can thus evaluate the theoretical minima from Eq.~\ref{eq:score_comparison_floors} and perform the same weighted average as for the loss to obtain a \textit{loss floor}  in Fig.~\ref{fig:score_comparison_training}. 
    The validation curves settle near their floors, and the final relative gaps are between $1\%$ and $5\%$
    (Table~\ref{tab:score_validation_gaps}), so (at least at the chosen validation parameter values) the two network sets are near-optimally trained.
    The minibatch training losses are variable as each minibatch draws a new parameter combination with its own population minimum.
    \begin{table}[ht!]
        \centering
        \small
        \begin{tabular}{@{}lrrrr@{}}
            \toprule
            Estimator & Seed & Final loss & Estimated floor & Relative gap \\
            \midrule
            ET-DSM & $42$ & $0.7871$ & $0.7731$ & $2.0\%$ \\
            ET-DSM & $43$ & $0.8711$ & $0.8652$ & $1.0\%$ \\
            Hyv\"arinen & $42$ & $-0.6439$ & $-0.6790$ & $5.2\%$ \\
            Hyv\"arinen & $43$ & $-0.5965$ & $-0.6098$ & $2.2\%$ \\
            \bottomrule
        \end{tabular}
        \caption{\textbf{Final weighted validation losses and estimated population floors.} }
        \label{tab:score_validation_gaps}
    \end{table}

    \paragraph{Compute Budgets.}
    The three estimators share the clean distribution and the noise-parameter law above.
    Each direct estimation network is trained on $327.68$ million fresh clean--noisy pairs under $80{,}000$ draws of the parameter combination; the
    posterior model, which is also the model behind the diffusion experiments of App.~\ref{app:diffusion_full5}, was
    trained on $22.4$ billion pairs under $560{,}000$ draws of the parameter combination.
    Naturally, the networks are not comparable, as the direct estimation networks estimate the score alone, while we use the same posterior network for all three presented applications.
    For the posterior route, Eq.~\ref{eq:score_mc_error_decomposition} separates the error at an observation into the
    squared bias of the learned posterior, which additional draws do not reduce, and the Monte Carlo variance, which
    decays as $1/N$.
    Table~\ref{tab:score_mc_budget} holds the two trained posteriors (i.e., two random seeds) fixed at the default parameter combination and measures the error over nine posterior sample budgets, with ten repetitions per model.
\todo{modified: posterior-budget plot uses the full minipage width.}

    \begin{table}[ht!]
        \centering
        \begin{minipage}[c]{0.52\linewidth}
            \centering
            \small
            \begin{tabular}{@{}rcc@{}}
                \toprule
                Posterior samples $N$ & Score NMSE $\downarrow$ & Score cosine $\uparrow$ \\
                \midrule
                $1$   & $1.446 \pm 0.110$ & $0.574 \pm 0.039$ \\
                $2$   & $0.698 \pm 0.061$ & $0.681 \pm 0.034$ \\
                $4$   & $0.368 \pm 0.052$ & $0.780 \pm 0.041$ \\
                $8$   & $0.196 \pm 0.018$ & $0.856 \pm 0.025$ \\
                $16$  & $0.112 \pm 0.012$ & $0.909 \pm 0.016$ \\
                $32$  & $0.071 \pm 0.010$ & $0.938 \pm 0.013$ \\
                $64$  & $0.050 \pm 0.008$ & $0.961 \pm 0.009$ \\
                $128$ & $0.039 \pm 0.006$ & $0.973 \pm 0.007$ \\
                $256$ & $0.034 \pm 0.006$ & $0.980 \pm 0.005$ \\
                \bottomrule
            \end{tabular}
        \end{minipage}\hfill
        \begin{minipage}[c]{0.45\linewidth}
            \centering
            \includegraphics[width=\linewidth]{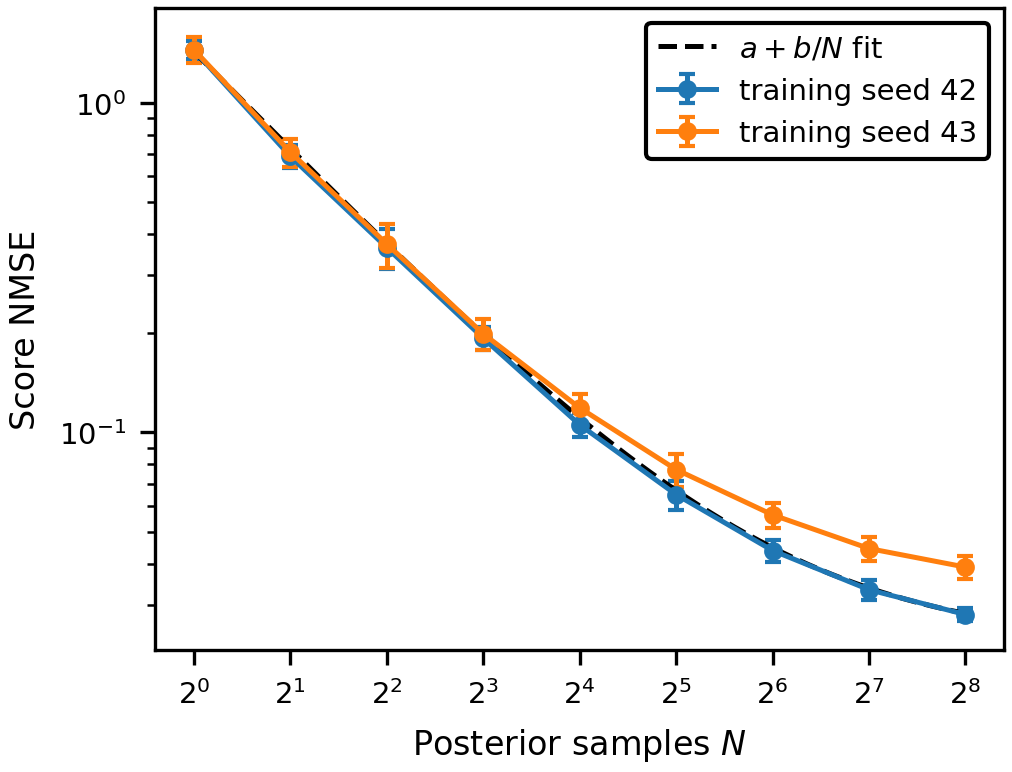}
        \end{minipage}
        \caption{\textbf{Score accuracy versus posterior draw budget.} Left: score NMSE and cosine similarity versus the number of posterior draws per observation;
        the standard deviations are over $20$ model--repetition units from two independently trained models: ten repetitions per model.
        Right: per-model score NMSE over the same budgets with the pooled $a+b/N$ fit (dashed); error bars are per-model.
        }
        \label{tab:score_mc_budget}
    \end{table}
    Fitting $a+b/N$ jointly over the nine posterior sample budgets gives an NMSE floor $a=0.023$ and slope $b=1.41$. The two models have nearly identical slopes ($1.409$ and $1.406$) and different floors ($0.017$ and $0.029$), matching the separation in Eq.~\ref{eq:score_mc_error_decomposition} between posterior variance and per-model approximation bias (which differs between model instances).

\todo{modified: Richardson posterior models are introduced here.}

\begin{figure}[ht!]
        \centering
        \includegraphics[width=.7\linewidth]{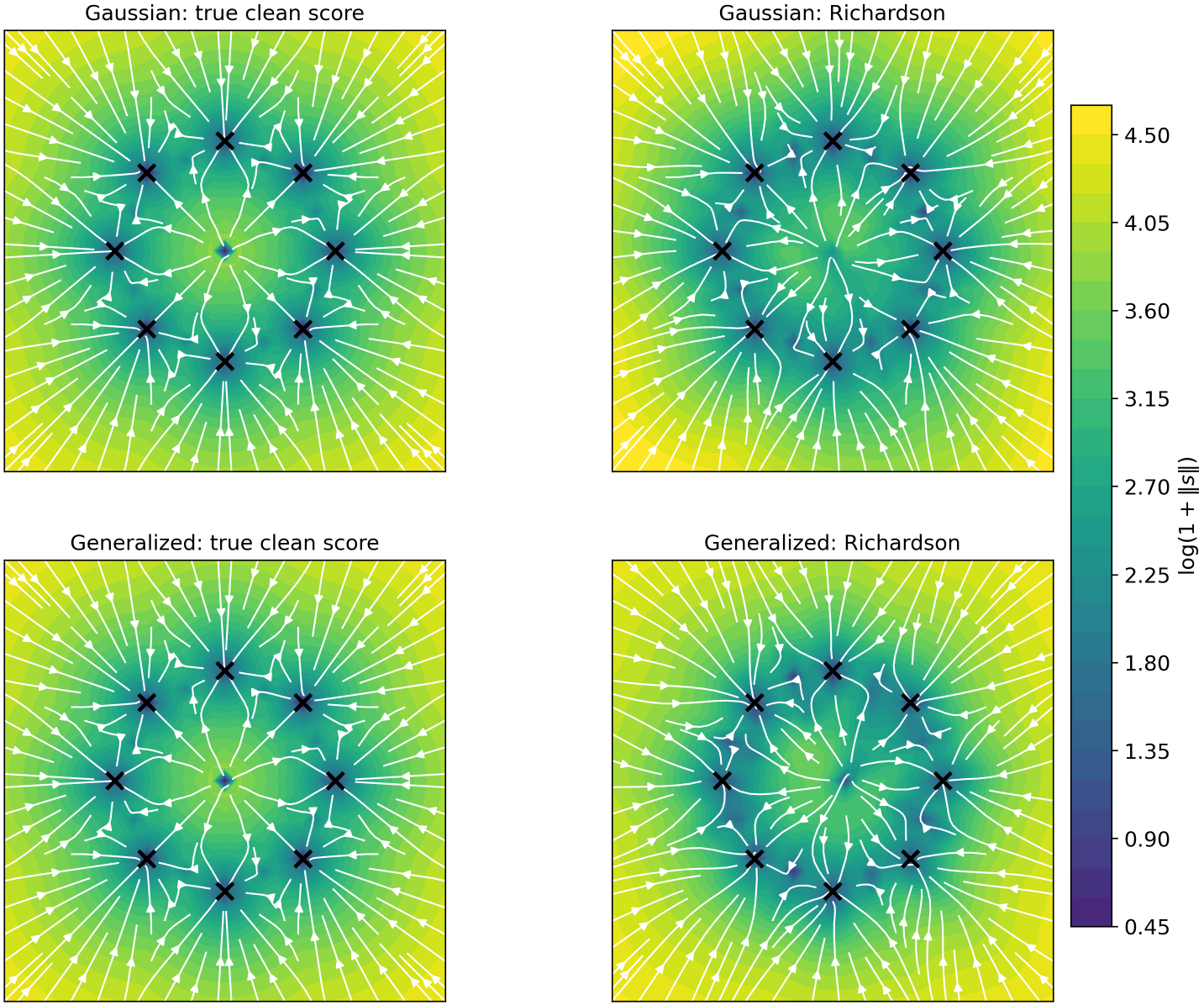}
        \caption{\textbf{Clean data score fields for the Eight-Gaussians dataset.} \textbf{Top:} Gaussian, \textbf{bottom:} generalized
            Gaussian noise. In each row, the left panel shows the analytic ground-truth clean field,
            and the right panel shows the field obtained via
            Richardson extrapolation.
        }
        \label{fig:score_field}
    \end{figure}

    \paragraph{Clean score fields.}
    Fig.~\ref{fig:score_field} presents the approximate clean score field obtained via the Richardson extrapolation
    (Sec.~\ref{sec:score_estimation}) and the analytic ground truth. We use two MLP-based
    \textit{Engression} posteriors \citep{shen_engression_2024}, conditioned on $\log\sigma$ and trained over
    $\sigma\in[0.01,10]$, for isotropic Gaussian $(\beta,\lambda)=(2,1)$ and generalized Gaussian
    $(\beta,\lambda)=(1.4,1.8)$ noise. The Richardson pair $(\varepsilon_1,\varepsilon_2)$ is the adjacent pair from
    that training-range noise grid with the lowest clean-score MSE. As the models were not trained to target the
    small-noise regime, the comparison is mostly qualitative.

    \begin{figure}[ht!]
        \centering
        \includegraphics[width=.7\linewidth]{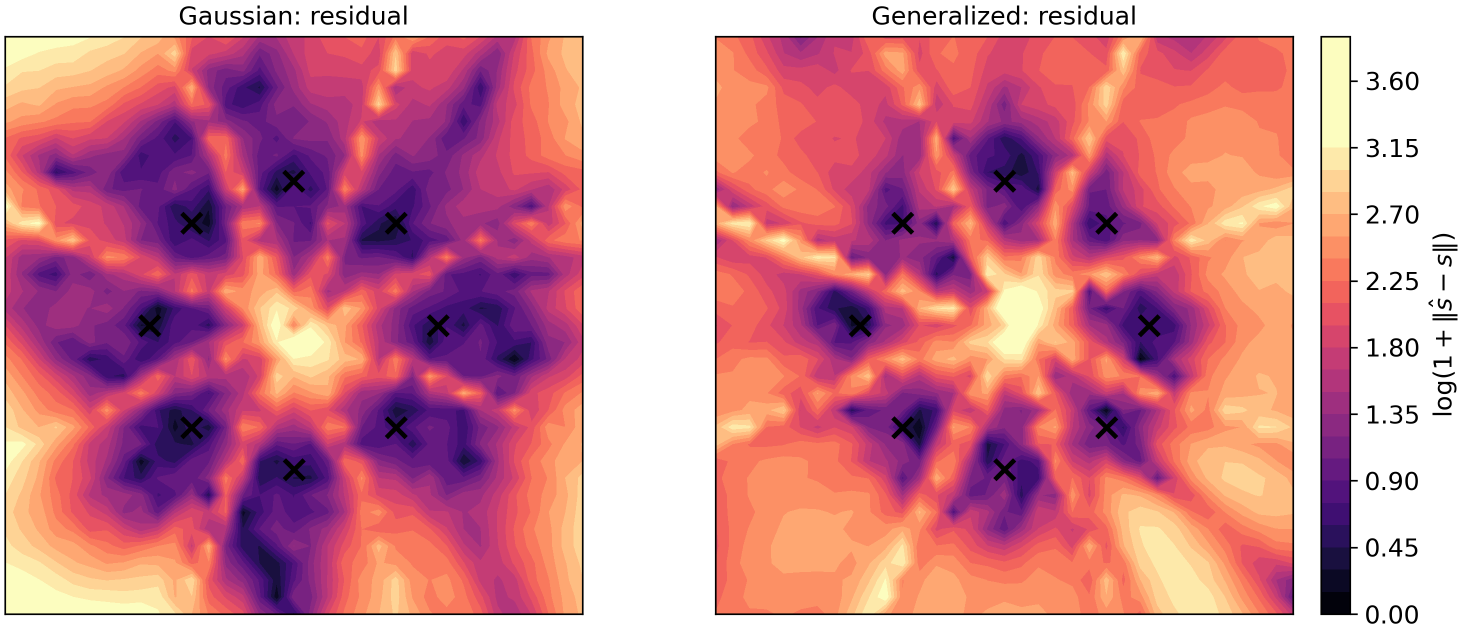}
        \caption{\textbf{Residual magnitudes for the clean score-field comparison in Fig.~\ref{fig:score_field}.}
         \textbf{Left:} Gaussian,  and \textbf{right:} generalized Gaussian noise.
            The color encodes $\log(1+\|\widehat s_{\mathrm{RE}}(y)-\nabla_y\log p(y)\|)$.
        }
        \label{fig:app_score_field_residual}
    \end{figure}

\newpage

    \subsubsection{Derivations}
    \label{app:proof_richardson}

    \paragraph{Small-noise limit.}

    First, Cor.~\ref{cor:gen_gaussian} holds for any choice of parameters $(\beta,\lambda,\Sigma)$. The Mahalanobis
    metric reduces to the Euclidean one in whitened coordinates:
    $$
    \|u\|^\beta_{\Sigma^{-1}}=\|\Sigma^{-1/2}u\|_2^\beta.
    $$
    Now write $\Sigma=\varepsilon^2\Sigma_0$. For the generalized Gaussian distribution,
    \begin{equation}
        q_\varepsilon(u)\ \propto\ \exp \Big( - \frac{\lambda}{\beta~ \varepsilon^\beta} ~
            \| \Sigma_0^{-1/2} u \|_2^\beta \Big),
        \label{eq:q_epsilon}
    \end{equation}
    and $\|u\|_{\Sigma^{-1}}=\varepsilon^{-1}\|\Sigma_0^{-1/2}u\|_2$. Let $P_\varepsilon(\cdot\mid Y=y)$ denote the
    posterior induced by the clean density $p$ and noising density $q_\varepsilon$, and let
    $m_\varepsilon=p*q_\varepsilon$. The Energy-Tweedie identity gives
    \begin{equation}
        s_{m_\varepsilon} (y) = -\frac{\lambda}{\beta} \; \frac{1}{ \varepsilon^\beta}
            \nabla^{\mathrm{PD}}_y ~ \EE_{X\sim P_\varepsilon(\cdot\mid Y=y)}
            [\| \Sigma_0^{-1/2} (X - y)\|_2^\beta ].
        \label{eq:score_eps_identity}
    \end{equation}

    This can be approximated via Monte Carlo samples from a trained posterior $X^{(k)} \sim \widehat{P}_{\theta,
    \varepsilon}(\cdot \mid Y=y)$:
    \begin{equation}
        \widehat{s}_{\varepsilon}(y) = -\frac{\lambda}{\beta ~\varepsilon^\beta} \nabla_y^{\mathrm{PD}} \frac{1}{K}
            \sum_{k=1}^K\left\|\Sigma_0^{-1 / 2}\left(X^{(k)}-y\right)\right\|_2^\beta,
        \label{eq:score_eps}
    \end{equation}
    (or equivalently using Eq.~\ref{eq:sm_finite_sample} if one does not wish to use automatic differentiation).

    For the limiting clean score, assume $p(y)>0$, $p$ is differentiable in a neighborhood of $y$, and the scaled
    kernels $q_\varepsilon$ form an approximate identity for both $p$ and $\nabla p$ at $y$:
    $$
    (p*q_\varepsilon)(y)\to p(y),\qquad (\nabla p*q_\varepsilon)(y)\to \nabla p(y).
    $$
    A basic sufficient condition is continuity of $p$ and $\nabla p$ near $y$, together with a dominated-convergence
    condition allowing convolution with the generalized-Gaussian tails. Under these conditions,
    $$
    \nabla_y m_\varepsilon(y)=\nabla_y(p*q_\varepsilon)(y)=(\nabla p*q_\varepsilon)(y)\to \nabla p(y),
    $$
    while $m_\varepsilon(y)\to p(y)>0$. Therefore
    $$
    s_{m_\varepsilon}(y)=\nabla_y\log m_\varepsilon(y)
    =\frac{\nabla_y m_\varepsilon(y)}{m_\varepsilon(y)}
    \longrightarrow \frac{\nabla_y p(y)}{p(y)}=\nabla_y\log p(y).
    $$
    Combining this with Eq.~\ref{eq:score_eps_identity} gives the small-noise clean-score limit used in
    Sec.~\ref{sec:score_estimation}. The stronger $\gC^7$ condition used for the Richardson expansion below should be
    read together with this dominated-convergence condition when the expansion is exchanged with the integral.

    \paragraph{Richardson error order.}
    The Richardson extrapolation estimator (Eq.~\ref{eq:richardson_extrapolation}) is defined for two small
    $\varepsilon_1, \varepsilon_2$ such that $\varepsilon_1=a \varepsilon, \varepsilon_2=b \varepsilon, ~~ a, b>0,
    a \neq b, \varepsilon \ll 1$ and assuming the same regularity and $p \in \gC^7$ around $y$.

    Note that $q_\varepsilon$:
    $$q_\varepsilon(u)\ \propto\ \exp \Big( - \frac{\lambda}{\beta~ \varepsilon^\beta} ~
        \| \Sigma_0^{-1/2} u \|_2^\beta \Big)$$
    is: a) symmetric: $q_\varepsilon(u) = q_\varepsilon(-u)$, and b), in the scale family:
    $$
    q_\varepsilon(\varepsilon u) \propto \exp \Big( - \frac{\lambda}{\beta~ \varepsilon^\beta} ~
        \| \Sigma_0^{-1/2} \varepsilon u \|_2^\beta \Big)
    = \exp \Big( - \frac{\lambda}{\beta~ \varepsilon^\beta} \varepsilon^\beta ~
        \| \Sigma_0^{-1/2} u \|_2^\beta \Big)
    = \exp \Big( - \frac{\lambda}{\beta} ~ \| \Sigma_0^{-1/2} u \|_2^\beta \Big)
    $$

    For the normalization constant $Z_\varepsilon=\int_{\mathbb{R}^d} \exp \left(-\frac{\lambda}{\beta
    \varepsilon^\beta}\left\|\Sigma_0^{-1 / 2} z\right\|_2^\beta\right) d z$, we have after changing the variables
    $z=\varepsilon v$, $d z=\left|\operatorname{det}\left(\frac{\partial z}{\partial v}\right)\right| d
    v=\left|\operatorname{det}\left(\varepsilon I_d\right)\right| d v = \varepsilon^d dv$:
    $$
    Z_\varepsilon=\varepsilon^d \int_{\mathbb{R}^d} \exp \left(-\frac{\lambda}{\beta}\left\|\Sigma_0^{-1 / 2}
        v\right\|_2^\beta\right) d v=\varepsilon^d Z_*,
    $$
    where
    $$
    Z_* \Def \int_{\mathbb{R}^d} \exp \left(-\frac{\lambda}{\beta}\left\|\Sigma_0^{-1 / 2} v\right\|_2^\beta\right) d v
    $$

    The noisy marginal is, by definition:
    $$
    m_\varepsilon(y)=(p*q_\varepsilon)(y)=\int p(x)\,q_\varepsilon(y-x)\,dx.
    $$
    Applying the change of variables $x=y-\varepsilon u$, so $d x = \varepsilon^d d u$:
    $$
    m_\varepsilon(y)=\varepsilon^d\int p(y-\varepsilon u)\,q_\varepsilon(\varepsilon u)\,du.
    $$

    Using the scaling relations established above, the rescaled weight
    $$
    w(u)\Def \varepsilon^d q_\varepsilon(\varepsilon u) =
    Z_*^{-1}\exp\left(-\frac{\lambda}{\beta}\|\Sigma_0^{-1/2}u\|_2^\beta\right)
    $$
    is independent of $\varepsilon$, and is even:
    $$
    w(u)=w(-u).
    $$

    Assume $p(y)>0$ and $p\in \gC^7$ near $y$, with the dominated-convergence condition from
    above. As the generalized Gaussian has finite moments for all polynomial
    orders~\citep{dytso_analytical_2018}, the same holds for $w$. Thus, we have the following expansion for $p$ for
    small increments $\varepsilon u$:
    $$
    p(y-\varepsilon u) = \sum_{k=0}^{5}\frac{(-\varepsilon)^k}{k!}(u^\top\nabla_y)^k p(y)\;+\;
    \O(\varepsilon^6\|u\|^6).
    $$

    Integrating against $w(u)\,du$ in the expression for the noisy marginal, all odd-order terms vanish by symmetry of
    $w$, hence
    $$
    m_\varepsilon(y) = p(y)+\varepsilon^2 A_2(y)+\varepsilon^4 A_4(y)+\O(\varepsilon^6),
    $$
    where
    $$
    A_2(y)=\frac12\int (u^\top\nabla_y)^2 p(y)\,w(u)\,du, \qquad A_4(y)=\frac1{24}\int (u^\top\nabla_y)^4
        p(y)\,w(u)\,du.
    $$

    Therefore
    $$
    \nabla_y m_\varepsilon(y) =
    \nabla_y p(y)+\varepsilon^2 \nabla_y A_2(y)+\varepsilon^4 \nabla_y A_4(y)+\O(\varepsilon^6),
    $$
    and so
    $$
    s_{m_\varepsilon}(y) = \nabla_y\log m_\varepsilon(y) =
    \frac{\nabla_y m_\varepsilon(y)}{m_\varepsilon(y)} =
    \nabla_y\log p(y) + \varepsilon^2 B_2(y) + \varepsilon^4 B_4(y) + O(\varepsilon^6),
    $$
    for some vector fields $B_2,B_4$.

    To find these, we can write:

    $$
    m_\varepsilon(y) =
    p(y)\left( 1+\varepsilon^2 \frac{A_2(y)}{p(y)}+\varepsilon^4\frac{A_4(y)}{p(y)}+\O(\varepsilon^6) \right),
    $$

    and
    $$
    \frac{1}{m_\varepsilon(y)} = \frac{1}{p(y)} \left( 1+\varepsilon^2
        \frac{A_2(y)}{p(y)}+\varepsilon^4\frac{A_4(y)}{p(y)}+\O(\varepsilon^6) \right)^{-1}.
    $$

    Using $(1+z)^{-1}=1-z+z^2+O(z^3)$ with $z=\varepsilon^2 \frac{A_2(y)}{p(y)}+\varepsilon^4\frac{A_4(y)}{p(y)} +
    \O(\varepsilon^6)$ gives
    $$
    \frac{1}{m_\varepsilon(y)} = \frac{1}{p(y)} -\varepsilon^2\frac{A_2(y)}{p(y)^2} +\varepsilon^4\left(
        \frac{A_2(y)^2}{p(y)^3}-\frac{A_4(y)}{p(y)^2} \right) + \O(\varepsilon^6).
    $$

    Thus, for the score (simplifying notation slightly):
    $$
        s_{m_\varepsilon}(y) = \frac{\nabla_y m_\varepsilon(y)}{m_\varepsilon(y)}
        = \Big( \nabla_y p+\varepsilon^2 \nabla_y A_2+\varepsilon^4 \nabla_y A_4+ \O(\varepsilon^6) \Big)
        \Big( \frac1p-\varepsilon^2\frac{A_2}{p^2}
            +\varepsilon^4\big(\frac{A_2^2}{p^3}-\frac{A_4}{p^2}\big) + \O(\varepsilon^6) \Big).
    $$

    The order 1 term is
    $$
        \frac{\nabla_y p(y)}{p(y)}=\nabla_y\log p(y).
    $$

    The order $\varepsilon^2$ term is
    $$
        \frac{\nabla_y A_2(y)}{p(y)}-\frac{A_2(y)}{p(y)^2}\nabla_y p(y) = \frac{\nabla_y
            A_2(y)}{p(y)}-\frac{A_2(y)}{p(y)}\,\nabla_y\log p(y) \Def B_2 (y).
    $$

    The order $\varepsilon^4$ term is:
    $$
        \frac{\nabla_y A_4(y)}{p(y)} -\frac{A_2(y)}{p(y)^2}\nabla_y A_2(y)
            +\left(\frac{A_2(y)^2}{p(y)^3}-\frac{A_4(y)}{p(y)^2}\right)\nabla_y p(y) \Def B_4(y).
    $$

    Now define the ``exact noisy score'' Richardson extrapolator as:
    $$
    s_{\mathrm{RE}}(y) \Def
    \frac{\varepsilon_2^2\,s_{m_{\varepsilon_1}}(y)-\varepsilon_1^2\,s_{m_{\varepsilon_2}}(y)}
    {\varepsilon_2^2-\varepsilon_1^2}
    $$
    for $\varepsilon_1 \neq \varepsilon_2$.

    We will substitute our expansion
    $$
        s_{m_{\varepsilon_i}}(y) = \nabla_y\log p(y) + \varepsilon_i^2 B_2(y) + \varepsilon_i^4 B_4(y)
        + \O(\varepsilon_i^6),
    $$
        and, for the sake of brevity, denote for any $y$ the clean score as $s$ and the remainder as $r_i =
        \O(\varepsilon_i^6)$.
    We get for the numerator:
    $$
    \begin{aligned}
        &\varepsilon_2^2\big(s+\varepsilon_1^2B_2+\varepsilon_1^4 B_4 + r_1\big)
        -\varepsilon_1^2\big(s+\varepsilon_2^2B_2+\varepsilon_2^4 B_4 + r_2 \big) \\
        &\quad= (\varepsilon_2^2 - \varepsilon_1^2) s
        + (\varepsilon_2^2 \varepsilon_1^2 - \varepsilon_1^2\varepsilon_2^2) B_2
        + (\varepsilon_2^2\varepsilon_1^4 - \varepsilon_1^2\varepsilon_2^4) B_4
        + \varepsilon_2^2 r_1 - \varepsilon_1^2 r_2.
    \end{aligned}
    $$
    The $B_2$ terms cancel. Using $\varepsilon_2^2\varepsilon_1^4-\varepsilon_1^2\varepsilon_2^4 =
    -\varepsilon_1^2\varepsilon_2^2(\varepsilon_2^2-\varepsilon_1^2)$, we get for the numerator:
    $$
    \varepsilon_2^2 s_{m_{\varepsilon_1}}(y) - \varepsilon_1^2 s_{m_{\varepsilon_2}}(y)
    = (\varepsilon_2^2-\varepsilon_1^2)\Big(s(y) - \varepsilon_1^2\varepsilon_2^2 B_4(y)\Big)
    + \varepsilon_2^2 r_1 - \varepsilon_1^2 r_2.
    $$

    Dividing by $\varepsilon_2^2-\varepsilon_1^2$:
    $$
        s_{\mathrm{RE}}(y) = s(y)-B_4(y)\varepsilon_1^2\varepsilon_2^2 + \frac{\varepsilon_2^2 r_1-\varepsilon_1^2
            r_2}{\varepsilon_2^2-\varepsilon_1^2}.
    $$

    Since $r_i=O(\varepsilon_i^6)$,
    $$
    \varepsilon_2^2 r_1-\varepsilon_1^2 r_2 = O(\varepsilon_2^2\varepsilon_1^6+\varepsilon_1^2\varepsilon_2^6),
    $$
    hence
    $$
    s_{\mathrm{RE}}(y) = \nabla_y\log p(y) - B_4(y)\varepsilon_1^2\varepsilon_2^2
    + \O\left(
    \frac{\varepsilon_2^2\varepsilon_1^6+\varepsilon_1^2\varepsilon_2^6}{|\varepsilon_2^2-\varepsilon_1^2|}
    \right).
    $$

    In particular: let $\varepsilon_1=a \varepsilon, \quad \varepsilon_2=b \varepsilon, \quad a, b>0,
    a \neq b.$ Then: $\varepsilon_1^2 \varepsilon_2^2=a^2 b^2 \varepsilon^4 =
    \O\left(\varepsilon^4\right)$ which means:
    $$\frac{\varepsilon_2^2 \varepsilon_1^6+\varepsilon_1^2 \varepsilon_2^6}
    {\left|\varepsilon_2^2-\varepsilon_1^2\right|}=\frac{b^2 a^6+a^2 b^6}{\left|b^2-a^2\right|}
    \varepsilon^6 = \O\left(\varepsilon^6\right).$$

    Thus we arrive at, for any $\varepsilon \ll 1$:
    $$
        s_{\mathrm{RE}}(y)=\nabla_y\log p(y)+ \O(\varepsilon^4).
    $$

    Accordingly, the estimator version is
    $$
        \hat s_{\mathrm{RE}}(y)
        =
        \frac{\varepsilon_2^2 \hat s_{\varepsilon_1}(y)-\varepsilon_1^2 \hat s_{\varepsilon_2}(y)}
        {\varepsilon_2^2-\varepsilon_1^2},
    $$
    with
    $$
    \hat s_{\mathrm{RE}}(y) = \nabla_y\log p(y)+\O(\varepsilon_1^2\varepsilon_2^2)+\text{estimation error}.
    $$

\todo{modified: Richardson variance and noise-level ratio guidance from the rebuttal.}
    At a fixed observation $y$, let $e_i=\hat s_{\varepsilon_i}(y)-s_{m_{\varepsilon_i}}(y)$ have finite conditional
    covariance matrices $V_i=\operatorname{Cov}(e_i\mid y)$ and
    $C_{12}=\operatorname{Cov}(e_1,e_2\mid y)$, taken over the randomness in the score estimates. Then
    \begin{equation}
        \operatorname{Cov}\bigl(\hat s_{\mathrm{RE}}(y)\mid y\bigr)
        =\frac{\varepsilon_2^4 V_1+\varepsilon_1^4 V_2
        -\varepsilon_1^2\varepsilon_2^2(C_{12}+C_{12}^{\top})}
        {(\varepsilon_2^2-\varepsilon_1^2)^2}.
        \label{eq:richardson_variance}
    \end{equation}
    The weights are $r^2/(r^2-1)$ and $-1/(r^2-1)$ for $r=\varepsilon_2/\varepsilon_1$;
    fixing $r$ away from one controls variance amplification.

    \clearpage
    \subsection{Appendix for Parameter Estimation}
    \label{app:param}

    \subsubsection{Theoretical Appendix}

    This subsection collects the formal results supporting the estimation criterion in
    Sec.~\ref{sec:param_estimation}, specialized to the generalized Gaussian case.

    For reference, the population criterion from Eq.~\ref{eq:param_optim} can be
    written as
    \begin{equation}
        J_0(\beta,\lambda,\Sigma)
        =
        \EE_{Y\sim m}\left[\left\|s_m(Y)-\lambda G(Y;\beta,\Sigma)\right\|_2^2\right],
        \label{eq:param_optim_pop}
    \end{equation}
    where
    $$
        G(Y;\beta,\Sigma)
        =
        -\frac{1}{\beta}\nabla_Y^{\mathrm{PD}}
        \mathrm{ES}_{\Sigma^{-1},\beta}\big(P(\cdot\mid Y),Y\big)
    $$
    is the residual field previously defined in Eq.~\ref{eq:G(y)}.

    Throughout this subsection, we adopt the regularity condition discussed in App.~\ref{app:proof_gen_gaussian_score}:
    $\EE\!\left[\|y-X\|_{\Sigma^{-1}}^{\beta-1}\mid Y=y\right]<\infty$.
    As discussed there, this condition is satisfied whenever the data density is locally bounded near the observation
    $y$.
    
    The profiled scalar scale is the least-squares coefficient in Eq.~\ref{eq:lambda_profile}; when a finite valid set
    $\Lambda=[\lambda_{\min},\lambda_{\max}]$ is imposed, the implementation uses
    $\Pi_\Lambda(a)=\min\{\lambda_{\max},\max\{\lambda_{\min},a\}\}$.

    Theorem~\ref{thm:identifiability} is the population statement: without a trace constraint, the true parameters are
    identified only up to the standard elliptical scale equivalence; with $\Tr(\Sigma)=d$, the normalized representative
    is unique. Proposition~\ref{prop:plugin_consistency} turns this population identification into consistency of the
    plug-in noise-parameter estimator when the auxiliary score estimator and Monte Carlo posterior field are consistent.
    Corollary~\ref{cor:lambda_profile} justifies profiling $\lambda$, which is the reduction used by
    Alg.~\ref{alg:noise_param}.
    \begin{theorem}[Oracle Identifiability of the Noise Parameters]
        \label{thm:identifiability}
        Let $\Theta \Def \{(\beta \in (0, 2], \lambda > 0, \Sigma \succ 0) \}$ and let the true parameters of the
        noising distribution be denoted by $\theta^*=(\beta^*, \lambda^*, \Sigma^*)$.
        \textnormal{1)} The optimizers under the true posterior $P(\cdot \mid Y)$ of the population-level objective
        $J_0$ in Eq.~\ref{eq:param_optim_pop} include the equivalence class
        $$
        \{(\beta^*, c^{\beta^*/2} \, \lambda^*, c \, \Sigma^*) \mid c \in \R^+\}.
        $$
        In particular,
        $$
        J_0(\beta^*, c^{\beta^*/2} \, \lambda^*, c \, \Sigma^*) =
        J_0(\beta^*, \lambda^*, \Sigma^*) \equiv 0, ~ \forall c \in \R^+.
        $$

        \textnormal{2)} Assume the true equivalence class is represented by the normalized tuple
        $\theta^*=(\beta^*,\lambda^*,\Sigma^*)$ with
        $$\Tr(\Sigma^*) = d,$$
        that is $\Theta_d=\{(\beta, \lambda, \Sigma): \beta \in(0,2], \lambda>0, \Sigma \succ 0,
        \operatorname{Tr}(\Sigma)=d\}$. Further, assume \textit{convolution identifiability} for the clean data
        distribution $p$: for every $\theta \in \Theta_d$, if $(p * f_\theta)(y) = 0$ for
        $m$-a.e. $y$, then $f_\theta(u)=0$ for (Lebesgue-)a.e. $u$.
        Here,
        $$
        f_\theta = q_{\theta^*} (\nabla_u \Phi_{\theta}(u)-\nabla_u \Phi_{\theta^*}(u)), \qquad
        \Phi_{\theta}(u) \Def \frac{\lambda}{\beta}\|u\|_{\Sigma^{-1}}^\beta,
        $$
        where $\Phi_{\theta} = c \gE = \frac{\lambda}{\beta} \,\gE$, and $\gE$ is the potential of the generalized Gaussian.

        Then, the restricted population optimization problem with $\Tr(\Sigma)=d$ admits a unique restricted optimizer,
        namely the representative parameter tuple $(\beta^*, \lambda^*, \Sigma^*)$.
    \end{theorem}
    The proof is given in App.~\ref{app:proof_identifiability}. It first records the scale orbit of equivalent
    generalized-Gaussian parameterizations and then uses the trace constraint plus convolution identifiability to pin
    down the normalized representative.

    \begin{proposition}[Consistency of the Plug-in Estimator]
        \label{prop:plugin_consistency}
        Let $\theta=(\beta,\lambda,\Sigma)$ denote the noise parameters and let $\Theta\subset\Theta_d$ be a compact
        subset of the restricted parameter space containing the true parameter tuple
        $\theta^*=(\beta^*,\lambda^*,\Sigma^*)$.
        Assume that the posterior term in the criterion is correctly specified, i.e., it uses $P(\cdot\mid Y=y)$, and
        use $G_{\beta,\Sigma}(y)$ defined in Eq.~\ref{eq:G(y)}.
        For $Y_1,\ldots,Y_n\iid m$, let
        $$
        J_n(\theta)\Def\frac{1}{n}\sum_{i=1}^n
        \left\|s_m(Y_i)-\lambda G_{\beta,\Sigma}(Y_i)\right\|_2^2 \quad \text{, and} ~
        \widehat J_n(\theta)\Def\frac{1}{n}\sum_{i=1}^n
        \left\|\widehat s_{m,n}(Y_i)-\lambda G_{\beta,\Sigma}(Y_i)\right\|_2^2
        $$
        be the sample-based versions of the population criterion $J_0$ in Eq.~\ref{eq:param_optim_pop} based on the
        oracle and the plug-in score estimator $\widehat s_{m,n}$, respectively.
        Assume $J_0$ is continuous on $\Theta$, and that $J_n$ is uniformly consistent:
        $$
        \sup_{\theta\in\Theta} \left|J_n(\theta)-J_0(\theta)\right| \convp 0.
        $$
        For the score terms, assume that $\widehat s_{m,n}$ is empirically $L_2$-consistent on the evaluation sample:
        $$
        \frac1n\sum_{i=1}^n \left\| \widehat s_{m,n}(Y_i)-s_m(Y_i) \right\|_2^2 = o_p(1),
        $$
        and the oracle residual w.r.t. the Energy-Tweedie term is empirically bounded in the second moment:
        $$
            \frac1n\sum_{i=1}^n \sup_{\theta\in\Theta}
            \left\| s_m(Y_i)-\lambda G_{\beta,\Sigma}(Y_i) \right\|_2^2 = O_p(1).
        $$
        By Theorem~\ref{thm:identifiability}, $\theta^*$ is the unique minimizer of $J_0$ on $\Theta$,
        thus if $\widehat\theta_n$ is any approximate minimizer satisfying
        $$
        \widehat J_n(\widehat\theta_n) \le \inf_{\theta\in\Theta} \widehat J_n(\theta) + o_p(1),
        $$
        then $$\widehat\theta_n \convp \theta^*.$$
        Now suppose that $G_{\beta,\Sigma}$ is replaced by a Monte Carlo estimate
        $\widehat G_{K_n,\beta,\Sigma}$ computed from $K_n$ posterior samples per evaluation point. If
        $$
        \frac{1}{n} \sum_{i=1}^n \sup_{\theta\in\Theta}
        \lambda^2\left\|\widehat G_{K_n,\beta,\Sigma}(Y_i)-G_{\beta,\Sigma}(Y_i)\right\|_2^2 = o_p(1),
        $$
        then with the corresponding Monte Carlo criterion:
        $$
            \widehat J_{n,K_n}^{\mathrm{MC}}(\beta,\lambda,\Sigma) \Def \frac1n \sum_{i=1}^n
            \left\| \widehat s_{m,n}(Y_i)-\lambda \widehat G_{K_n,\beta,\Sigma}(Y_i) \right\|_2^2,
        $$
        the same conclusion holds for its approximate minimizers:
        $\widehat\theta^{\mathrm{MC}}_{n,K_n} \convp \theta^*.$
    \end{proposition}
    The proof is given in App.~\ref{app:proof_plugin_consistency}. 

    \begin{corollary}[$\lambda$ Profile Closed-Form]
        With $G(y;\beta,\Sigma)$ defined in Eq.~\ref{eq:G(y)}, for every fixed $\beta$ and $\Sigma$ with
        $\EE_{Y\sim m}[\|G(Y; \beta, \Sigma)\|_2^2]>0$, the unconstrained optimizer for $\lambda$ in
        Eq.~\ref{eq:param_optim_pop} is given by Eq.~\ref{eq:lambda_profile}. If this value is positive, it is also the
        optimizer under the constraint $\lambda>0$; otherwise the constrained infimum is approached as
        $\lambda\downarrow0$. For a finite valid interval $\Lambda=[\lambda_{\min},\lambda_{\max}]$, the constrained
        optimizer is the projected coefficient $\Pi_\Lambda[\lambda^*(\beta,\Sigma)]$.
        Furthermore, in the Gaussian case this simplifies to:
        \begin{equation}
            \lambda^*(\beta=2, \Sigma) =
            \frac{\EE_{Y \sim m}\left[\langle s_m(Y), \Sigma^{-1}(\EE[X \mid Y]-Y) \rangle\right]}
            {\EE_{Y \sim m}\left[\|\Sigma^{-1}(\EE[X \mid Y]-Y)\|^2_2\right]},
            \label{eq:lambda_profile_gaussian}
        \end{equation}
        and in the isotropic case $\Sigma = \sigma^2 I_d$ this further simplifies to:
        \begin{equation}
            \lambda^*(\beta=2, \sigma^2 I_d) =
            \sigma^2\,
            \frac{\EE_{Y \sim m}\left[\langle s_m(Y), \EE[X \mid Y]-Y \rangle\right]}
            {\EE_{Y \sim m}\left[\|\EE[X \mid Y]-Y\|^2_2\right]}.
            \label{eq:lambda_profile_isotropic}
        \end{equation}
        In words, this is the global least-squares projection coefficient of the noisy score onto the residual field
        induced by the fixed $(\beta,\Sigma)$.
        \label{cor:lambda_profile}
    \end{corollary}
    The proof is given in App.~\ref{app:proof_lambda_profile}. 

    The algorithm below combines these ingredients in the form used in the estimation experiment: $\Sigma$ is
    parameterized with trace normalization, $\lambda$ is profiled by Cor.~\ref{cor:lambda_profile} and projected to the
    valid interval $\Lambda$, and the remaining shape parameters are updated by gradient steps on the profiled
    objective.

    \begin{algorithm}[ht!]
    \caption{\textbf{Profiled Noise Parameter Estimation}}
    \label{alg:noise_param}
    \begin{algorithmic}[1]
        \REQUIRE Estimation samples $Y_1,\ldots,Y_n\sim m$, fixed posterior sampler
        $P_\theta(\cdot\mid Y)$, fixed score estimator $\widehat s_m$, lower tail bound
        $\beta_{\min}>0$, scale interval $\Lambda=[\lambda_{\min},\lambda_{\max}]$,
        posterior sample count $K$, step sizes $\eta_t$.

        \STATE Parameterize
        $$
        \beta_\zeta
        =
        \beta_{\min}+(2-\beta_{\min})\operatorname{sigmoid}(\zeta),
        \qquad
        \Sigma_\phi
        =
        d\,\frac{L_\phi L_\phi^\top}{\Tr(L_\phi L_\phi^\top)},
        $$
        where $L_\phi$ is lower triangular with positive diagonal.

        \STATE Initialize $(\zeta,\phi)$.

        \WHILE{not converged}
            \STATE Sample a minibatch $\mathcal I_t\subset\{1,\ldots,n\}$ and set
            $\beta=\beta_\zeta$, $\Sigma=\Sigma_\phi$.

            \STATE For each $i\in\mathcal I_t$, draw
            $X_i^{(1)},\ldots,X_i^{(K)}\sim P_\theta(\cdot\mid Y_i)$ and compute
            $$
            \widehat G_{i,K}(\beta,\Sigma)
            =
            \Sigma^{-1}\frac1K\sum_{k=1}^K
            \|X_i^{(k)}-Y_i\|_{\Sigma^{-1}}^{\beta-2}
            (X_i^{(k)}-Y_i).
            $$

            \STATE Profile the scalar scale:
            $$
            \widehat\lambda_t
            =
            \Pi_\Lambda
            \left[
            \frac{
            \sum_{i\in\mathcal I_t}
            \langle \widehat s_m(Y_i),\widehat G_{i,K}(\beta,\Sigma)\rangle
            }{
            \sum_{i\in\mathcal I_t}
            \|\widehat G_{i,K}(\beta,\Sigma)\|_2^2
            }
            \right],
            $$
            where $\Pi_\Lambda$ is the projection onto the interval $\Lambda$.
            \STATE Take a gradient step on the profiled objective $\hat J_n^{\mathrm{prof}}(\beta,\Sigma) = \hat J_n \left( \beta, \hat\lambda_t, \Sigma \right)$:
            $$
            (\zeta,\phi)
            \leftarrow
            (\zeta,\phi)
            -
            \eta_t
            \nabla_{\zeta,\phi}
            \frac1{|\mathcal I_t|}
            \sum_{i\in\mathcal I_t}
            \left\|
            \widehat s_m(Y_i)
            -
            \widehat\lambda_t\widehat G_{i,K}(\beta_\zeta,\Sigma_\phi)
            \right\|_2^2.
            $$
        \ENDWHILE

        \STATE Set $\widehat\beta=\beta_\zeta$ and $\widehat\Sigma=\Sigma_\phi$.

        \STATE Recompute $\widehat G_{i,K}(\widehat\beta,\widehat\Sigma)$ on all estimation samples and set
        $$
        \widehat\lambda
        =
        \Pi_\Lambda
        \left[
        \frac{
        \sum_{i=1}^n
        \langle \widehat s_m(Y_i),\widehat G_{i,K}(\widehat\beta,\widehat\Sigma)\rangle
        }{
        \sum_{i=1}^n
        \|\widehat G_{i,K}(\widehat\beta,\widehat\Sigma)\|_2^2
        }
        \right].
        $$

        \RETURN $(\widehat\beta,\widehat\lambda,\widehat\Sigma)$.
    \end{algorithmic}
    \end{algorithm}

\newpage
    \subsubsection{Extra Empirical Results}
    \label{app:param_extra_results}

    We present the full  results of the experiment in Tab.~\ref{tab:param_estimation_SPD} in Tab.~\ref{tab:param_full_spd_family_controls}. To recap:
    the correctly specified family estimates $(\beta,\lambda,u,v)$;
    the Gaussian control fixes $\beta=2$ while retaining the full learned SPD matrix; the
    isotropic generalized control fixes $u=v=0$ while estimating $(\beta,\lambda)$; and the isotropic Gaussian  control imposes both restrictions. Every family is crossed with the four score/posterior source combinations on the same paired estimation and held-out observations. The truth is $(\beta^*,\lambda^*,u^*,v^*)=(1.4,1.8,0.25,0.35)$, and all held-out errors use the same oracle noisy score.
    As in Table~\ref{tab:param_estimation_SPD}, the ``oracle'' score and posterior are the importance-sampled ground truth, and the ``learned'' score and posterior
    are the Hyvärinen estimator and the SPD-conditioned Engressor.
    The values are estimated from $384$ estimation and $384$ held-out observations per noise level, pooled over the three noise levels (totaling $1152$ observations each), with $256$ posterior samples per observation.
    We generate 5 replicates of the data, each paired with its own trained score estimator (in the learned-score case). For the learned posterior entries, each data replicate is additionally fitted with each of $3$ separately trained posterior models.
    Score estimates and posterior draws are generated once per observation before the fit and held fixed during the optimization, so the plug-in objective is deterministic given the data replicate and the same samples serve every fitted family.
\todo{modified}

    \begin{table*}[ht!]
    \centering
    \setlength{\tabcolsep}{3pt}
    \resizebox{\textwidth}{!}{%
    \begin{tabular}{@{}llccccccc@{}}
        \toprule
        Fitted family & \shortstack{Score/\\Posterior} & $\widehat\beta$ & $\widehat\lambda$ & $\widehat u$
            & $\widehat v$ & $\|\widehat S-S^\star\|_F$ & \shortstack{Held-out\\MSE} & NMSE \\
        \midrule
        generalized & \shortstack{oracle/\\oracle}
            & $1.4094 \pm 0.0062$ & $1.7814 \pm 0.0041$ & $0.2537 \pm 0.0043$ & $0.3505 \pm 0.0037$
            & $0.0079 \pm 0.0048$ & $0.0728 \pm 0.0049$ & $0.0061 \pm 0.0004$ \\
        generalized & \shortstack{learned/\\oracle}
            & $1.4118 \pm 0.0075$ & $1.7733 \pm 0.0145$ & $0.2493 \pm 0.0075$ & $0.3512 \pm 0.0059$
            & $0.0112 \pm 0.0055$ & $0.0737 \pm 0.0062$ & $0.0061 \pm 0.0005$ \\
        generalized & \shortstack{oracle/\\learned}
            & $1.4379 \pm 0.0111$ & $1.7952 \pm 0.0098$ & $0.2557 \pm 0.0036$ & $0.3463 \pm 0.0056$
            & $0.0118 \pm 0.0063$ & $0.1747 \pm 0.0219$ & $0.0145 \pm 0.0018$ \\
        generalized & \shortstack{learned/\\learned}
            & $1.4398 \pm 0.0120$ & $1.7875 \pm 0.0150$ & $0.2512 \pm 0.0061$ & $0.3472 \pm 0.0066$
            & $0.0118 \pm 0.0057$ & $0.1759 \pm 0.0212$ & $0.0146 \pm 0.0017$ \\
        \midrule
        \shortstack[l]{forced Gaussian\\anisotropic} & \shortstack{oracle/\\oracle}
            & $\equiv 2$ & $1.4697 \pm 0.0210$ & $0.2588 \pm 0.0087$ & $0.3466 \pm 0.0043$
            & $0.0155 \pm 0.0106$ & $0.4109 \pm 0.0215$ & $0.0342 \pm 0.0016$ \\
        \shortstack[l]{forced Gaussian\\anisotropic} & \shortstack{learned/\\oracle}
            & $\equiv 2$ & $1.4642 \pm 0.0290$ & $0.2552 \pm 0.0078$ & $0.3475 \pm 0.0086$
            & $0.0161 \pm 0.0051$ & $0.4123 \pm 0.0183$ & $0.0343 \pm 0.0014$ \\
        \shortstack[l]{forced Gaussian\\anisotropic} & \shortstack{oracle/\\learned}
            & $\equiv 2$ & $1.5132 \pm 0.0224$ & $0.2585 \pm 0.0105$ & $0.3452 \pm 0.0062$
            & $0.0185 \pm 0.0115$ & $0.4813 \pm 0.0266$ & $0.0400 \pm 0.0022$ \\
        \shortstack[l]{forced Gaussian\\anisotropic} & \shortstack{learned/\\learned}
            & $\equiv 2$ & $1.5076 \pm 0.0300$ & $0.2548 \pm 0.0086$ & $0.3463 \pm 0.0076$
            & $0.0161 \pm 0.0080$ & $0.4824 \pm 0.0237$ & $0.0401 \pm 0.0020$ \\
        \midrule
        \shortstack[l]{forced isotropic\\generalized} & \shortstack{oracle/\\oracle}
            & $1.1206 \pm 0.0240$ & $1.8078 \pm 0.0154$ & $\equiv 0$ & $\equiv 0$
            & $0.6083 \pm 0.0000$ & $1.5910 \pm 0.0342$ & $0.1324 \pm 0.0025$ \\
        \shortstack[l]{forced isotropic\\generalized} & \shortstack{learned/\\oracle}
            & $1.1262 \pm 0.0208$ & $1.7979 \pm 0.0156$ & $\equiv 0$ & $\equiv 0$
            & $0.6083 \pm 0.0000$ & $1.5931 \pm 0.0337$ & $0.1325 \pm 0.0024$ \\
        \shortstack[l]{forced isotropic\\generalized} & \shortstack{oracle/\\learned}
            & $1.1241 \pm 0.0200$ & $1.8183 \pm 0.0177$ & $\equiv 0$ & $\equiv 0$
            & $0.6083 \pm 0.0000$ & $1.7145 \pm 0.0495$ & $0.1426 \pm 0.0034$ \\
        \shortstack[l]{forced isotropic\\generalized} & \shortstack{learned/\\learned}
            & $1.1295 \pm 0.0149$ & $1.8087 \pm 0.0178$ & $\equiv 0$ & $\equiv 0$
            & $0.6083 \pm 0.0000$ & $1.7159 \pm 0.0501$ & $0.1428 \pm 0.0034$ \\
        \midrule
        \shortstack[l]{forced isotropic\\Gaussian} & \shortstack{oracle/\\oracle}
            & $\equiv 2$ & $1.3650 \pm 0.0332$ & $\equiv 0$ & $\equiv 0$
            & $0.6083 \pm 0.0000$ & $2.5008 \pm 0.0270$ & $0.2081 \pm 0.0019$ \\
        \shortstack[l]{forced isotropic\\Gaussian} & \shortstack{learned/\\oracle}
            & $\equiv 2$ & $1.3591 \pm 0.0381$ & $\equiv 0$ & $\equiv 0$
            & $0.6083 \pm 0.0000$ & $2.5043 \pm 0.0304$ & $0.2084 \pm 0.0023$ \\
        \shortstack[l]{forced isotropic\\Gaussian} & \shortstack{oracle/\\learned}
            & $\equiv 2$ & $1.4005 \pm 0.0343$ & $\equiv 0$ & $\equiv 0$
            & $0.6083 \pm 0.0000$ & $2.5634 \pm 0.0746$ & $0.2133 \pm 0.0050$ \\
        \shortstack[l]{forced isotropic\\Gaussian} & \shortstack{learned/\\learned}
            & $\equiv 2$ & $1.3946 \pm 0.0392$ & $\equiv 0$ & $\equiv 0$
            & $0.6083 \pm 0.0000$ & $2.5667 \pm 0.0775$ & $0.2135 \pm 0.0053$ \\
        \bottomrule
    \end{tabular}%
    }
    \caption{\textbf{Estimated noise parameters, shape error, and held-out score error for four fitted families and four score/posterior source combinations.} Setup and errors are the same as in Tab.~\ref{tab:param_estimation_SPD}.
    $\widehat S=S(\widehat u,\widehat v)$ and $S^\star=S(u^\star,v^\star)$ are the fitted and true trace-normalized shape matrices; held-out MSE is the unnormalized numerator of the NMSE.
    }
    \label{tab:param_full_spd_family_controls}
\end{table*}
     The full table confirms the conclusions of Tab.~\ref{tab:param_estimation_SPD} under every
     source combination: the correctly specified family has the lowest held-out error and
     recovers the non-diagonal geometry throughout ($\|\widehat S-S^\star\|_F=0.0079$--$0.0118$),
     and the family ordering is identical for all four source combinations.
     Replacing the oracle score with the learned Hyvärinen score changes MSE by at most $0.0035$ at a fixed family and posterior source.
    The constrained fits again compensate in characteristic ways regardless of the estimator: forcing Gaussian tails  keeps the SPD coordinates accurate but lowers $\widehat\lambda$ to $1.46$--$1.51$, while forcing isotropy drags $\widehat\beta$ down to $1.12$--$1.13$.
\todo{modified: family-control errors are presented in Table~\ref{tab:param_full_spd_family_controls}.}

    \clearpage
    \subsubsection{Proofs of Section~\ref{sec:param_estimation}}
    \label{app:proof_param}
    The next few subsections give the proofs for the formal statements above.

    \paragraph{Proof of Theorem~\ref{thm:identifiability}}
    \label{app:proof_identifiability}

    \textit{Part 1.}
    Similar to what we did in App.~\ref{app:proof_richardson}, we first use scale invariance. For any $c>0$, we claim that
    $$
    \mathrm{ES}_{(c\Sigma)^{-1},\beta}(P,y) = c^{-\beta/2}\,\mathrm{ES}_{\Sigma^{-1},\beta}(P,y),
    $$
    which makes the term inside the expectation in $J_0$ pointwise invariant to rescaling of $(\lambda, \Sigma)$, hence
    the conclusion holds in expectation, as well.

    For the energy score, we have:
    \begin{align*}
        \mathrm{ES}_{(c\Sigma)^{-1},\beta}(P,y)
        &= \EE_{Y \sim P} \left[ \|Y-y\|_{(c\Sigma)^{-1}}^\beta \right]
        - \tfrac{1}{2} \EE_{Y,Y^\prime \iid P}
        \left[ \|Y-Y^\prime\|_{(c\Sigma)^{-1}}^\beta \right] \\
        &= c^{-\beta/2} \left(
        \EE_{Y \sim P} \left[ \|Y-y\|_{\Sigma^{-1}}^\beta \right]
        - \tfrac{1}{2} \EE_{Y,Y^\prime \iid P}
        \left[ \|Y-Y^\prime\|_{\Sigma^{-1}}^\beta \right]\right) \\
        &= c^{-\beta/2}\,\mathrm{ES}_{\Sigma^{-1},\beta}(P,y).
    \end{align*}

    By linearity of the path-derivative:
    $$
    \nabla_y^{\mathrm{PD}}\mathrm{ES}_{(c\Sigma)^{-1},\beta}(P,y)
    = c^{-\beta/2}\nabla_y^{\mathrm{PD}}\mathrm{ES}_{\Sigma^{-1},\beta}(P,y), ~~ \forall y,
    $$
    therefore:
    $$
        \frac{c^{\beta/2}\lambda}{\beta}\nabla_y^{\mathrm{PD}}\mathrm{ES}_{(c\Sigma)^{-1},\beta}(P,y)
        = \frac{\lambda}{\beta}\nabla_y^{\mathrm{PD}}\mathrm{ES}_{\Sigma^{-1},\beta}(P,y).
    $$

    As $s_m(y)$ does not depend on our belief about the parameters, this means that the term inside the expectation in
    $J_0$ is pointwise invariant to this rescaling of $(\lambda, \Sigma)$, hence the expectation is, as well:
    $$
    J_0(\beta, c^{\beta/2}\lambda, c\Sigma)=J_0(\beta, \lambda, \Sigma).
    $$

    By Theorem~\ref{thm:et}, at the true tuple $(\beta^*, \lambda^*, \Sigma^*)$, the vector inside $J_0$ is
    zero $m$--\textit{a.e}. Therefore, $J_0(\beta^*, \lambda^*, \Sigma^*)=0$. Since $J_0\ge 0$, the true tuple is a
    global optimizer. By scale invariance, the whole orbit is also made of global optimizers. This proves part 1.

    \textit{Part 2.}
    As noted, we are assuming that the path derivatives in $J_0$ are finite for all $\theta$ in the parameter class
    under consideration; in particular, the local boundedness condition discussed in
    App.~\ref{app:proof_gen_gaussian_score} is sufficient for the singularity at $X=y$.

    First, impose the restriction:
    $$
        \Tr(\Sigma) = d,
    $$
    where $d$ is the dimension of the ambient space. As $\Tr( c \Sigma) = c \Tr(\Sigma)$, this uniquely pins down the
    constant $c$ in the above equivalence class.

    For a parameter tuple $\theta=(\beta, \lambda, \Sigma)$, introduce the generalized-Gaussian potential and its
    gradient:
    $$
    \Phi_{\theta}(u) \;\Def\; \frac{\lambda}{\beta}\|u\|_{\Sigma^{-1}}^\beta, \qquad
    \Psi_{\theta}(u) \;\Def\; \nabla_u \Phi_{\theta}(u)
    = \lambda \Sigma^{-1}\|u\|_{\Sigma^{-1}}^{\beta-2}u.
    $$

    Under the equivalence-class transformation $(\lambda,\Sigma)\mapsto(c^{\beta/2}\lambda,c\Sigma)$, write
    $\theta_c\Def(\beta, c^{\beta/2}\lambda, c\Sigma)$. Then
    $$
        \|u\|_{(c\Sigma)^{-1}}^\beta
        = \big(u^\top(c\Sigma)^{-1}u\big)^{\beta/2}
        = c^{-\beta/2}\|u\|_{\Sigma^{-1}}^\beta,
    $$
    and hence the potential is invariant:
    $$
    \Phi_{\theta_c}(u)
    = \frac{c^{\beta/2}\lambda}{\beta}\|u\|_{(c\Sigma)^{-1}}^\beta
    = \frac{\lambda}{\beta}\|u\|_{\Sigma^{-1}}^\beta
    = \Phi_{\theta}(u).
    $$

    Note the following relations:
    \begin{enumerate}
        \item Bayes' formula:
        $$
        p(x\mid Y=y) = \frac{p(x)q_{\theta^*}(y-x)}{\int p(x)q_{\theta^*}(y-x)\,dx}
        = \frac{p(x)q_{\theta^*}(y-x)}{m(y)}.
        $$

        \item
        With $(p*h)(y)\Def \int_\X p(x)h(y-x)\,dx$,
        $$
        m(y)s_m(y)
        = m(y)\nabla_y\log m(y)
        = \nabla_y m(y)
        = (p*\nabla q_{\theta^*})(y)
        = -\left(p*\left[q_{\theta^*}\Psi_{\theta^*}\right]\right)(y).
        $$

    \end{enumerate}

    Denote $\theta^* \Def (\beta^*, \lambda^*, \Sigma^*)$ the true parameter tuple, and
    $\tilde{\theta} \Def (\tilde{\beta}, \tilde{\lambda}, \tilde{\Sigma})$ the restricted optimizer tuple.

    First, we have:
    $$
        \nabla_y q_{\theta^*}(y-x)
        = -\lambda^* (\Sigma^*)^{-1}\|y-x\|_{(\Sigma^*)^{-1}}^{\beta^*-2}(y-x)q_{\theta^*}(y-x)
        = -q_{\theta^*}(y-x)\Psi_{\theta^*}(y-x).
    $$
    Thus, by Theorem~\ref{thm:et}, under the true posterior and the correct parameters,
    $$
        m(y) \frac{\lambda^*}{\beta^*} \nabla_y^{\mathrm{PD}}
        \mathrm{ES}_{(\Sigma^*)^{-1}, \beta^*}(P(\cdot \mid Y=y), y)
            =
    - (p * \nabla q_{\theta^*})(y)
    = \left(p * \left[q_{\theta^*}\Psi_{\theta^*}\right]\right)(y).
    $$

    We \textit{claim} that under $\tilde{\theta}=(\tilde{\beta}, \tilde{\lambda}, \tilde{\Sigma})$,
    $$
    m(y) \frac{\tilde{\lambda}}{\tilde{\beta}}
    \nabla_y^{\mathrm{PD}}\mathrm{ES}_{\tilde{\Sigma}^{-1},\tilde{\beta}}(P(\cdot\mid Y=y),y)
    = \left(p * \left[q_{\theta^*}\Psi_{\tilde{\theta}}\right]\right)(y)
    = \int_\X p(x)q_{\theta^*}(y-x)\Psi_{\tilde{\theta}}(y-x)\,dx.
    $$

    By the same derivation as Thm.~\ref{thm:et}, we have:
    $$
    \begin{aligned}
        \frac{\tilde{\lambda}}{\tilde{\beta}}
        \nabla_y^{\mathrm{PD}}\mathrm{ES}_{\tilde{\Sigma}^{-1},\tilde{\beta}}(P(\cdot\mid Y=y),y)
        &= -\tilde{\lambda}\tilde{\Sigma}^{-1}
        \EE_{X \sim P(\cdot\mid Y=y)}
        \left[\|X-y\|_{\tilde{\Sigma}^{-1}}^{\tilde{\beta}-2}(X-y)\right] \\
        &= \EE_{X \sim P(\cdot\mid Y=y)}\left[\Psi_{\tilde{\theta}}(y-X)\right].
    \end{aligned}
    $$

    Using Bayes' formula, this is equal to the following under the \textit{true} posterior:
    $$
        \int_{\X}\Psi_{\tilde{\theta}}(y-x)\frac{p(x)q_{\theta^*}(y-x)}{m(y)}\,dx.
    $$

    Hence:
    $$
    m(y)\frac{\tilde{\lambda}}{\tilde{\beta}}
    \nabla_y^{\mathrm{PD}}\mathrm{ES}_{\tilde{\Sigma}^{-1},\tilde{\beta}}(P(\cdot\mid Y=y),y)
    = \int_\X p(x)
    \underbrace{\left[q_{\theta^*}(y-x)\Psi_{\tilde{\theta}}(y-x)\right]}_{h_{\theta^*,\tilde{\theta}}(y-x)}
    \,dx.
    $$

    Thus, by the definition of (vector-valued) convolution:
    $$
    m(y)\frac{\tilde{\lambda}}{\tilde{\beta}}
    \nabla_y^{\mathrm{PD}}\mathrm{ES}_{\tilde{\Sigma}^{-1},\tilde{\beta}}(P(\cdot\mid Y=y),y)
    = \left(p * \left[q_{\theta^*}\Psi_{\tilde{\theta}}\right]\right)(y).
    $$

    Note that:
    $$
        m(y)s_m(y)
        = \nabla_y m(y)
        = (p*\nabla q_{\theta^*})(y)
        = -\left(p*\left[q_{\theta^*}\Psi_{\theta^*}\right]\right)(y).
    $$

    We wish to connect to the term inside the integrand in Eq.~\ref{eq:param_optim_pop}. By linearity of convolution
    applied component-wise:
    $$
        m(y)\left[
        s_m(y)+\frac{\tilde{\lambda}}{\tilde{\beta}}
        \nabla_y^{\mathrm{PD}}\mathrm{ES}_{\tilde{\Sigma}^{-1},\tilde{\beta}}(P(\cdot\mid Y=y),y)
        \right]
        = \left(p*\left[q_{\theta^*}\left(\Psi_{\tilde{\theta}}-\Psi_{\theta^*}\right)\right]\right)(y).
    $$

    We know by Thm.~\ref{thm:et} and the first part of this proof that the normalized true tuple
    $\theta^*=(\beta^*, \lambda^*, \Sigma^*)$ is feasible for the restricted problem and satisfies $J_0(\theta^*) = 0$.
    Hence the restricted minimum value is \textit{at most} 0, meaning it is exactly zero (as $J_0 \geq 0$) and every
    restricted optimizer satisfies $J_0(\tilde{\theta})=0$ by definition. Hence the bracket on the LHS vanishes
    $m$--a.e.

    Thus:
    $$
        \left(p*\left[q_{\theta^*}\left(\Psi_{\tilde{\theta}}-\Psi_{\theta^*}\right)\right]\right)(y)
        = 0, ~~ m\text{--a.e.}
    $$

    Next, impose \textbf{the convolutional identifiability assumption}, i.e., that the data distribution is ``complete''
    under convolution, namely:
    $$
        p*f \stackrel{m-\mathrm{a.e.}}{=} 0
        \implies f \stackrel{\mathrm{a.e.}}{=} 0
    $$
    for $f_\theta = q_{\theta^*} (\Psi_{\tilde{\theta}}-\Psi_{\theta^*})$.

    As a side note: a stronger sufficient condition is given by the Fourier convolution theorem: the Fourier transform
    of $p$ -- i.e., the data distribution's \textit{characteristic function} -- has no zeros:
    $$
        \left( \widehat{p}(\omega)~\widehat{f}(\omega)=0 \implies \widehat{f}(\omega) = 0 \right)
        \Leftrightarrow \widehat{p} (\omega)\neq 0, ~~ \forall \omega.
    $$

    Continuing: as the Generalized Gaussian density $q_{\theta^*}$ is strictly positive everywhere, this must mean that:
    $$
        \left(\Psi_{\tilde{\theta}}-\Psi_{\theta^*}\right)(y) = 0, ~~ m\text{--a.e.}.
    $$
    As $p$ is a probability density and $q_{\theta^*} >0$ everywhere, this means that $m$-a.e. implies (Lebesgue) a.e.,
    hence:
    $$
        \Psi_{\tilde{\theta}} ~\stackrel{a.e.}{=} ~ \Psi_{\theta^*}
    $$

    We thus have so far:
    \begin{enumerate}
        \item The restricted optimizer achieves the global minimum loss $J_0 = 0$
        \item This leads to a.e. equality of $\Psi_{\tilde{\theta}}$ and $\Psi_{\theta^*}$, as proven above.
    \end{enumerate}

    We will show that further:
    \begin{enumerate}
        \item The above a.e. equality implies that $\tilde\theta$ lies in the same (optimal) scale orbit as $\theta^*$.
        \item The trace restriction $\Tr(\Sigma)=d=\Tr(\Sigma^*)$ uniquely identifies the member of the orbit,
        which is moreover $\tilde\theta=\theta^*$.
    \end{enumerate}

    Reminder: for $\theta=(\beta,\lambda,\Sigma)$,
    $$
        \Psi_\theta(u) \Def \lambda \Sigma^{-1}\|u\|_{\Sigma^{-1}}^{\beta-2}u.
    $$
    Thus $\Psi_\theta$ is a continuous deterministic vector-valued function on $\R^d\setminus\{0\}$. Since the two sides
    are continuous on this set, the a.e. equality implies, for every $u \neq 0$,
    $$
        \tilde{\lambda}\tilde{\Sigma}^{-1}\|u\|_{\tilde{\Sigma}^{-1}}^{\tilde{\beta}-2}u
        =
        \lambda^*(\Sigma^*)^{-1}\|u\|_{(\Sigma^*)^{-1}}^{\beta^*-2}u.
    $$

    The scalar radial factors are positive, so the vector parts $\tilde{\Sigma}^{-1}u$ and
    $(\Sigma^*)^{-1}u$ must be collinear.
    Pre-multiplying by $\Sigma^*$ and writing $B \Def \Sigma^*\tilde{\Sigma}^{-1}$ gives
    $$
        \tilde{\lambda}\|u\|_{\tilde{\Sigma}^{-1}}^{\tilde{\beta}-2}Bu
        =
        \lambda^*\|u\|_{(\Sigma^*)^{-1}}^{\beta^*-2}u.
    $$
    Hence $Bu$ is a positive scalar multiple of $u$ for every $u\neq 0$. We also have $B e_i = a_i e_i$ for a basis
    $\{e_i\}$ and constants $a_i$, and that
    $B(e_i + e_j)$ must be parallel to $e_i + e_j$. Thus $a_i = a_j$ for all $i, j$, and hence $B = b I$ for some $b>0$.
    Equivalently, for $c \Def b^{-1}>0$,
    $$
        \tilde{\Sigma}=c\Sigma^*.
    $$

    Substituting $\tilde{\Sigma}=c\Sigma^*$ into the full identity above gives
    $$
        \tilde{\lambda}c^{-\tilde{\beta}/2}
        \|u\|_{(\Sigma^*)^{-1}}^{\tilde{\beta}-2}(\Sigma^*)^{-1}u
        =
        \lambda^*\|u\|_{(\Sigma^*)^{-1}}^{\beta^*-2}(\Sigma^*)^{-1}u.
    $$
    Since $\Sigma^*$ is positive definite and $u\neq 0$, we can cancel $(\Sigma^*)^{-1}u$. Defining the Mahalanobis
    radius
    $$
        r \Def \sqrt{u^\top(\Sigma^*)^{-1}u},
    $$
    we obtain the scalar relation
    $$
        \tilde{\lambda}c^{-\tilde{\beta}/2}r^{\tilde{\beta}-2}
        =
        \lambda^*r^{\beta^*-2}, \quad \text{for all } r>0.
    $$

    This identity can hold for all $r>0$ only if the powers of $r$ agree:
    $$
        \tilde{\beta}-2=\beta^*-2,
    $$
    hence $\tilde{\beta}=\beta^*$. Plugging back in gives
    $$
        \tilde{\lambda}c^{-\beta^*/2}=\lambda^*,
    $$
    or equivalently $\tilde{\lambda}=c^{\beta^*/2}\lambda^*$. Therefore,
    $$
        \tilde{\theta}=(\beta^*,c^{\beta^*/2}\lambda^*,c\Sigma^*),
    $$
    so the restricted optimizer $\tilde{\theta}$ lies in the same optimal orbit as $\theta^*$. Finally, since
    $\Tr(\tilde{\Sigma})=d$ and $\Tr(\Sigma^*)=d$, we have $cd=d$, hence $c=1$ and
    $\tilde{\theta}=\theta^*$.
    Thus the normalized true tuple is the unique restricted optimizer. Thus, restricting the scale of the ``covariance''
    uniquely pins down the minimizer as it pins down the constant $c$ above. This also proves that the above minimizer
    equivalence class is the only one for Eq.~\ref{eq:param_optim_pop}, otherwise we'd not be able to obtain
    \textit{the} unique minimizer (i.e., by contradiction). Furthermore, this restricted unique minimizer coincides with
    the true parameter tuple or the normalized representative.

    \qed

    \paragraph{Proof of Proposition~\ref{prop:plugin_consistency}}
    \label{app:proof_plugin_consistency}

    Reminder: following Corollary~\ref{cor:lambda_profile}, we have the following shorthand:
    $$
    G_{\beta,\Sigma}(y) \Def -\frac{1}{\beta} \nabla_y^{\mathrm{PD}}
    \mathrm{ES}_{\Sigma^{-1},\beta} \bigl(P(\cdot\mid Y=y),y\bigr).
    $$
    The population criterion $J_0$ can then be written as:
    $$
    J_0(\theta) = \EE_{Y\sim m} \left[ \left\| s_m(Y)-\lambda G_{\beta,\Sigma}(Y) \right\|_2^2 \right].
    $$
    Using this we can define:
    \begin{itemize}
        \item \textit{the oracle empirical criterion:}
        $$
            J_n(\beta,\lambda,\Sigma) \Def \frac1n \sum_{i=1}^n
            \left\| s_m(Y_i)-\lambda G_{\beta,\Sigma}(Y_i) \right\|_2^2,
        $$
        \item \textit{the plug-in empirical criterion:}
        $$
            \widehat J_n(\beta,\lambda,\Sigma) \Def \frac1n \sum_{i=1}^n
            \left\| \widehat s_{m,n}(Y_i)-\lambda G_{\beta,\Sigma}(Y_i) \right\|_2^2.
        $$
        \item \textit{the Monte Carlo plug-in empirical criterion}, using $K_n$ Monte Carlo samples to estimate
        $G_{\beta,\Sigma}$:
        $$
            \widehat J_{n,K_n}^{\mathrm{MC}}(\beta,\lambda,\Sigma) \Def \frac1n \sum_{i=1}^n
            \left\| \widehat s_{m,n}(Y_i)-\lambda \widehat G_{K_n,\beta,\Sigma}(Y_i) \right\|_2^2.
        $$
    \end{itemize}
    For $\theta=(\beta,\lambda,\Sigma)$, write $G_\theta(y) \Def G_{\beta,\Sigma}(y)$ and let

    $$
    e_i \Def \widehat s_{m,n}(Y_i)-s_m(Y_i), \qquad
    r_i(\theta) \Def s_m(Y_i)-\lambda G_\theta(Y_i).
    $$

    We wish to see what happens to the plug-in empirical criterion when a consistent plug-in estimator is used.
    We have:
    $$
    \widehat s_{m,n}(Y_i) - \lambda G_{\beta,\Sigma}(Y_i)
    =
    \widehat s_{m,n}(Y_i) - s_m(Y_i) + s_m(Y_i) - \lambda G_{\beta,\Sigma}(Y_i)
    =
    e_i + r_i(\theta).
    $$
    Therefore, for each $\theta=(\beta,\lambda,\Sigma)\in\Theta$,
    $$
    \begin{aligned}
    \widehat J_n(\theta)-J_n(\theta)
    &=
    \frac1n \sum_{i=1}^n
    \left[
    \left\| \widehat s_{m,n}(Y_i)-\lambda G_{\beta,\Sigma}(Y_i) \right\|_2^2
    -
    \left\| s_m(Y_i)-\lambda G_{\beta,\Sigma}(Y_i) \right\|_2^2
    \right] \\
    &=
    \frac1n \sum_{i=1}^n
    \left[
    \|r_i(\theta)+e_i\|_2^2-\|r_i(\theta)\|_2^2
    \right] \\
    &=
    \frac1n \sum_{i=1}^n
    \left(2\langle e_i,r_i(\theta)\rangle+\|e_i\|_2^2\right).
    \end{aligned}
    $$

    The proposition assumes directly that
    $$
        \sup_{\theta\in\Theta} \left|J_n(\theta) - J_0(\theta)\right| \convp 0.
    $$
    It remains to bound the plug-in error $\widehat J_n-J_n$.
    By the triangle and Cauchy--Schwarz inequalities:
    $$
    \begin{aligned}
        \sup_{\theta\in\Theta} \left|\widehat J_n(\theta)-J_n(\theta)\right|
        &=
        \sup_{\theta\in\Theta} \left| \frac1n \sum_{i=1}^n
        \left(2\langle e_i,r_i(\theta)\rangle+\|e_i\|_2^2\right) \right|
        \\
        &\leq \sup_{\theta\in\Theta} \frac1n \sum_{i=1}^n
        \left( 2\left|\langle e_i, r_i(\theta)\rangle \right| +\|e_i\|_2^2\right) \\
        &\leq
        \frac{2}{n} \sum_{i=1}^n \|e_i\|_2 \sup_{\theta\in\Theta} \|r_i(\theta)\|_2
        + \frac1n \sum_{i=1}^n \|e_i\|_2^2.
    \end{aligned}
    $$

    We have assumed that the plug-in estimator has an empirical $L_2$-consistency on the evaluation sample:
    $$
    \frac1n \sum_{i=1}^n \left\| e_i \right\|^2_2 =
    \frac1n \sum_{i=1}^n \left\| \widehat s_{m,n}(Y_i)-s_m(Y_i) \right\|_2^2 = o_p(1).
    $$
    This is an empirical analogue of Fisher-divergence consistency, the latter being common for score-matching
    estimators \citep{song_sliced_2019}.
    We leave it as a high-level assumption since its verification depends on the exact score estimator, model class,
    data regularity, and other factors.
    This means that:
    $$
    \sup_{\theta\in\Theta} \left|\widehat J_n(\theta)-J_n(\theta)\right| \leq
    \frac{2}{n} \sum_{i=1}^n \|e_i\|_2 \sup_{\theta\in\Theta} \|r_i(\theta)\|_2 + o_p(1).
    $$
    We can further handle the remaining term by applying the Cauchy--Schwarz inequality again:
    $$
    \frac{2}{n} \sum_{i=1}^n \|e_i\|_2 \sup_{\theta\in\Theta} \|r_i(\theta)\|_2 \leq
    2 \left( \frac1n \sum_{i=1}^n \|e_i\|_2^2 \right)^{1/2}
    \left( \frac1n \sum_{i=1}^n \sup_{\theta\in\Theta} \|r_i(\theta)\|_2^2 \right)^{1/2}.
    $$
    So by the same assumption, the first factor is $o_p(1)$.
    For the second term, we have assumed that it is empirically bounded in the second moment:
    $$
    \frac1n \sum_{i=1}^n \sup_{\theta\in\Theta} \left\| s_m(Y_i) - \lambda G_\theta(Y_i) \right\|_2^2
    = O_p(1).
    $$
    Thus, the whole is $o_p(1)$:
    $$\sup_{\theta\in\Theta} \left|\widehat J_n(\theta)-J_n(\theta)\right| = o_p(1).$$

    By the triangle inequality, we have:
    $$
        \sup_{\theta\in\Theta}\left|\widehat J_n(\theta)-J_0(\theta)\right|
        \leq \sup_{\theta\in\Theta}\left|\widehat J_n(\theta)-J_n(\theta)\right|
        +\sup_{\theta\in\Theta}\left|J_n(\theta)-J_0(\theta)\right|.
    $$
    The first term is $o_p(1)$ by the above, and the second term is $o_p(1)$ by assumption. Hence:
    $$
        \sup_{\theta\in\Theta} \left|\widehat J_n(\theta)-J_0(\theta) \right| = o_p(1).
    $$
    We will use Theorem 5.7 in \citet{vaart_asymptotic_1998} with $M_n(\theta) = -\widehat J_n(\theta)$ and
    $M(\theta) = -J_0(\theta)$.
    First, the last result means:
    $$
        \widehat J_n(\widehat\theta_n) \leq \inf_{\theta\in\Theta} \widehat J_n(\theta) + o_p(1)
        \leq \widehat J_n(\theta^*) + o_p(1).
    $$
    Second: let
    $$
        A_{\varepsilon}=\left\{\theta\in\Theta: d\left(\theta,\theta^*\right) \geq \varepsilon\right\}.
    $$
    As $\Theta$ is assumed compact and $J_0$ is assumed continuous, $J_0$ attains its minimum on $A_{\varepsilon}$.
    Theorem~\ref{thm:identifiability} already gives $\theta^*$ as the unique minimizer on the compact set $\Theta$.
    Hence:
    $$
    \inf_{\theta\in A_{\varepsilon}} J_0(\theta) > J_0(\theta^*).
    $$

    Thus, by Theorem 5.7 in \citet{vaart_asymptotic_1998}, we have that
    $$
    \widehat\theta_n \convp \theta^*.
    $$

    \paragraph{Monte Carlo extension:}
    We denote the Monte Carlo plug-in for $G_{\beta,\Sigma}$ as:
    $$
        \widehat G_{K_n,\beta,\Sigma}(Y_i) \Def \Sigma^{-1} \frac{1}{K_n} \sum_{k=1}^{K_n}
        \left\|X_{i,k}-Y_i\right\|_{\Sigma^{-1}}^{\beta-2}\left(X_{i,k}-Y_i\right),
    $$
    with $X_{i,k}$ being samples from the correctly specified posterior $P(\cdot\mid Y=Y_i)$, and $K_n$ denoting the Monte
    Carlo budget such that the posterior-sampling error vanishes asymptotically.
    For this, we assume a similar empirical $L_2$-consistency in the evaluation sample as we did for the other score
    estimator:
    $$
    \frac{1}{n} \sum_{i=1}^n \sup_{\theta\in\Theta}
    \lambda^2\left\|\widehat G_{K_n,\beta,\Sigma}(Y_i)-G_{\beta,\Sigma}(Y_i)\right\|_2^2 = o_p(1).
    $$
    Thus, the argument follows along similar lines to the proof above.
    Defining
    $$
    a_i(\theta) \Def \widehat s_{m,n}(Y_i) - \lambda G_{\beta,\Sigma}(Y_i), \quad
    \Delta_i(\theta) \Def \lambda \left( \widehat G_{K_n,\beta,\Sigma}(Y_i)-G_{\beta,\Sigma}(Y_i) \right).
    $$

    Then the Monte Carlo residual satisfies
    $$
    \widehat s_{m,n}(Y_i) - \lambda \widehat G_{K_n,\beta,\Sigma}(Y_i) = a_i(\theta)-\Delta_i(\theta).
    $$
    Therefore, for each $\theta\in\Theta$,
    $$
    \begin{aligned}
        \widehat J_{n,K_n}^{\mathrm{MC}}(\theta) - \widehat J_n(\theta)
        &= \frac1n \sum_{i=1}^n
        \left( \|a_i(\theta)-\Delta_i(\theta)\|_2^2 - \|a_i(\theta)\|_2^2 \right) \\
        &= \frac1n \sum_{i=1}^n
        \left( -2\langle a_i(\theta),\Delta_i(\theta)\rangle + \|\Delta_i(\theta)\|_2^2 \right).
    \end{aligned}
    $$
    Taking the supremum and applying Cauchy--Schwarz as before gives
    $$
    \begin{aligned}
        \sup_{\theta\in\Theta} \left| \widehat J_{n,K_n}^{\mathrm{MC}}(\theta) - \widehat J_n(\theta) \right|
        &\le \frac2n \sum_{i=1}^n \sup_{\theta\in\Theta} \|a_i(\theta)\|_2 \|\Delta_i(\theta)\|_2
        + \frac1n \sum_{i=1}^n \sup_{\theta\in\Theta} \|\Delta_i(\theta)\|_2^2 \\
        &\le 2 \left( \frac1n \sum_{i=1}^n \sup_{\theta\in\Theta} \|a_i(\theta)\|_2^2 \right)^{1/2}
        \left( \frac1n \sum_{i=1}^n \sup_{\theta\in\Theta} \|\Delta_i(\theta)\|_2^2 \right)^{1/2} \\
        &\qquad+ \frac1n \sum_{i=1}^n \sup_{\theta\in\Theta} \|\Delta_i(\theta)\|_2^2.
    \end{aligned}
    $$
    The second part of the first term and the last term are $o_p(1)$ by assumption, since:
    $$
        \|\Delta_i(\theta)\|_2^2 =
        \lambda^2 \left\| \widehat G_{K_n,\beta,\Sigma}(Y_i) - G_{\beta,\Sigma}(Y_i) \right\|_2^2.
    $$
    Thus, the question is what is the order of $\frac1n \sum_{i=1}^n \sup_{\theta\in\Theta} \|a_i(\theta)\|_2^2$?
    Reusing notation from the first part of the proof:
    $$
    e_i = \widehat s_{m,n}(Y_i) - s_m(Y_i), \quad
    r_i(\theta) = s_m(Y_i) - \lambda G_{\beta,\Sigma}(Y_i),
    $$
    we have:
    $$
    a_i(\theta) = \widehat s_{m,n}(Y_i) - \lambda G_{\beta,\Sigma}(Y_i) = e_i+r_i(\theta).
    $$

    Now, using the triangle inequality:
    $$
    \frac1n \sum_{i=1}^n \sup_{\theta\in\Theta} \|a_i(\theta)\|_2^2
    \le \frac2n \sum_{i=1}^n \|e_i\|_2^2
    + \frac2n \sum_{i=1}^n \sup_{\theta\in\Theta} \|r_i(\theta)\|_2^2.
    $$
    The first term is, as discussed, $o_p(1)$ by assumption; for the second term we have:
    $$
    \frac2n \sum_{i=1}^n \sup_{\theta\in\Theta} \|r_i(\theta)\|_2^2
    = \frac2n \sum_{i=1}^n \sup_{\theta\in\Theta}
    \left\|s_m(Y_i)-\lambda G_{\beta,\Sigma}(Y_i)\right\|_2^2 = O_p(1)
    $$
    by assumption on the oracle residual of the Energy-Tweedie term in the Proposition.
    Thus:
    $$
        \sup_{\theta\in\Theta} \left| \widehat J_{n,K_n}^{\mathrm{MC}}(\theta)-\widehat J_n(\theta) \right| = o_p(1).
    $$
    By the triangle inequality, we now have:
    $$
        \sup_{\theta\in\Theta} \left| \widehat J_{n,K_n}^{\mathrm{MC}}(\theta)-J_0(\theta) \right|
        \le \sup_{\theta\in\Theta}
        \left| \widehat J_{n,K_n}^{\mathrm{MC}}(\theta)-\widehat J_n(\theta) \right|
        + \sup_{\theta\in\Theta} |\widehat J_n(\theta)-J_0(\theta)| = o_p(1).
    $$
    Thus, for any $\widehat\theta^{\mathrm{MC}}_{n,K_n}$ satisfying:
    $$
    \widehat J_{n,K_n}^{\mathrm{MC}}(\widehat\theta^{\mathrm{MC}}_{n,K_n})
    \le \inf_{\theta\in\Theta} \widehat J_{n,K_n}^{\mathrm{MC}}(\theta) + o_p(1)
    $$
    the same argument as above applies, giving us:
    $$
    \widehat\theta^{\mathrm{MC}}_{n,K_n} \convp \theta^*.
    $$

    \qed

    \paragraph{Proof of Corollary~\ref{cor:lambda_profile}}
    \label{app:proof_lambda_profile}

    Starting from the population objective $J_0(\beta, \lambda, \Sigma)$, and fixing $\beta$ and $\Sigma$, define
    $$
    G(y; \beta, \Sigma)
    \Def \Sigma^{-1} ~\EE_{X \sim P(\cdot \mid Y=y)}\left[\|X-y\|_{\Sigma^{-1}}^{\beta-2}(X-y)\right].
    $$
    Then:
    $$
    \begin{aligned}
    J_0
    &= \EE_{Y \sim m} \big[ \| s_m(Y)
        + \frac{\lambda}{\beta}\nabla_Y^{\mathrm{PD}}\mathrm{ES}_{\Sigma^{-1},\beta}(P(\cdot \mid Y), Y)\|^2 \big] \\
    &= \EE_{Y \sim m} \big[ \| s_m(Y) - \lambda G(Y; \beta, \Sigma) \|^2 \big] \\
    &= \EE_{Y \sim m} \big[ \|s_m(Y)\|_2^2 - 2 \lambda \langle s_m(Y), G(Y; \beta, \Sigma) \rangle
        + \lambda^2 \|G(Y; \beta, \Sigma)\|_2^2 \big].
    \end{aligned}
    $$

    Setting the first-order condition $\frac{\partial J_0}{\partial \lambda} = 0$ gives, when
    $\EE_{Y\sim m}[\|G(Y; \beta, \Sigma)\|_2^2]>0$:
    $$
    \lambda^*(\beta, \Sigma)
    = \frac{\EE_{Y \sim m}\left[\langle s_m(Y), G(Y; \beta, \Sigma) \rangle\right]}
    {\EE_{Y \sim m}\left[\|G(Y; \beta, \Sigma)\|^2_2\right]}.
    $$

    Now, in the anisotropic Gaussian case $\beta=2$, we have
    $$
    G(Y; 2, \Sigma) = \Sigma^{-1} \left( \EE[X\mid Y] - Y \right).
    $$
    Hence:
    $$
    \lambda^*(\beta=2, \Sigma)
    =
    \frac{\EE_{Y \sim m}\left[\left\langle s_m(Y), \Sigma^{-1} \left( \EE[X\mid Y] - Y \right) \right\rangle\right]}
    {\EE_{Y \sim m}\left[\left\|\Sigma^{-1} \left( \EE[X\mid Y] - Y \right)\right\|^2_2\right]}.
    $$
    Finally, in the isotropic case --- $\Sigma = \sigma^2 I_d$ --- we can cancel the variances in the numerator and
    denominator, giving:
    $$
    \lambda^*(\beta=2, \sigma^2 I_d)
    =
    \sigma^2
    \frac{\EE_{Y \sim m}\left[\left\langle s_m(Y), \EE[X\mid Y] - Y \right\rangle\right]}
    {\EE_{Y \sim m}\left[\left\|\EE[X\mid Y] - Y\right\|^2_2\right]}.
    $$

    \qed

    \clearpage
    \subsection{Appendix for Energy-Tweedie Diffusion}
    \label{app:diffusion}

    \subsubsection{Diffusion Along Arbitrary Noise-Parameter Paths: Eight Gaussians}
    \label{app:diffusion_full5}

    This section details the five-parameter Eight-Gaussians experiment summarized in Sec.~\ref{sec:diffusion}. The
    noise law is the generalized Gaussian of Eq.~\ref{eq:gen_gaussian} with the trace-normalized shape matrix
    $S(u,v)=\bigl(\begin{smallmatrix}1+u & v\\ v & 1-u\end{smallmatrix}\bigr)$ parameterized by disk
    coordinates $u^2+v^2<1$.
    Noise magnitude along a path is matched through the RMS displacement $r_t=r_{\eta_t}$ of
    Eq.~\ref{eq:noise_rms_scales} with $d=2$. The schedules below prescribe $r_t$, which fixes the reference RMS
    $\rho_t=r_t\,\lambda_t^{1/\beta_t}$ along any path.
    The full conditioning vector is $(\beta,\log\lambda,\log\rho,u,v)$.
    The four reverse sampling paths of Sec.~\ref{sec:diffusion}, stated over path progress $t\in[0,1]$, are:
    \begin{equation}
    \begin{aligned}
        \text{Isotropic, fixed tail:}\quad &(\beta_t,\lambda_t,u_t,v_t)=(1.4,\,1.8,\,0,\,0),\\
        \text{Fixed non-diagonal, fixed tail:}\quad &(\beta_t,\lambda_t,u_t,v_t)=(1.4,\,1.8,\,0,\,0.6),\\
        \text{Rotating covariance, fixed tail:}\quad &(\beta_t,\lambda_t)=(1.4,\,1.8),\quad
            (u_t,v_t)=0.6\,(\cos\theta_t,\sin\theta_t),\\ \theta_t=\tfrac{\pi}{2}+2\pi t,\\
        \text{Joint five-parameter path:}\quad &\beta_t=1.4+0.5\sin^2(\pi t),\quad \lambda_t=1.8+0.4\sin(2\pi t),\\
            (u_t,v_t)\ \text{as rotating}.
    \end{aligned}
    \label{eq:full5_paths}
    \end{equation}

\begin{figure}[!ht]
        \centering
        \includegraphics[width=.75\linewidth]{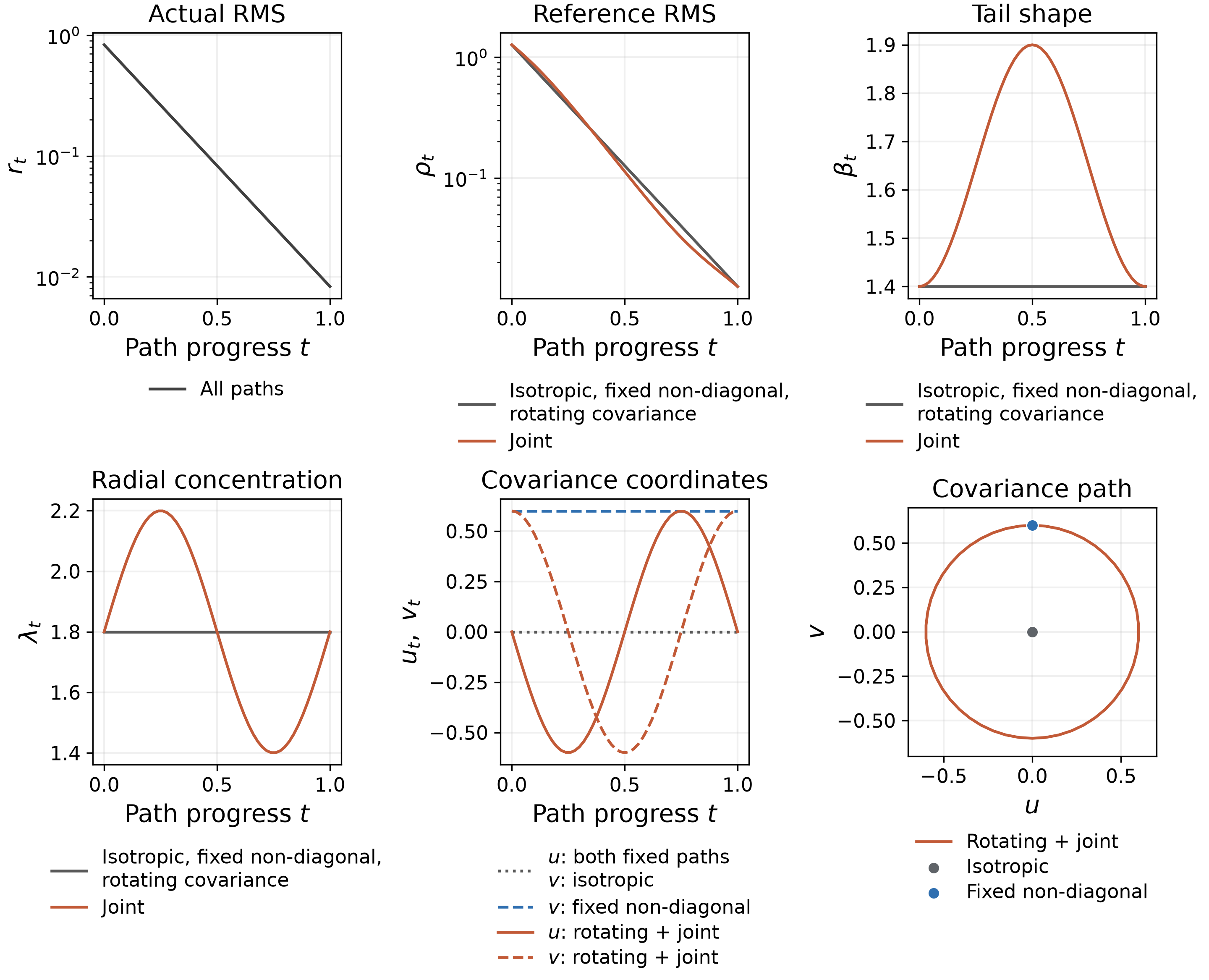}
        \caption{\textbf{Four reverse paths through the Eight-Gaussians noise-parameter space.}
        The panels show actual RMS $r_t$, reference RMS $\rho_t$, tail shape $\beta_t$, radial concentration
        $\lambda_t$, covariance coordinates $u_t,v_t$, and the path in the $(u,v)$ plane, with $t\in[0,1]$ denoting path progress.
        Each distinct trajectory is drawn once; panel legends identify the paths it represents.
        In the coordinate panel, solid and dashed lines denote $u_t$ and $v_t$, while the dotted line groups shared zero coordinates.
        Fixed covariance paths are shown as points.}
        \label{fig:full5_parameter_paths}
    \end{figure}

\begin{figure}[ht!]
        \centering
        \includegraphics[width=\linewidth]{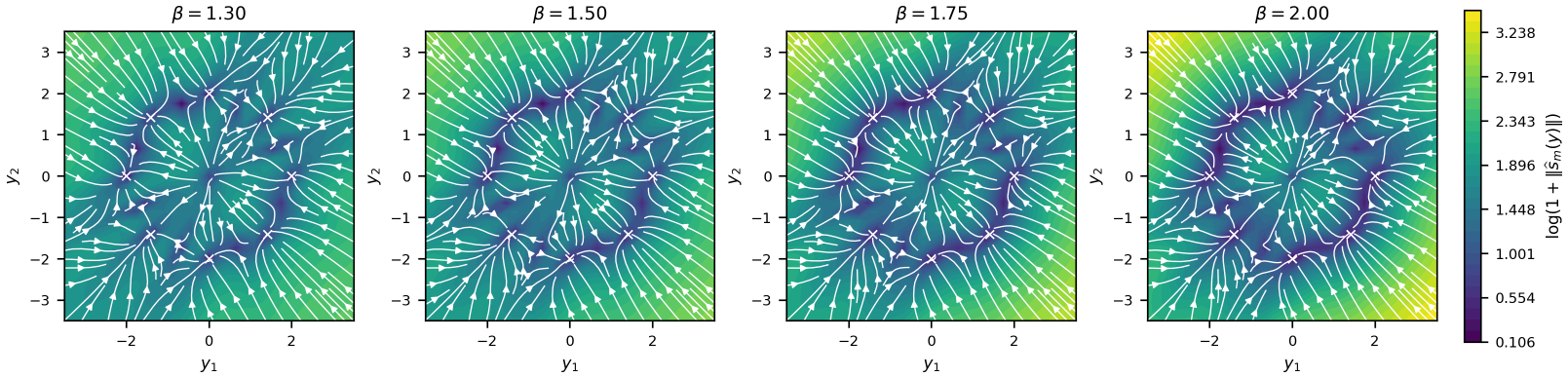}
        \includegraphics[width=\linewidth]{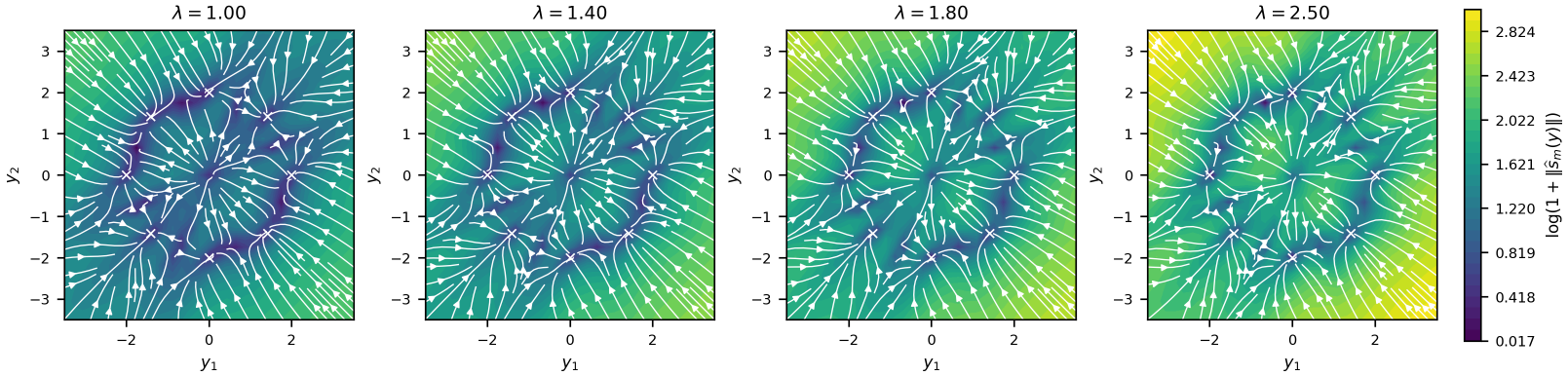}
        \includegraphics[width=\linewidth]{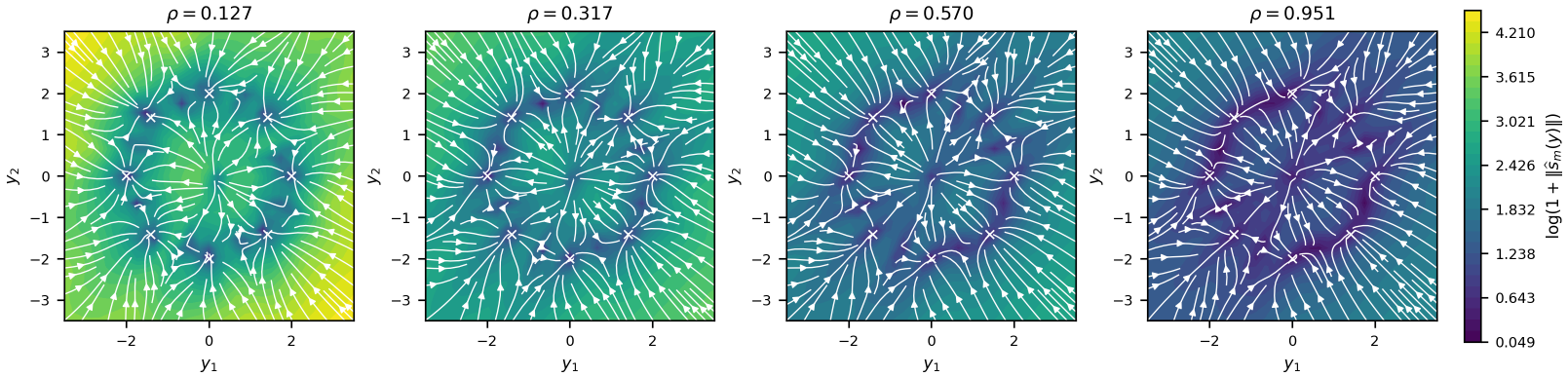}
        \includegraphics[width=\linewidth]{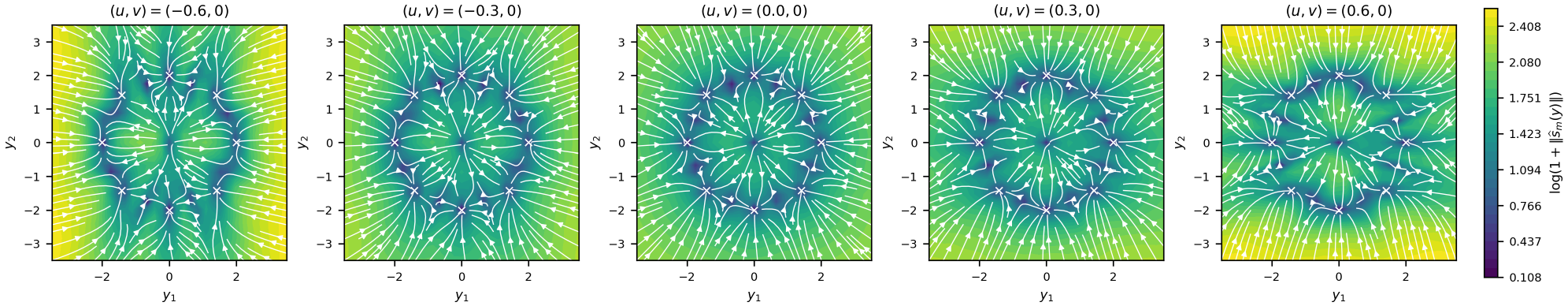}
        \includegraphics[width=\linewidth]{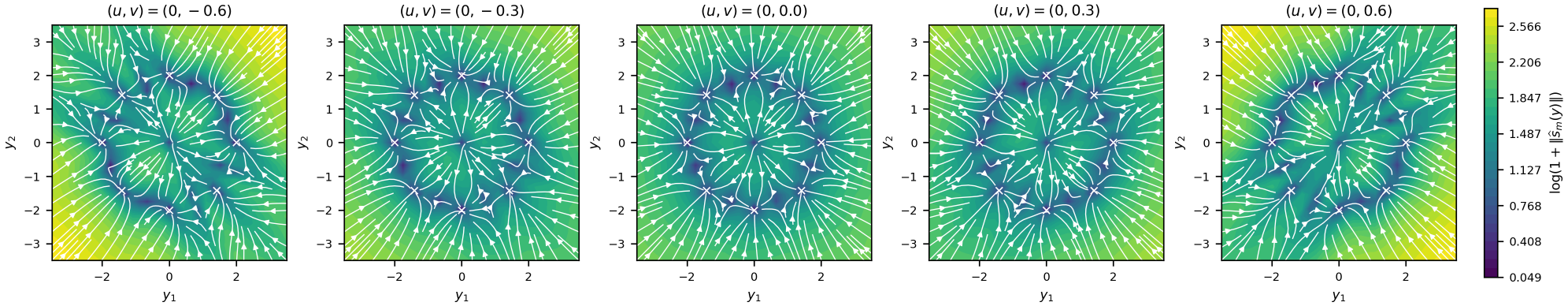}
        \caption{\textbf{One-parameter Energy-Tweedie score-field sweeps of the five-parameter conditional model.}}
        \label{fig:full5_score_sweeps}
    \end{figure}

    Figure~\ref{fig:full5_parameter_paths} displays the four noise-parameter paths.
    The parameter-path panel plots $r_t$, $\rho_t$, and $(\beta_t,\lambda_t,u_t,v_t)$ against path progress $t$; all four paths share the schedule $r_t$.
    Figure~\ref{fig:full5_score_sweeps} shows the full one-parameter score-field sweeps behind
    Fig.~\ref{fig:diffusion_score}: each row varies one conditioning parameter while the remaining parameters stay at
    the base parameter combination $(\beta,\lambda,\rho,u,v)=(1.4,\,1.8,\,0.57,\,0,\,0.6)$ (the $u$ and $v$ rows set the
    respective other disk coordinate to zero). From top to bottom, the rows use
    $\beta\in\{1.3,1.5,1.75,2.0\}$, $\lambda\in\{1.0,1.4,1.8,2.5\}$,
    $\rho\in\{0.13,0.32,0.57,0.95\}$ (the $\sigma$ sweep of App.~\ref{app:score_comparison}), $u\in\{-0.6,\dots,0.6\}$, and
    $v\in\{-0.6,\dots,0.6\}$. Streamlines show normalized score direction and color shows magnitude. The field
    responds to each parameter as expected: raising $\beta$
    toward $2$ moves the field from center-seeking toward the mean-seeking, ring-concentrating Gaussian regime,
    $\lambda$ and $\rho$ rescale the field and the noise magnitude, and $(u,v)$ tilt the field along the
    corresponding covariance geometry.
%
%
%

    \paragraph{Generation protocol.} We train two independent five-parameter conditional Engression models
    (training seeds 42 and 43) and sample each along the four reverse paths of Eq.~\ref{eq:full5_paths}, sharing
    one RMS schedule: $51$ geometrically spaced levels with $r_t$ decreasing from $0.83$ to $0.0083$. The
    rotating covariance completes a full rotation over the path, and the joint path matches $r_t$ at
    every level through $\rho_t$, with endpoints pinned to the rotating path --- the joint-vs-rotating comparison
    therefore isolates the intermediate route through $(\beta,\lambda,\rho)$.
    Sampling uses annealed Langevin with $10$ steps per level and
    $N=64$ posterior draws per score evaluation; each model is sampled with three seeds of $4{,}096$ final samples,
    pooled to $12{,}288$ samples per model and path. Signed energy distance to clean data is computed on fixed
    seed-stratified $4{,}096$-sample subsets against an independent $4{,}096$-point clean reference; mode coverage
    and radial error use all pooled samples.

    Table~\ref{tab:full5_diffusion_paths} reports mean $\pm$ sample
    standard deviation across the two trained models. Its endpoint metrics are signed energy distance to clean data,
    total variation from uniform mode weights, minimum mode mass, and mean radial distance to the ring; the final row
    is an independent clean-sample reference of the same size.

    \begin{table}[ht!]
        \centering
        \small
        \setlength{\tabcolsep}{4pt}
        \begin{tabular}{@{}lcccc@{}}
            \toprule
            Path & Signed ED & Mode-weight TV & Min.\ mode mass & Mean radial error \\
            \midrule
            Isotropic, fixed tail & $0.0058 \pm 0.0009$ & $0.017 \pm 0.002$ & $0.119 \pm 0.001$ & $0.233 \pm 0.001$ \\
            Fixed non-diagonal, fixed tail & $0.0081 \pm 0.0053$ & $0.039 \pm 0.006$ & $0.111 \pm 0.002$ & $0.237 \pm 0.002$ \\
            Rotating covariance, fixed tail & $0.0072 \pm 0.0027$ & $0.058 \pm 0.003$ & $0.104 \pm 0.003$ & $0.227 \pm 0.003$ \\
            Joint five-parameter path & $0.0072 \pm 0.0030$ & $0.055 \pm 0.005$ & $0.105 \pm 0.003$ & $0.220 \pm 0.005$ \\
            \midrule
            Independent clean draws & $0.0007 \pm 0.0017$ & $0.009 \pm 0.000$ & $0.121 \pm 0.000$ & $0.160 \pm 0.001$ \\
            \bottomrule
        \end{tabular}
        \caption{\textbf{Endpoint generation metrics for the four Eight-Gaussians noise-parameter paths.}}
        \label{tab:full5_diffusion_paths}
    \end{table}
    All eight modes retain more than $1\%$ mass on every path, run, and model. The four paths have similar endpoint
    estimates, and the paired joint-minus-rotating differences are negligible: signed ED
    $0.00002\pm0.00030$, mode-weight TV $-0.0031\pm0.0014$, minimum mode mass $0.0007\pm0.0001$, mean radial error
    $-0.0073\pm0.0020$. Routing the reverse process through simultaneously varying $(\beta,\lambda,\sigma)$ therefore
    incurs no measurable endpoint cost relative to the matched covariance-only control.

\begin{figure}[ht!]
        \centering
        \includegraphics[width=.9\linewidth]{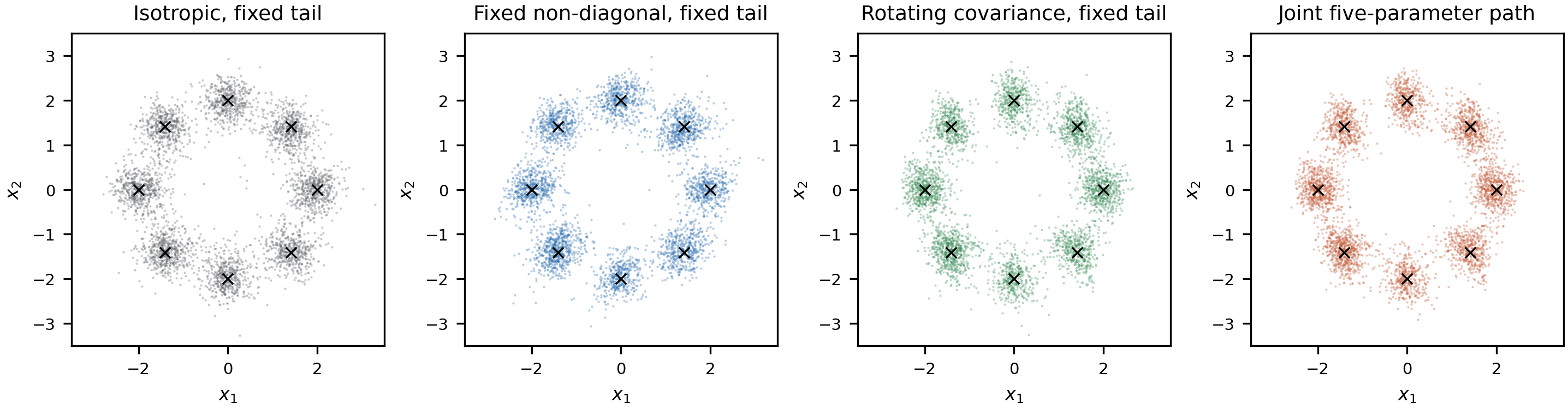}
        \caption{\textbf{Endpoint samples for each of the four paths (both trained models pooled), with the clean
            Eight-Gaussians reference.}}
        \label{fig:full5_final_samples}
    \end{figure}

\begin{figure}[ht!]
        \centering
        \includegraphics[width=\linewidth]{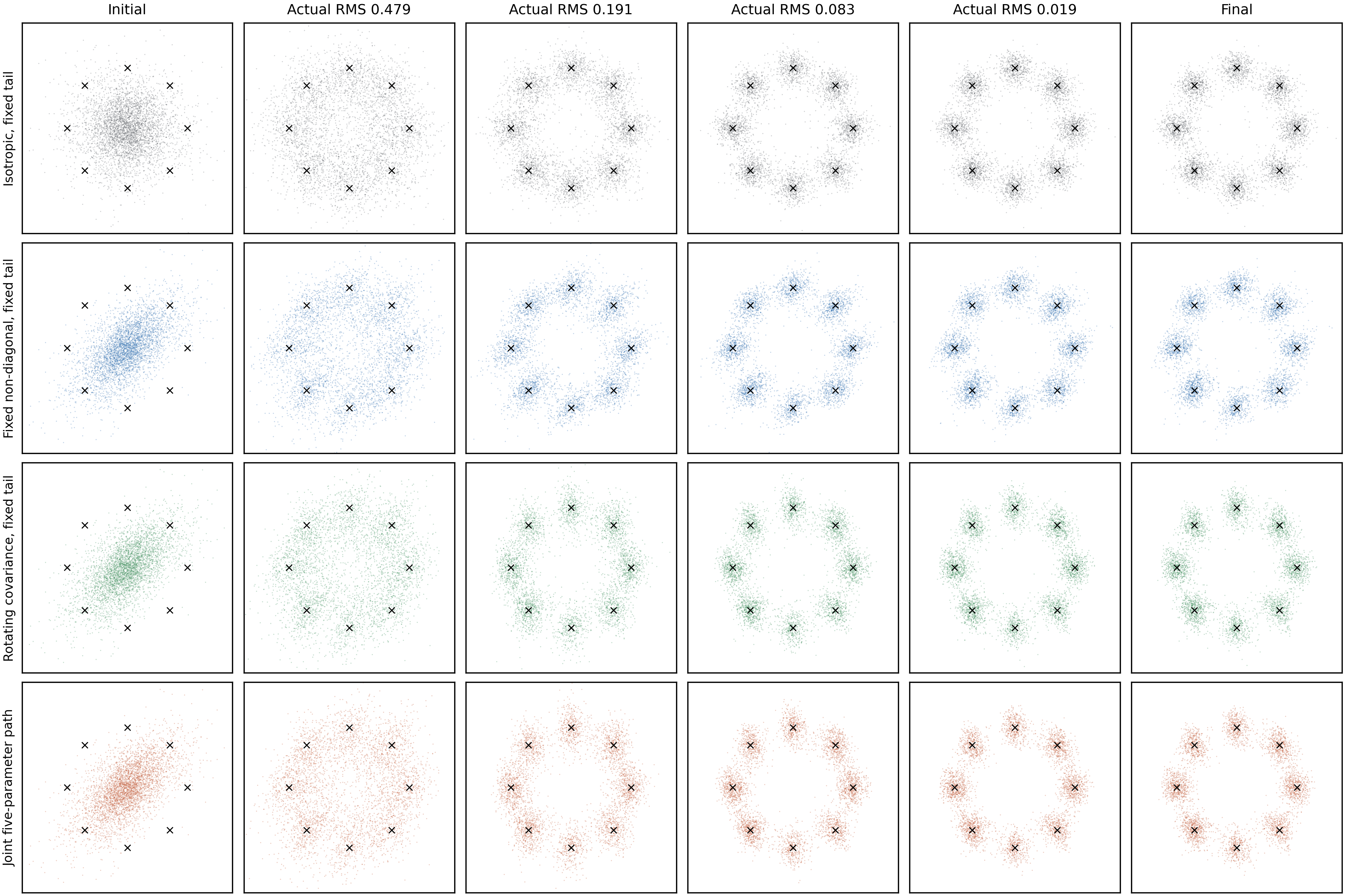}
        \caption{\textbf{Reverse-chain states along the shared RMS schedule $r_t$ for the four paths (rows), from the
            initial noise level to the endpoint (columns).}}
        \label{fig:full5_progress}
    \end{figure}

    Figures~\ref{fig:full5_final_samples} and~\ref{fig:full5_progress} show the endpoint samples and the
    reverse-chain states along the schedule.

\begin{figure}[ht!]
        \centering
        \includegraphics[width=.65\linewidth]{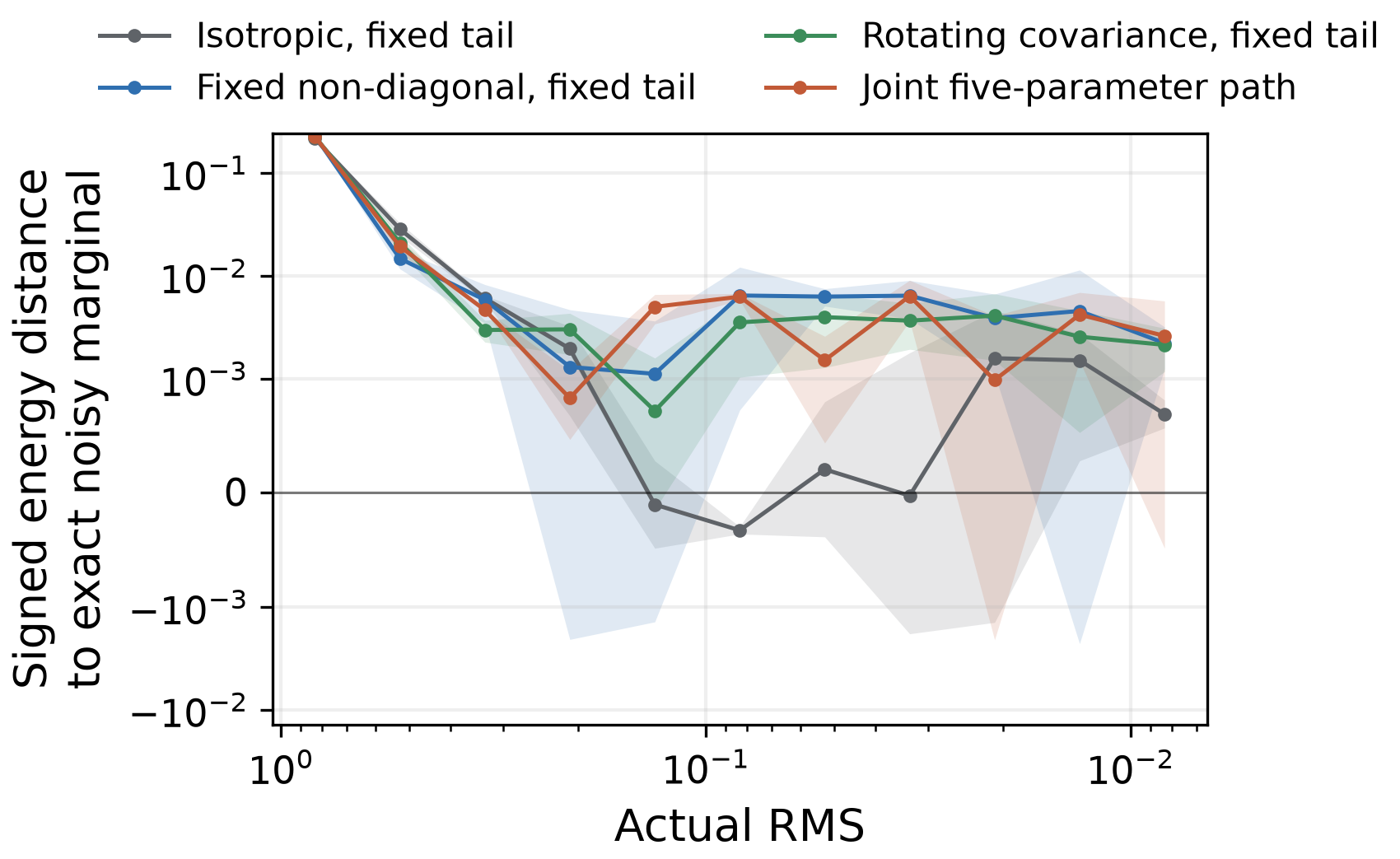}
        \caption{\textbf{Signed energy distance between the reverse-chain state and forward-noised clean data at the
            matching noise level, along the four paths.} Curves show mean $\pm$ sample standard deviation across the
            two trained models. The vertical axis uses a symmetric logarithmic scale.}
        \label{fig:full5_marginal_tracking}
    \end{figure}

    Figure~\ref{fig:full5_marginal_tracking} tracks the reverse chain against the \emph{exact noisy marginal} at each level: the distance drops from $0.015$--$0.028$ at $r_t=0.53$ to $0.003$--$0.006$ at $r_t=0.33$ and remains near
    zero thereafter on all four paths, indicating that the sampler follows the intended sequence of noisy marginals
    rather than only reaching a good endpoint. This check also covers the small-noise levels, where importance
    sampling no longer provides a reliable ground-truth score.

    \paragraph{Fixed-parameter path isolation.}
    We additionally compare \emph{constant-parameter} paths: along each path, $(\beta,\lambda,u,v)$ stay fixed and
    only the RMS $r_t$ decreases. Each arm changes a single coordinate of the reference value
    $(\beta,\lambda,u,v)=(1.4,1.8,0,0.6)$ and holds it for the entire path, so differences against the reference
    arm isolate the contribution of that coordinate. All arms use the same two trained five-parameter posterior models, three
    sampling seeds per model, a shared $49$-level RMS schedule $r_t$ from $0.8329$ to $0.0100$, and identical
    Langevin and posterior-sampling budgets. Each run produces $4{,}096$ endpoint draws; the three runs are pooled
    within each model before nonlinear metrics are computed. Table~\ref{tab:full5_fixed_parameter_paths} reports
    mean $\pm$ sample standard deviation across the two trained models. The tracking maximum is the largest signed
    energy distance to the exact noisy marginal over the post-burn-in checkpoints and is distinct from endpoint
    quality. Signed energy distance uses a fixed seed-stratified subset of $4{,}096$ generated draws and an
    independent clean reference of the same size; mode metrics use all $12{,}288$ pooled draws per trained model.
    \begin{table*}[ht!]
        \centering
        \small
        \resizebox{\textwidth}{!}{%
        \begin{tabular}{@{}llcccc@{}}
            \toprule
            Parameter & Fixed value & \shortstack{Signed\\ED $\downarrow$} & \shortstack{Mode-weight\\TV $\downarrow$}
                & \shortstack{Min. mode\\mass $\uparrow$} & \shortstack{Max. tracking\\ED $\downarrow$} \\
            \midrule
            \shortstack[l]{Clean-draw\\floor} & \shortstack[l]{Independent\\draws}
                & $0.00071 \pm 0.00171$ & $0.0090 \pm 0.0001$ & $0.1208 \pm 0.0002$ & --- \\
            \shortstack[l]{Ref. fixed\\path} & \shortstack[l]{$\beta=1.4,\lambda=1.8$\\$(u,v)=(0,0.6)$}
                & $0.00746 \pm 0.00560$ & $0.0461 \pm 0.0044$ & $0.1056 \pm 0.0009$ & $0.0267 \pm 0.0010$ \\
            $\beta$ & $1.3$
                & $0.00787 \pm 0.00621$ & $0.0470 \pm 0.0061$ & $0.1053 \pm 0.0001$ & $0.0246 \pm 0.0056$ \\
            $\beta$ & $1.6$
                & $0.00666 \pm 0.00506$ & $0.0432 \pm 0.0030$ & $0.1065 \pm 0.0030$ & $0.0276 \pm 0.0046$ \\
            $\beta$ & $1.8$
                & $0.00622 \pm 0.00554$ & $0.0391 \pm 0.0037$ & $0.1078 \pm 0.0030$ & $0.0206 \pm 0.0001$ \\
            $\beta$ & $2.0$
                & $0.00627 \pm 0.00674$ & $0.0337 \pm 0.0089$ & $0.1101 \pm 0.0018$ & $0.0355 \pm 0.0030$ \\
            $\lambda$ & $1.0$
                & $0.00771 \pm 0.00555$ & $0.0372 \pm 0.0061$ & $0.1119 \pm 0.0026$ & $0.0177 \pm 0.0035$ \\
            $\lambda$ & $2.5$
                & $0.00801 \pm 0.00642$ & $0.0439 \pm 0.0106$ & $0.1070 \pm 0.0019$ & $0.0227 \pm 0.0006$ \\
            $(u,v)$ & $(0,0)$
                & $0.00468 \pm 0.00088$ & $0.0249 \pm 0.0040$ & $0.1112 \pm 0.0006$ & $0.0344 \pm 0.0040$ \\
            $(u,v)$ & $(0.6,0)$
                & $0.00357 \pm 0.00038$ & $0.0220 \pm 0.0004$ & $0.1151 \pm 0.0006$ & $0.0193 \pm 0.0045$ \\
            $(u,v)$ & $(0,-0.6)$
                & $0.00431 \pm 0.00067$ & $0.0310 \pm 0.0012$ & $0.1097 \pm 0.0013$ & $0.0193 \pm 0.0007$ \\
            $(u,v)$ & $(-0.6,0)$
                & $0.00363 \pm 0.00134$ & $0.0300 \pm 0.0009$ & $0.1143 \pm 0.0033$ & $0.0226 \pm 0.0010$ \\
            \bottomrule
        \end{tabular}%
        }
        \caption{\textbf{Endpoint and post-burn-in marginal-tracking metrics for the fixed-parameter path isolation study.}}
        \label{tab:full5_fixed_parameter_paths}
    \end{table*}

    Covariance geometry gives the clearest endpoint differences. Relative to the reference fixed path,
    $(u,v)=(0.6,0)$ lowers signed ED from $0.00746$ to $0.00357$ and mode-weight TV from $0.0461$ to $0.0220$,
    while raising minimum mode mass from $0.1056$ to $0.1151$. The radial parameters have smaller or
    metric-dependent effects: $\beta=1.8$ improves all four reported metrics, while $\lambda=1$ gives the lowest
    tracking point estimate but nearly unchanged endpoint signed ED. Endpoint quality and worst-case marginal
    tracking therefore induce different descriptive rankings; with two trained-model seeds, the table is to be read more as a preliminary exploration.

    \newpage
    \subsubsection{Energy-Tweedie Diffusion on MNIST}
    \label{app:diffusion_mnist}
    This section details the MNIST experiment summarized in Sec.~\ref{sec:diffusion}. The posterior model is a single conditional Engression network (hidden width $2{,}560$) conditioned on the five-parameter noise specification $(\beta,\lambda,\rho,u,v)$, with the reference RMS $\rho$ of Eq.~\ref{eq:noise_rms_scales} and
    $d=784$. Training uses $\rho\in[0.025,6]$, $\beta\in[1.3,2]$,
    $\lambda\in[1,2.5]$, and covariance-disk radius $\|(u,v)\|\le 0.7$. The noise covariance shape is
    $\Sigma_0(u,v)=I_{392}\otimes S(u,v)$ with
    $S(u,v)=\bigl(\begin{smallmatrix}1+u & v\\ v & 1-u\end{smallmatrix}\bigr)$ acting on horizontally adjacent
    pixel pairs, so $(u,v)$ correlate neighboring pixels.

\begin{figure}[!ht]
        \centering
        \includegraphics[width=.75\linewidth]{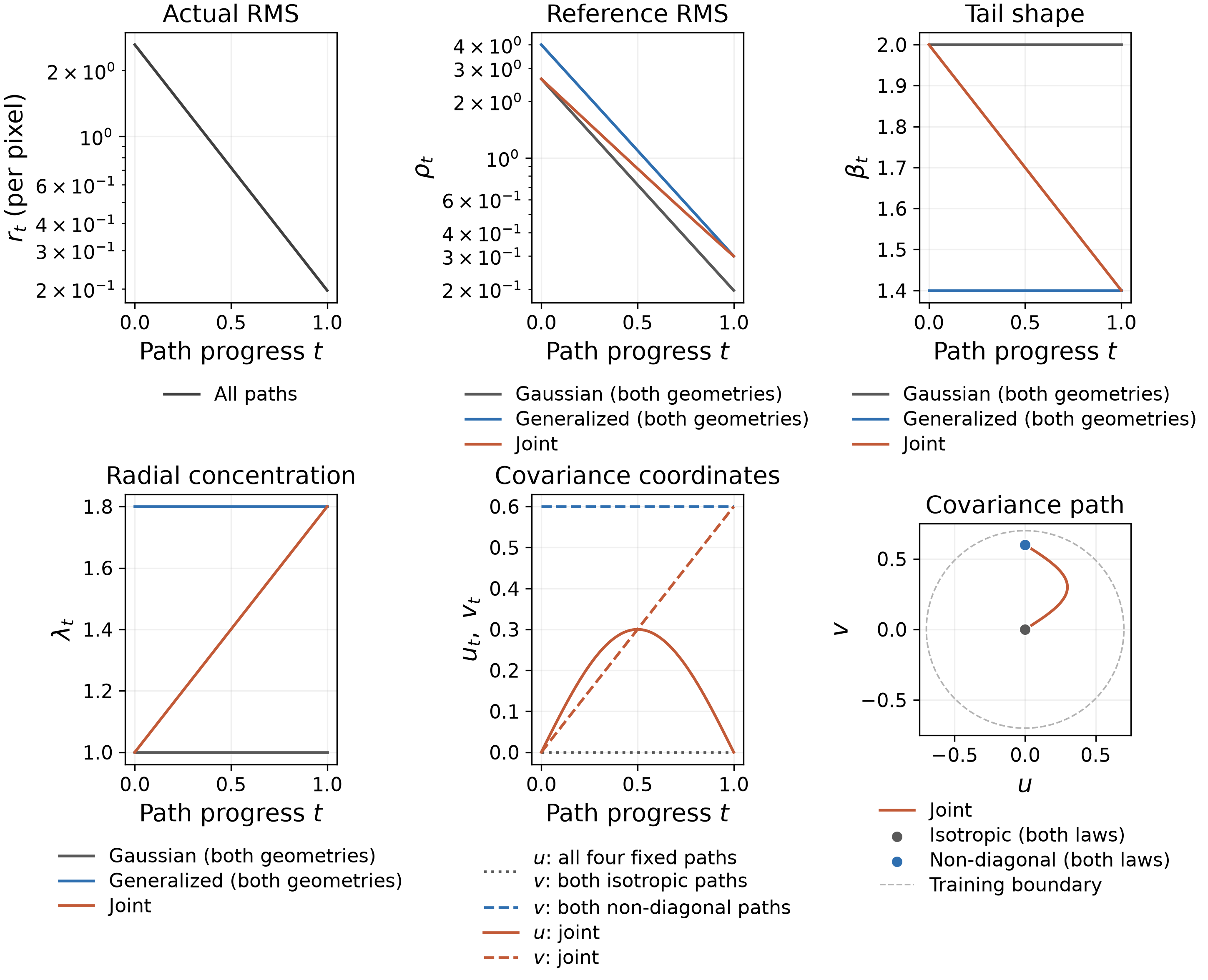}
        \caption{\textbf{Noise-magnitude, radial-law, and covariance trajectories for the five MNIST sampling paths.}
        The panels show actual per-pixel RMS $r_t$, reference RMS $\rho_t$, tail shape $\beta_t$, radial concentration
        $\lambda_t$, covariance coordinates $u_t,v_t$, and the path in the $(u,v)$ plane, with $t\in[0,1]$ denoting path progress.
        Each distinct trajectory is drawn once; panel legends identify the paths it represents.
        In the coordinate panel, solid and dashed lines denote $u_t$ and $v_t$, while the dotted line groups shared zero coordinates.
        Fixed covariance paths are shown as points; the dashed circle bounds the training region.}
        \label{fig:mnist_paths}
    \end{figure}

    \paragraph{Sampling paths.} All paths share one per-pixel RMS schedule $r_t$, geometric over $64$ levels with
    $r_t = 2.63\,(0.197/2.63)^{t}$, and differ only in the noise law and covariance geometry:
    \begin{equation}
    \begin{aligned}
        \text{Gaussian / isotropic:}\quad &(\beta_t,\lambda_t,u_t,v_t)=(2,\,1,\,0,\,0),\\
        \text{Gaussian / non-diagonal:}\quad &(\beta_t,\lambda_t,u_t,v_t)=(2,\,1,\,0,\,0.6),\\
        \text{Generalized / isotropic:}\quad &(\beta_t,\lambda_t,u_t,v_t)=(1.4,\,1.8,\,0,\,0),\\
        \text{Generalized / non-diagonal:}\quad &(\beta_t,\lambda_t,u_t,v_t)=(1.4,\,1.8,\,0,\,0.6),\\
        \text{Joint five-parameter path:}\quad &\beta_t=2-0.6t,\quad \lambda_t=1+0.8t,\quad u_t=0.3\sin(\pi t),\quad v_t=0.6t.
    \end{aligned}
    \label{eq:mnist_paths}
    \end{equation}
    Figure~\ref{fig:mnist_paths} displays the five sampling paths.
    The path panel plots $(\rho_t,r_t,\beta_t,\lambda_t,u_t,v_t)$ against path progress and includes the $(u,v)$  trajectories inside the training region.

    The Gaussian/isotropic path is ordinary Gaussian diffusion, where Eq.~\ref{eq:ES_diffusion_sample} reduces to
    the posterior-mean Tweedie residual. The joint path starts at the Gaussian/isotropic start and ends at the
    generalized/non-diagonal endpoint, so the corruption transitions continuously from isotropic Gaussian into a     heavy-tailed anisotropic law as the noise shrinks, and was not tuned for performance.

\todo{replaced three MNIST sampling paths with four fixed factorial paths and one jointly varying five-parameter path}
    Sampling uses annealed Langevin with $1{,}000$ updates at the first level and $100$ at each subsequent level,
    $N=256$ posterior draws per score evaluation, a shared step-size curve, and a final posterior draw at the
    terminal level $r_t=0.197$. Each path is sampled with three seeds ($2718$, $3141$, $5772$) of $256$ images,
    pooled to $768$ images per path; every generated image enters the evaluation without any selection.

    \todol{TODO: check this, check refs}
    \paragraph{Evaluation.} The primary metric is the signed unbiased energy distance between penultimate-layer
    features of a fixed, publicly available MNIST classifier (ONNX Model Zoo \textit{MNIST-12}; $256$-dimensional
    features; test accuracy $0.989$), computed between generated samples and a fixed bank of $2{,}048$ held-out real
    images. The classifier features discard pixel-level nuisance variation while any discrepancy in feature space
    shows a discrepancy between the underlying distributions; the energy distance admits an unbiased estimator and
    equals twice the squared maximum mean discrepancy for the associated distance-induced kernel
    \citep{szekely_energy_2017,sejdinovic_equivalence_2013}, making it a close relative of the kernel inception distance \citep{binkowski_demystifying_2018} while remaining within the energy-score toolkit used
    throughout this paper. Because the estimator is unbiased, its scale has a direct null reference: repeated
    real-vs-real draws of disjoint held-out images with the identical estimator and reference bank give a null
    distribution whose $95$th percentile at the pooled sample size is $0.017$. Companion metrics are the mean
    classifier confidence and the total variation between the predicted class distribution and the test-set class
    prior. Reported values are pooled point estimates over $768$ images per path with seed-stratified $95\%$
    bootstrap confidence intervals. The real-vs-real feature energy distance is reported without a percentile
    interval because the energy distance is a degenerate U-statistic at the null; the measured null distribution
    supplies its reference scale instead.
    \begin{table}[ht!]
        \centering
        \small
        \begin{tabular}{@{}lccc@{}}
            \toprule
            Path & Feature ED $\downarrow$ & Mean confidence $\uparrow$ & Class TV $\downarrow$ \\
            \midrule
            Real MNIST vs.\ real reference & $0.015$ & $0.993$ {\scriptsize$[0.989, 0.996]$} & $0.074$ {\scriptsize$[0.058, 0.117]$} \\
            Gaussian / isotropic & $0.574$ {\scriptsize$[0.506, 0.711]$} & $0.922$ {\scriptsize$[0.910, 0.934]$} & $0.162$ {\scriptsize$[0.133, 0.198]$} \\
            Gaussian / non-diagonal & $0.247$ {\scriptsize$[0.215, 0.360]$} & $0.964$ {\scriptsize$[0.956, 0.971]$} & $0.111$ {\scriptsize$[0.089, 0.150]$} \\
            Generalized / isotropic & $0.637$ {\scriptsize$[0.548, 0.808]$} & $0.908$ {\scriptsize$[0.893, 0.921]$} & $0.113$ {\scriptsize$[0.089, 0.152]$} \\
            Generalized / non-diagonal & $0.218$ {\scriptsize$[0.186, 0.324]$} & $0.966$ {\scriptsize$[0.959, 0.973]$} & $0.085$ {\scriptsize$[0.066, 0.124]$} \\
            Joint five-parameter path & $0.303$ {\scriptsize$[0.259, 0.424]$} & $0.960$ {\scriptsize$[0.951, 0.968]$} & $0.107$ {\scriptsize$[0.081, 0.142]$} \\
            \bottomrule
        \end{tabular}
        \caption{\textbf{Feature energy distance, classifier confidence, and class-distribution total variation for real
            MNIST and the five sampling paths.}}
        \label{tab:mnist_quantitative}
    \end{table}
\todo{replaced three-path MNIST metrics with the complete five-path quantitative comparison}

\begin{figure}[ht!]
        \centering
        \includegraphics[width=.95\textwidth]{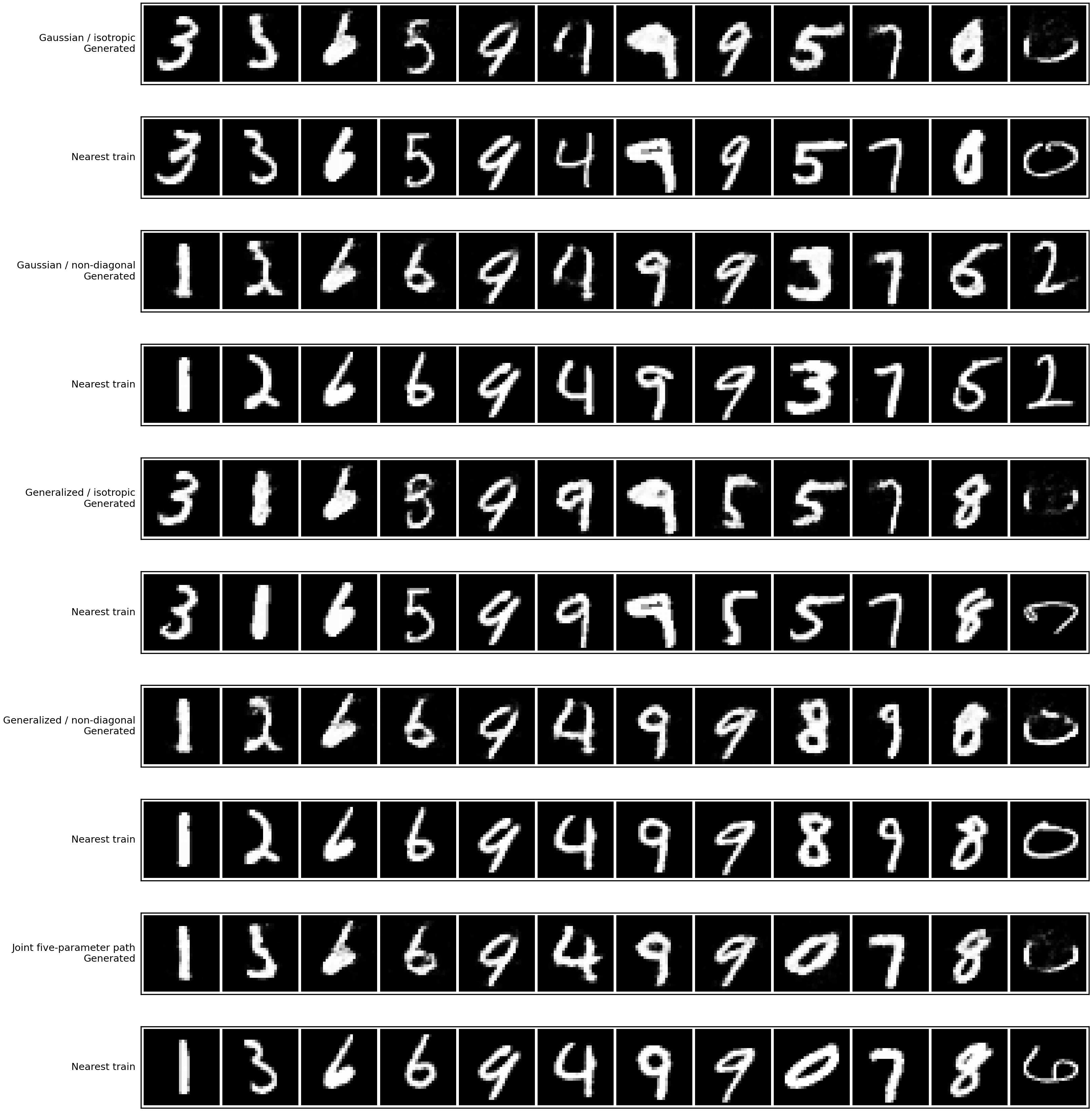}
        \caption{\textbf{Generated samples (upper rows) and their nearest MNIST training images in pixel RMSE (lower rows)
            for each of the five sampling paths.}}
        \label{fig:mnist_nearest_train}
    \end{figure}

\begin{figure}[ht!]
        \centering
        \includegraphics[width=.7\linewidth]{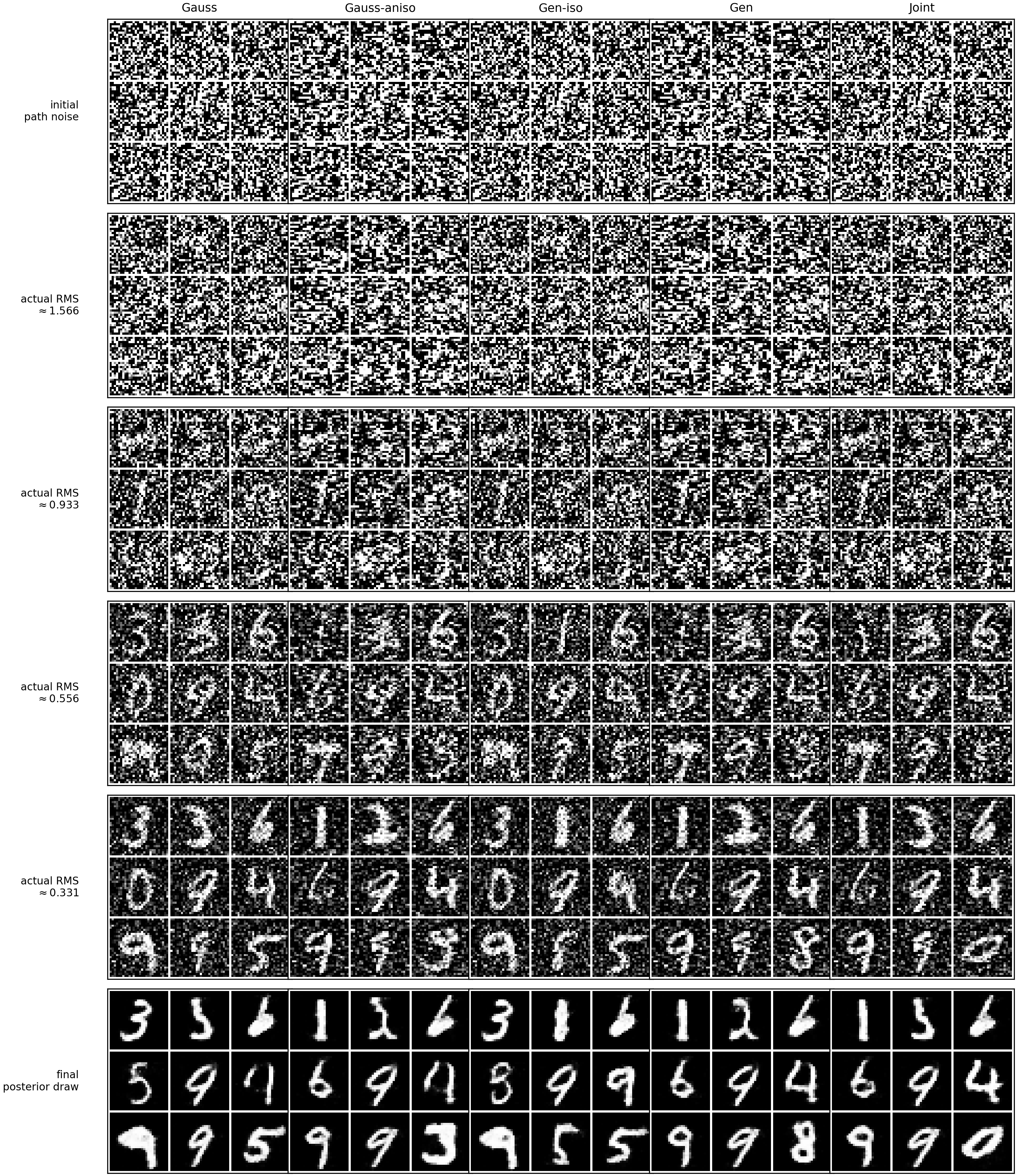}
        \caption{\textbf{Reverse-chain states for the five MNIST paths (columns) at matched per-pixel RMS levels $r_t$
            (rows), followed by the final posterior draw.} Each path--level cell contains the same nine chains in a
            $3\times3$ grid.}
        \label{fig:mnist_progress}
    \end{figure}

\begin{figure}[ht!]
        \centering
        \includegraphics[width=.9\linewidth]{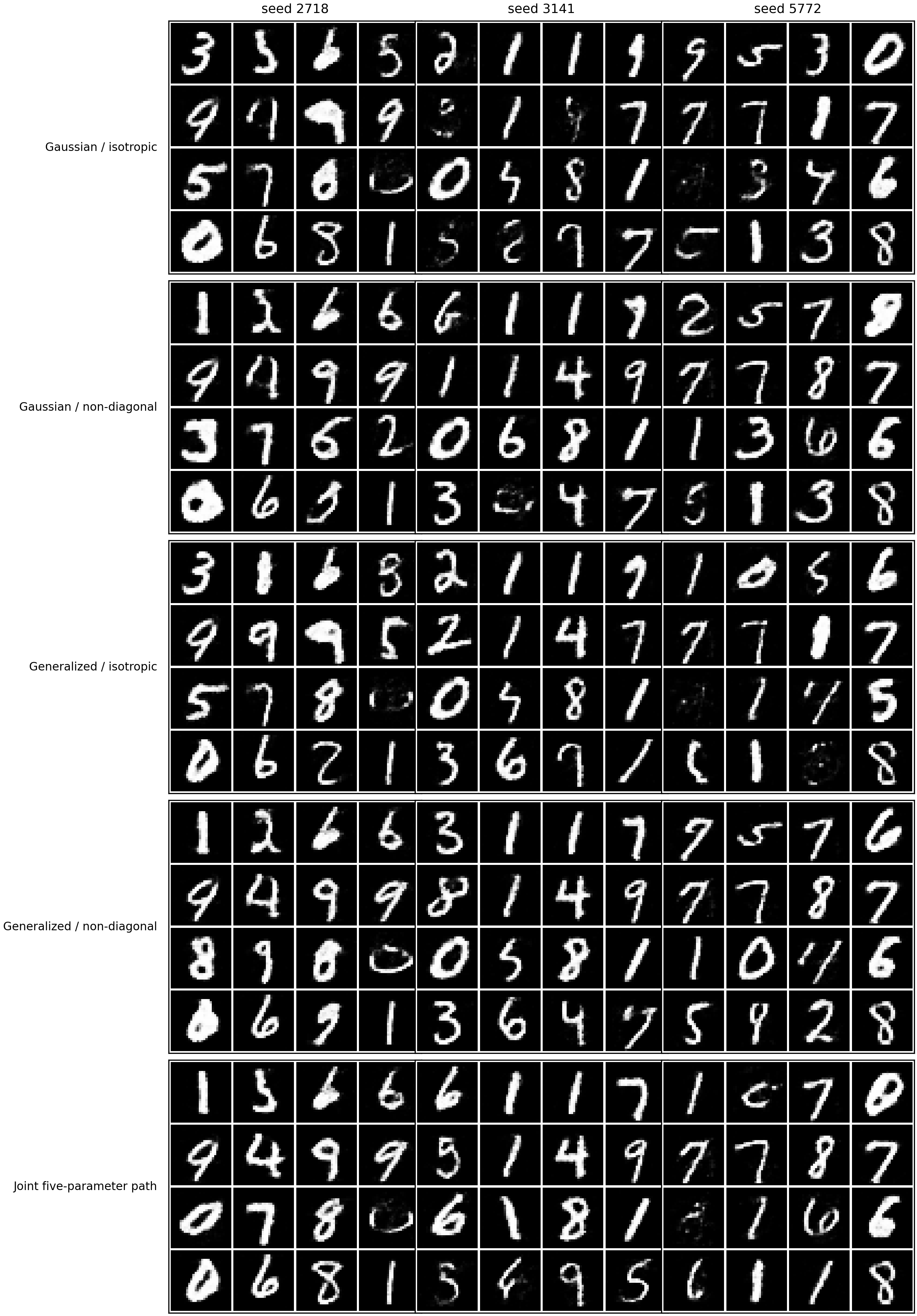}
        \caption{\textbf{Unselected endpoint samples for each path (rows) and sampling seed (columns).}}
        \label{fig:mnist_seeds}
    \end{figure}

    All ten classes are present on every path. The two fixed non-diagonal paths give the lowest feature energy
    distances. The generalized/non-diagonal path has the best point estimate on all three metrics, although its
    intervals overlap those of the Gaussian/non-diagonal path. Both isotropic paths have larger feature energy
    distance, while the joint path lies between the non-diagonal and isotropic controls. As a memorization check,
    mean pixel RMSE to the nearest training image is $0.139$--$0.143$ across the generated paths, compared with
    $0.147$ for held-out real images. Figure~\ref{fig:mnist_nearest_train} pairs generated examples with their nearest
    training neighbors, showing similar strokes and poses without exact pixel-space duplicates. It contains twelve
    unselected endpoint samples from the common figure seed for each path.
    Figures~\ref{fig:mnist_progress} and~\ref{fig:mnist_seeds} show the reverse chains at matched $r_t$ and the
    endpoint samples for each path and seed. The progress grid uses paths as columns and matched-RMS levels as rows,
    with the same nine chains in each cell.

    \clearpage
    \subsection{Further Connections}
    \label{app:connections}

    We present some further connective threads to the work presented in this section, specializing again  to the generalized Gaussian case.

    \subsubsection{Connection Between the Noise Geometry and the Energy Score Loss}
    \label{app:connections_geom}

    As indicated by Theorem~\ref{thm:et}, there is a correspondence between each \textit{noising} distribution
    parameterized by $\beta, \lambda, \Sigma$ and an energy score \textit{with the same parameters} whose
    path-derivative matches the score of the data noised by the former.

    One perspective is to notice that the Mahalanobis distance $\|\cdot\|_{\Sigma^{-1}}$ can be associated to the
    (position-independent) Riemannian inner product $\langle u, v \rangle = u^T \Sigma^{-1} v$. We thus get for the
    Riemannian gradient $\operatorname{grad}_{\Sigma^{-1}} f(x) = \Sigma~\nabla_x f(x).$ Now, one can define the
    Riemannian path-derivative analogously to the ordinary one, and get that, by Eq.~\ref{eq:gen_gaussian_score} the
    noisy score is proportional to the Riemannian gradient of some potential/energy:
    $$
    s_m(y) = -\lambda ~ \Sigma^{-1} \, \EE\left[  \|y - X\|_{\Sigma^{-1}}^{\beta - 2} ~ (y - X) \mid Y=y \right] =
        -\lambda ~ \Sigma^{-1} \operatorname{grad}^{\mathrm{PD}}_{\Sigma^{-1}} \phi(y),
    $$
    where we can solve for $\phi(y)$ and obtain:
    $$
    \phi(y) = \frac{1}{\beta}~ \EE  [ \|y - X \|_{\Sigma^{-1}}^{\beta} \mid Y=y ],
    $$
    i.e., the ``mean $\beta$-energy'' associated with the \textit{posterior} (w.r.t.~$Y$) of the generalized Gaussian
    elliptical distribution (i.e., ``energy model'') --- cf. Eq.~\ref{eq:gen_gaussian} --- in the geometry induced by
    $\Sigma$.


    \subsubsection{Denoising}
    \label{app:denoising}

    Denoising can be seen as both a problem setting and a generative modeling approach
    \citep{milanfar_denoising_2024}. In this subsection, we first focus on the denoising problem as such, with the
    generative modeling presented in Sec.~\ref{sec:diffusion}.

    In this case, the posterior $P(X\mid Y=y)$ can be considered a \textit{distributional denoiser} with $P_\theta(X\mid
    Y=y)$ denoting its estimator. An example of score-based denoising is \citet{kim_noise2score_2021}, who have focused
    on using the standard Tweedie formula in the denoising context, using denoising autoencoders
    \citep{vincent_extracting_2008} to extract the score. The vast majority of denoising work limits the noise to the
    isotropic Gaussian $\gN(0, \sigma^2 I)$.

    An interesting perspective is to consider \textit{ideal denoisers} in light of our results.
    As noted in \citet{milanfar_denoising_2024}, if $f(y, \alpha)$ denotes the denoising functional for noise level
        parameters $\alpha$%
    \footnote{One can consider $\alpha$ to be a vector in this case.}%
    , then the ideal denoiser will be a gradient of some scalar function (the ``energy'' or potential):
    $$
    f^\star(y, \alpha)=\nabla \gU(y, \alpha),
    $$
    which is equivalent to requiring that its Jacobian be \textit{symmetric} in the isotropic Gaussian case. Thus, the
    denoiser itself defines a \textit{conservative vector field}. We will be focusing on MSE-optimal denoisers in the
    following discussion.

    In the \textit{isotropic} Gaussian case, this is readily apparent from the ``basic'' Tweedie formula:
    $$
    f^\star(y, \sigma^2)\; = \;\EE[X\!\mid Y{=}y]\; =\; y+\sigma^2\,\nabla_y \log m(y)
    $$
    implies for the Jacobian:
    $$\nabla f^\star(y, \sigma^2) = I + \sigma^2\nabla^2 \! \log m(y),$$
    which is obviously symmetric.

    Moving to the \textit{anisotropic} Gaussian case --- $\alpha=(\beta=2, \lambda=1, \Sigma)$ --- we still have
    Eq.~\ref{eq:tweedie}:
    $$
    f^\star(y, \alpha)\; =\;y + \Sigma\,\nabla_y \log m(y).
    $$

    By App.~\ref{app:connections_geom}, this can be seen as the Riemannian gradient in the geometry induced by $\Sigma$:
    \begin{equation}
        f^\star(y, \alpha)\;=\;\operatorname{grad}_{\Sigma^{-1}} \! \underbrace{\Big(\tfrac12\|y\|^2_{\Sigma^{-1}}+\log
            m(y)\Big)}_{\gU},
        \label{eq:riemannian_denoiser}
    \end{equation}
    where $\gU$ is the potential function mentioned above for this case.
    The Jacobian of such an ideal denoiser:
    $$\nabla f^\star(y) = I + \Sigma \nabla^2\!\log m(y)$$
    is not necessarily symmetric (as it depends on $\Sigma$, i.e. the anisotropy). However, we have:
    \begin{equation}
        (\nabla f^\star(y))^\top \Sigma^{-1} = \Sigma^{-1} \nabla f^\star(y),
    \end{equation}
    i.e., it is self-adjoint in the $\Sigma^{-1}$ geometry, or equivalently, it is symmetric in whitened coordinates.
    This represents a generalization of the criterion for ideal denoisers to \textit{anisotropic} Gaussian noise ---
    \textit{self-adjointness}.

    A common alternative MAP-optimal denoiser $f_{\mathrm{MAP}}(y, \alpha)$ can be obtained from the \textit{proximal operator}
    \citep{milanfar_denoising_2024, boyd_convex_2004}:
    $$
    \operatorname{prox}_{R}^{\Sigma^{-1}} (y) = \argmin_x (\tfrac12 \|x - y\|_{\Sigma^{-1}}^2 + R(x)),
    $$

    where $R(x)$ is the negative log-prior on the clean data $x$. Under additive Gaussian noise, the MAP estimator is the proximal
    operator of the negative log-prior. If the prior is also Gaussian, then the posterior is Gaussian, so the MAP
    estimate coincides with the posterior mean (equivalently, the Wiener filter); hence the resulting field matches
    Eq.~\ref{eq:riemannian_denoiser}.

    Finally, still in the anisotropic Gaussian case, we have by Tweedie:
    $$f^\star(y) - y = \Sigma\, s_m(y)\;.$$
    However, by Thm.~\ref{thm:et}, we also have:
    \begin{equation}
        f^\star(y) - y = - \frac{\Sigma}{2} \nabla_y^{\mathrm{PD}}
            \mathrm{ES}_{\Sigma^{-1}, 2}\big(P(X \mid Y=y), y\big).
        \label{eq:denoiser_es}
    \end{equation}

    Thus, the potential $\gU$ that generates the conservative field is defined \textbf{\textit{either}} by the
    \textit{log-marginal} $\log m(y)$ (as previously shown in Eq.~\ref{eq:riemannian_denoiser}) \textbf{\textit{or}} the
    corresponding energy score as denoted by Eq.~\ref{eq:denoiser_es}:
    \begin{equation}
    \begin{aligned}
        \operatorname{grad}_{\Sigma^{-1}}\!\Big(\tfrac12\|y\|_{\Sigma^{-1}}^{2} \, + \, \log m(y)\Big) \; &= \;
            f^\star(y) \\
            &= \; \operatorname{grad}^{\mathrm{PD}}_{\Sigma^{-1}}\!\Big(\tfrac12\|y\|_{\Sigma^{-1}}^{2}
            \; - \; \tfrac12\,\mathrm{ES}_{\Sigma^{-1},2}\!\big(P(X\mid Y=y),y\big)\Big),
    \end{aligned}
        \label{eq:denoiser_equality}
    \end{equation}
    with the caveat that in the latter case, we need to take the \textit{path derivative} $\nabla_y^{\mathrm{PD}}$
    (Eq.~\ref{eq:path_derivative}).

    \subsection{Model Architectures and Training}
    \label{app:training}

Our posterior models adapt the implementation of \citet{shen_engression_2024}%
    \footnote{\url{https://github.com/xwshen51/engression}}%
    : stochastic residual MLPs conditioned on the noise parameters,
    with architectures summarized in Table~\ref{tab:posterior_training}.
    We train these models on clean--noisy pairs using the Euclidean energy score with exponent one for the non-diffusion use cases.
    Noise parameters enter through feature-wise linear modulation (FiLM): the noise scale for isotropic
    models, scale and covariance shape for parameter estimation, and all five noise parameters for the
    five-parameter Eight-Gaussians and MNIST models.

    \begin{table}[htbp]
        \centering
        \small
        \setlength{\tabcolsep}{3pt}
        \begin{tabular}{@{}lccrcr@{}}
            \toprule
            Posterior & $(L,H,Z)$ & FiLM width & Batch size (k) & Learning rate & Steps (k) \\
            \midrule
            Gaussian, isotropic & $(6,200,100)$ & $16$ & $40$ & $2\cdot10^{-5}\to2\cdot10^{-6}$ & $240$ \\
            Generalized, isotropic & $(6,100,100)$ & linear & $80$ & $3\cdot10^{-5}\to2\cdot10^{-6}$ & $200$ \\
            Five-parameter & $(6,250,100)$ & $64$ & $40$ & $1.5\cdot10^{-5}\to10^{-6}$ & $280$ \\
            Parameter estimation & $(6,250,100)$ & $64$ & $40$ & $1.5\cdot10^{-5}\to10^{-6}$ & $280$ \\
            MNIST & $(16,2560,128)$ & $256$ & $1.536$ & $1.5\cdot10^{-5}\to1.5\cdot10^{-6}$ & $100$ \\
            \bottomrule
        \end{tabular}
        \caption{\textbf{Posterior architectures and training budgets.} The first four rows use Eight Gaussians.
        $L$ counts affine layers in the residual branches (two per block), $H$ is the hidden width, and $Z$
        is the Gaussian noise dimension per layer. FiLM width is the hidden width of the conditioning MLP;
        ``linear'' denotes a single affine map. Batch sizes and optimizer steps are reported in thousands.}
        \label{tab:posterior_training}
    \end{table}

    All posteriors use Adam with cosine learning-rate decay. Eight-Gaussians training accumulates two
    minibatches per optimizer step; MNIST uses one, with a $500$-step learning-rate warmup.
    Each fixed or five-parameter Eight-Gaussians posterior is reused for score estimation and diffusion;
    all five MNIST sampling paths share one posterior model.

    The five-parameter Eight-Gaussians posterior draws one noise-parameter combination per minibatch: with probability
    $0.35$ one of $18$ preselected combinations, namely $(\beta,\lambda)=(1.4,1.8)$ at the reference RMS levels
    $\rho\in\{0.32,0.57,0.95\}$ (Eq.~\ref{eq:noise_rms_scales}) and the covariance shapes
    $(u,v)\in\{(0,0),(0.6,0),(0,0.6),(0.6/\sqrt2,0.6/\sqrt2)\}$ together with six extremal parameter combinations, and
    otherwise $\beta$ uniform on $[1.3,2]$, $\lambda$ and $\rho$ log-uniform on $[1,2.5]$ and $[0.01,10]$, and
    $(u,v)$ uniform in area on the disk of radius $0.7$. 

    For plug-in parameter estimation, we reuse one score estimation network across the three noise-level settings in each replicate. Conditioning on the known log noise level avoids training three separate Hyv\"arinen networks; the method itself is parameter-agnostic (i.e., this is to save on compute, not required by the method).
    Each network has three hidden layers of width $256$ with SiLU activations and is trained by exact Hyv\"arinen score matching, using AdamW for $40{,}000$ steps with batch size $4{,}096$ and learning rate $10^{-3}$. Complete training configurations, seeds, and preprocessing are provided in the supplement.

    For the direct-score comparison in Sec.~\ref{sec:score_estimation}, ET-DSM and Hyv\"arinen use the
    MLP architecture described above, with seven inputs:
    the two noisy coordinates and the normalized conditioning vector $(\beta,\log\lambda,\log\rho,u,v)$,
    with the reference RMS $\rho$ defined in Eq.~\ref{eq:noise_rms_scales}.
    The linear output layer outputs the two score coordinates.

    We train each network for $80{,}000$ AdamW updates with batch size $4{,}096$, cosine learning-rate
    decay from $10^{-3}$ to $10^{-5}$, zero weight decay, and gradient clipping at norm $10$.
    Each minibatch uses one noise-parameter combination; the two estimators receive the same minibatches in order.

%
    \clearpage

    \section*{NeurIPS Paper Checklist}
    \begin{enumerate}

        \item {\bf Claims}
        \item[] Question: Do the main claims made in the abstract and introduction accurately reflect the paper's
            contributions and scope?
        \item[] Answer: \answerYes{}
        \item[] Justification: The abstract and introduction state the paper's main contributions: the Energy--Tweedie identity, its relationship to the classical setting, and its consequences for score estimation, noise-parameter estimation, and diffusion-style sampling.

        \item {\bf Limitations}
        \item[] Question: Does the paper discuss the limitations of the work performed by the authors?
        \item[] Answer: \answerYes{}
        \item[] Justification: The Discussion lists the limitations: the identity holds for the (large) additive Gibbs noise model, the noise-parameter estimation theory assumes a correctly specified posterior term and a sufficiently accurate auxiliary score estimator, the experiments are specialized to the generalized Gaussian case on low-dimensional synthetic data and MNIST, and large-scale image-generation benchmarks are out of scope.

        \item {\bf Theory assumptions and proofs}
        \item[] Question: For each theoretical result, does the paper provide the full set of assumptions and a complete
            (and correct) proof?
        \item[] Answer: \answerYes{}
        \item[] Justification: The theoretical results state their assumptions in the main text or immediately adjacent discussion, and the formal proofs are provided in the appendix.

        \item {\bf Experimental result reproducibility}
        \item[] Question: Does the paper fully disclose all the information needed to reproduce the main experimental
            results of the paper to the extent that it affects the main claims and/or conclusions of the paper
            (regardless of whether the code and data are provided or not)?
        \item[] Answer: \answerYes{}
        \item[] Justification: The experiments use well-known synthetic datasets and MNIST, and all the settings and steps needed to
                reproduce them are included.

        \item {\bf Open access to data and code}
        \item[] Question: Does the paper provide open access to the data and code, with sufficient instructions to
            faithfully reproduce the main experimental results, as described in supplemental material?
        \item[] Answer: \answerYes{}
        \item[] Justification: The data used in the experiments are synthetically generated or public (MNIST); the code, configurations, trained models, and instructions for reproducing the experimental results are released at \url{https://github.com/andleb/et}.

        \item {\bf Experimental setting/details}
        \item[] Question: Does the paper specify all the training and test details (e.g., data splits, hyperparameters,
            how they were chosen, type of optimizer) necessary to understand the results?
        \item[] Answer: \answerYes{}
        \item[] Justification: We specify the main details in the paper and the rest in the appendix and the public code repository.

        \item {\bf Experiment statistical significance}
        \item[] Question: Does the paper report error bars suitably and correctly defined or other appropriate
            information about the statistical significance of the experiments?
        \item[] Answer: \answerYes{}
        \item[] Justification: Where applicable, the reported experimental results include error bars or equivalent uncertainty summaries for the experiments supporting the empirical claims.

        \item {\bf Experiments compute resources}
        \item[] Question: For each experiment, does the paper provide sufficient information on the computer resources
            (type of compute workers, memory, time of execution) needed to reproduce the experiments?
        \item[] Answer: \answerYes{}
        \item[] Justification: The compute resources used are documented in the public code repository.

        \item {\bf Code of ethics}
        \item[] Question: Does the research conducted in the paper conform, in every respect, with the NeurIPS Code of
            Ethics \url{https://neurips.cc/public/EthicsGuidelines}?
        \item[] Answer: \answerYes{}
        \item[] Justification: The work conforms to the NeurIPS Code of Ethics.

        \item {\bf Broader impacts}
        \item[] Question: Does the paper discuss both potential positive societal impacts and negative societal impacts
            of the work performed?
        \item[] Answer: \answerYes{}
        \item[] Justification: The work is primarily theoretical and methodological: it derives fundamental identities
                and  validates them in controlled synthetic experiments.
                Accordingly, we do not identify a direct negative societal impact beyond the general possibility
                that improved score estimation or generative-modeling methodology could eventually be incorporated
                into downstream systems.

        \item {\bf Safeguards}
        \item[] Question: Does the paper describe safeguards that have been put in place for responsible release of data
            or models that have a high risk for misuse (e.g., pre-trained language models, image generators, or scraped
            datasets)?
        \item[] Answer: \answerNA{}
        \item[] Justification: The paper does not release a high-risk pretrained model, dataset, image generator, or other asset requiring controlled access safeguards.

        \item {\bf Licenses for existing assets}
        \item[] Question: Are the creators or original owners of assets (e.g., code, data, models), used in the paper,
            properly credited and are the license and terms of use explicitly mentioned and properly respected?
        \item[] Answer: \answerYes{}
        \item[] Justification: The public code repository documents the licenses of the software dependencies and third-party assets used to run the experiments.

        \item {\bf New assets}
        \item[] Question: Are new assets introduced in the paper well documented and is the documentation provided
            alongside the assets?
        \item[] Answer: \answerYes{}
        \item[] Justification: The paper releases new experimental code, synthetic-data generation scripts, and configuration files needed to reproduce the results.

        \item {\bf Crowdsourcing and research with human subjects}
        \item[] Question: For crowdsourcing experiments and research with human subjects, does the paper include the
            full text of instructions given to participants and screenshots, if applicable, as well as details about
            compensation (if any)?
        \item[] Answer: \answerNA{} 
        \item[] Justification: The paper does not involve crowdsourcing or research with human subjects.

        \item {\bf Institutional review board (IRB) approvals or equivalent for research with human subjects}
        \item[] Question: Does the paper describe potential risks incurred by study participants, whether such risks
            were disclosed to the subjects, and whether Institutional Review Board (IRB) approvals (or an equivalent
            approval/review based on the requirements of your country or institution) were obtained?
        \item[] Answer: \answerNA{} 
        \item[] Justification: The paper does not involve human subjects research.

        \item {\bf Declaration of LLM usage}
        \item[] Question: Does the paper describe the usage of LLMs if it is an important, original, or non-standard
            component of the core methods in this research? Note that if the LLM is used only for writing, editing, or
            formatting purposes and does \emph{not} impact the core methodology, scientific rigor, or originality of the
            research, declaration is not required.
        \item[] Answer: \answerNA{} 
        \item[] Justification: LLMs are not an important, original, or non-standard component of the core methods in this work. The use of LLMs was limited to ordinary writing, editing, formatting, and basic code-assistance purposes and thus did not affect the core methodology or scientific claims.

    \end{enumerate}

\end{document}